\documentclass{IEEEtran}
\usepackage[utf8]{inputenc}
\usepackage{amsmath}
\usepackage{amssymb}
\usepackage{amsthm}
\usepackage{xcolor}
\usepackage{mathtools}
\usepackage{soul}
\usepackage{url}  
\usepackage{algorithm}
\usepackage[noend]{algpseudocode}
\usepackage{hyperref}
\usepackage{enumitem}
\usepackage[style=ieee, backend=biber]{biblatex}
\bibliography{main}
\AtBeginBibliography{\footnotesize}

\usepackage{multirow}

\usepackage{graphicx}
\usepackage{subcaption}
\usepackage{todonotes}
\usepackage{booktabs}

\newtheorem{definition}{Definition}

\newif\ifdraft
\draftfalse
\usepackage{xparse}
\usepackage{xspace}

\makeatletter
\def\@addpunct2#1{\ifnum\spacefactor>\@m \else#1\fi}
\newcommand{\para}[1]{\noindent\textbf{#1\unskip\@addpunct2{.}}~~}
\makeatother

\newcommand{\adv}{\mathcal{A}}
\newcommand{\prover}{\mathcal{T}}
\newcommand{\encryptedPw}{\mathcal{R}}
\newcommand{\dataset}{D}
\newcommand{\model}{f}
\newcommand{\fingerprint}[1][]{}
\NewDocumentCommand{\pw}{o}{
  \IfNoValueTF{#1}
    {\mathcal{P}}
    {\mathcal{P}(#1)}%
}
\newcommand{\prf}{PoL\xspace}
\MakeRobust\prf

\NewDocumentCommand{\oracle}{o}{
  \IfNoValueTF{#1}
    {Or}
    {Or(#1)}%
}
\NewDocumentCommand{\verifier}{o}{
  \IfNoValueTF{#1}
    {\mathcal{V}}
    {\mathcal{V}(#1)}%
}

\newcommand{\vsr}[2]{\operatorname{VSR}\left(#1, #2\right)}
\newcommand{\encrypt}[1]{\operatorname{enc}(#1)}
\newcommand{\decrypt}[1]{\operatorname{dec}(#1)}
\newcommand{\hash}[1]{\operatorname{h}\left(#1\right)}
\newcommand{\redst}[1][115pt]{\bgroup\markoverwith {\textcolor{red}{\makebox[0pt][l]{\rule[0.5ex]{#1}{0.4pt}}\rule[1ex]{#1}{0.4pt}}}\ULon}

\newcommand{\ie}{\textit{i.e.,}\@\xspace}
\newcommand{\eg}{\textit{e.g.,}\@\xspace}

\newcommand{\pow}{PoL\xspace}

\newcommand{\re}{$\varepsilon_{repr}$\xspace}
\newcommand{\ret}{$\varepsilon_{repr}(t)$\xspace}%
\newcommand{\nre}{$||\varepsilon_{\text{repr}}||$\xspace}
\newcommand{\nret}{$||\varepsilon_{\text{repr}}(t)||$\xspace}
\newcommand{\rd}{$d_{\text{ref}}$\xspace} %

\newcommand{\pWeights}{\mathbb{W}}
\newcommand{\pBIndex}{\mathbb{I}}
\newcommand{\pSign}{\mathbb{H}}
\newcommand{\pMeta}{\mathbb{M}}

\algnewcommand\algorithmicoptional{\textbf{Optional:}}
\algnewcommand\Optional{\item[\algorithmicoptional]}%

\ifdraft
\newcommand{\nick}[1]{\textcolor{blue}{nick: #1}}
\newcommand{\varun}[1]{\textcolor{cyan}{VC: #1}}
\newcommand{\chris}[1]{\textcolor{teal}{CC: #1}}

\newcommand{\ilia}[1]{\textcolor{green}{ilia: #1}}
\newcommand{\anvinth}[1]{\textcolor{orange}{anvith: #1}}
\newcommand{\my}[1]{\textcolor{purple}{M.: #1}}
\newcommand{\natalie}[1]{{\color{red} [natalie: #1]}}
\newcommand{\stephan}[1]{{\color{violet} [stephan: #1]}}
\newcommand{\adam}[1]{{\color{orange} [adam: #1]}}
\newcommand{\vinith}[1]{{\color{brown} [vinith: #1]}}
\newcommand{\lucy}[1]{{\color{lightgray} [lucy: #1]}}
\else
\newcommand{\nick}[1]{}
\newcommand{\varun}[1]{}
\newcommand{\chris}[1]{}
\newcommand{\ilia}[1]{}
\newcommand{\anvinth}[1]{}
\newcommand{\my}[1]{}
\newcommand{\natalie}[1]{}
\newcommand{\stephan}[1]{}
\newcommand{\adam}[1]{}
\newcommand{\vinith}[1]{}
\newcommand{\lucy}[1]{}
\fi

\newcommand\blfootnote[1]{%
  \begingroup
  \renewcommand\thefootnote{}\footnote{#1}%
  \addtocounter{footnote}{-1}%
  \endgroup
}

\newtheorem{theorem}{Theorem}

\newtheorem{conjecture}{Conjecture}

\providecommand{\floor}[1]{ \lfloor #1 \rfloor }

\title{\vspace*{-0.5in}
{{\normalsize \rm In 42\textsuperscript{nd} IEEE Symposium on Security and Privacy\hrule}}
\vspace*{0.4in}Proof-of-Learning:  Definitions and Practice}

\author{Hengrui Jia\text{*}\IEEEauthorrefmark{4}, Mohammad Yaghini\text{*}\IEEEauthorrefmark{4}, Christopher A. Choquette-Choo\text{+}\IEEEauthorrefmark{4}, Natalie Dullerud\text{+}\IEEEauthorrefmark{4}, Anvith Thudi\text{+}\IEEEauthorrefmark{4}, Varun Chandrasekaran\IEEEauthorrefmark{2}, Nicolas Papernot\IEEEauthorrefmark{4}\vspace*{0.15cm} \\ 
University of Toronto and Vector Institute\IEEEauthorrefmark{4}, University of Wisconsin-Madison\IEEEauthorrefmark{2} 
}
\begin{document}

\maketitle

\begin{abstract}

Training machine learning (ML) models typically involves expensive iterative optimization. Once the model's final parameters are released, there is currently no mechanism for the entity which trained the model to prove that these parameters were indeed the result of this optimization procedure. Such a mechanism would support security of ML applications in several ways. For instance, it would simplify ownership resolution when multiple parties contest ownership of a specific model. It would also facilitate the distributed training across untrusted workers where Byzantine workers might otherwise mount a denial-of-service by returning incorrect model updates. 

In this paper, we remediate this problem by introducing the concept of proof-of-learning in ML. Inspired by research on both proof-of-work and verified computations, we observe how a seminal training algorithm, stochastic gradient descent, accumulates secret information due to its stochasticity. This produces a natural construction for a proof-of-learning which demonstrates that a party has expended the compute require to obtain a set of model parameters correctly. In particular, our analyses and experiments show that an adversary seeking to illegitimately manufacture a proof-of-learning needs to perform \textit{at least} as much work than is needed for gradient descent itself. 

We also instantiate a concrete proof-of-learning mechanism in both of the scenarios described above. In model ownership resolution, it protects the intellectual property of models released publicly. In distributed training, it preserves availability of the training procedure. Our empirical evaluation validates that our proof-of-learning mechanism is robust to variance induced by the hardware (\eg ML accelerators) and software stacks.
\end{abstract}
\blfootnote{\text{*}Joint lead authors; \text{+}joint secondary authors.}

\section{Introduction}
\label{sec:intro}

Training machine learning (ML) models is computationally and memory intensive~\cite{li__chuan_openais_2020}, often requiring hardware acceleration. GPUs~\cite{markidis2018nvidia}, TPUs~\cite{jouppi2017datacenter}, and FPGAs~\cite{putnam2014reconfigurable} are used to ensure efficient training. In the status quo, there is no way for an entity \textit{to prove that they have performed the work} required to train a model. This would be of immense utility in at least two settings. First, once a model is released publicly intentionally or unintentionally (\ie it is stolen), the model's owner may be interested in proving that they trained the model as a means to resolve and claim ownership---for instance, resolving claims related to model stealing attacks~\cite{tramer2016stealing, 2016arXiv160202697P, pal, copycat, knockoff}. Second, a model owner may seek to distribute the training~\cite{li2014scaling} across multiple workers (\eg virtual machines in a cloud) and requires guarantees of integrity of the computation performed by these workers. This would defend against some of the parties being corrupted accidentally (\eg due to hardware failure) or maliciously (\eg by an adversary which relies on Byzantine workers to perform denial-of-service attacks~\cite{blanchard2017machine}). 

In our work, we design a strategy that will allow a party--the \textit{prover}--to generate a proof that will allow another party--the \textit{verifier}--to verify the \textit{correctness} of the computation performed during training. In the case of ML, this translates to the prover generating a proof to support its claims that it has performed a specific set of computations required to obtain a set of model parameters. In the model stealing scenario, the prover would be the model owner, and the verifier would be a legal entity resolving ownership disputes. In the distributed learning scenario, the prover would be one of the workers, and the verifier the model owner. We name our strategy \textit{proof-of-learning (\prf)}. Unlike prior efforts related to proofs-of-work~\cite{dworknaor1992,preneel_proofs_1999}, our approach is not aimed at making computation expensive so as to inhibit denial-of-service attacks. 

When developing our concept for \prf, we consider only the training phase and not the inference phase; the cost of inference is generally much lower, and there already exist mechanisms to ensure the integrity of ML inference performed by another party~\cite{ghodsi_safetynets_nodate}. In our design, we wish to design a proof strategy that adds \textit{limited overhead} to the already computationally intensive process of training. Deep models do not have closed form solutions, and use variants of gradient descent as the de-facto choice for training. Additionally, stochastic gradient-based optimization methods used in deep learning, like stochastic gradient descent (SGD), update model parameters iteratively over long sequences by computing unbiased estimates of the true gradient~\cite{rumelhart1986learning}. Naturally, this sequence represents the work performed by the prover in training their model. We propose that \prf for ML should demonstrate two properties: (a) the prover performed the necessary optimization (expending computational resources) to train an ML model, and (b) these steps were computed \textit{correctly}, \ie that we have integrity of computation. 

There has been extensive research in proof systems related to other applications. Verified computations relates to settings where outcomes of outsourced computation (such as in client-server architectures) can be verified~\cite{setty2012making,setty2012taking,braun2013verifying,hawblitzel2015ironfleet,tan2017efficient}. Theoretical advances and efficient hardware design have enabled both smaller proofs and more efficient verification strategies~\cite{ishai2007efficient,walfish2015verifying}. The simplest scheme, however, involves \textit{duplicated execution} \ie re-executing the computation performed to verify the validity of the proof.

Following this intuition, we introduce in a general approach to obtain a \prf which enables verifying the computation performed during training (see \S~\ref{sec:pow}). We then instantiate a concrete \prf which utilizes the difficulty to invert gradient descent (see \S~\ref{sec:pow}). The added advantage here is that operations involving gradient descent are computed as part of the learning procedure, and can be used for generating the proof as well. In our work, the guarantees sought by the prover are analogous to those in the verifiable computations literature: given (i) a (random) distribution to draw the initial model weights from, (ii) the model's final weights, and (iii) a dataset, the prover must provide \textit{a sequence of batch indices and intermediate model updates} that, starting from the initialization, one can \textit{replicate} the path to the final model weights. This allows a verifier to recompute \textit{any} of the steps of gradient descent of their choosing to confirm the validity of the sequence provided. This in turn demonstrates that the prover has indeed performed the computation required to obtain the final parameter values. However, verification also requires the expensive process of gradient computation to verify the steps taken, as our proposal is based on re-execution. To make verification more computationally affordable, we introduce a heuristic for the verifier to select \textit{only a subset of the pairs of model parameter states} to verify. This allows the verifier to trade-off the confidence of verification with its cost: if the verifier \textit{randomly} picks a set of parameter pairs, then with sufficiently many choices, it can be confident of the proof's validity. 

There are many sequences that can be obtained from a given start state to a given final state (owing to the various sources of stochasticity involved in training). However, through our theoretical analysis in \S~\ref{sec:security_analysis}, we observe that obtaining these states through conventional training (\ie moving forward through the sequence) is more efficient than inverting gradient descent (\ie moving backwards through the sequence). Our analysis shows that inverting gradient descent takes \textit{at least} as much work as training. Thus, it is hard for an adversary to \textit{spoof} a \prf using such a strategy.

In summary, our contributions are the following: 
\begin{itemize}
    \item In \S~\ref{sec:formalizing}, we formalize the desiderata for a concept of proof-of-learning, the threat model we operate in, and introduce a formal protocol between the different actors involved in generating a \prf.
    \item In \S~\ref{sec:pow}, we introduce a general mechanism for \prf based on the observation that stochastic gradient descent utilized during training is difficult to invert.
    \item We analytically prove the correctness of our mechanism in \S~\ref{sec:correctness_analysis}, and then verify experimentally that it can be implemented despite hardware and software stochasticity. %
    \item We analyze the security of our proposed mechanism in \S~\ref{sec:security_analysis} through an analysis of entropy growth in gradient descent, and evaluate possible spoofing strategies an adversary may rely on to  pass verification.
    \item \textcolor{black}{Our code is open-sourced at \url{github.com/cleverhans-lab/Proof-of-Learning}. }
    
\end{itemize}
\section{Related Work}

\subsection{Proof-of-Work in Cryptography}

The concept of proof-of-work (or PoW), where one party proves to another that it has expended computational resources towards a computation result, was first introduced by Dwork and Naor~\cite{dworknaor1992}. The concept was motivated as a defense from denial-of-service (DoS) attacks against email and network providers. This was the main motivation for many later PoW functions as well, in which PoW functions force the adversary to expend significant computational resources, whether CPU or memory resources, in order to request access to the service. We revisit this motivation in \S~\ref{sec:formalizing}, but with the perspective of ML systems in mind. The term PoW itself was later introduced by Jakobsson and Juels~\cite{preneel_proofs_1999}. A key property of this formulation is that PoW relies largely on the existence of one-way functions popular in cryptography to establish an asymmetry between the party doing the computation and the party verifying that the computation was performed. 

In standard two-round PoW protocols, the prover receives a query including a cryptographic puzzle, frequently involving or indirectly based on a hashed randomly generated value or structure computed by the verifier. The prover solves the computational puzzle and returns the value, which the verifier either accepts as a solution to the problem or rejects. Generally, the process of solving the computational problem by the prover depends, directly or indirectly, on computation of a pre-image of a hashed random number generated and computed by the verifier, a known hard and expensive problem. 

Dwork and Naor~\cite{dworknaor1992} enumerated several PoW strategies predicated on integer square root module large prime problem: \eg the Fiat Shamir signature scheme and the Ong-Schnorr-Shamir signature scheme. Since then, many methods have been proposed for PoW functions. These initial PoW functions constituted CPU-bound functions and later memory-bound PoW functions gradually grew out of the field as well. Among PoW functions are partial hash inversion~\cite{preneel_proofs_1999}, moderately hard memory-bound functions~\cite{moderate_hard_mbound}, guided tour puzzle~\cite{guided_tour}, Diffie-Helman problem-based~\cite{diffiehellman}, Merkle-tree-based~\cite{merkle_tree_based_pow}, Hokkaido~\cite{hokkaido} and Cuckoo cycle~\cite{cuckoo_cycle}. 

Recently, systems that incorporate PoW have also been motivated by or used for various cryptocurrencies. Many current cryptocurrencies, such as Bitcoin and HashCash~\cite{hashcash, bitcoin}, employ systems based on PoW algorithms. Blockchain systems in cryptocurrency utilize a modified setup of the typical setting and actors in PoW frameworks for DoS attacks. In Bitcoin, miners competitively attempt to secure a payment as follows. First, they collect  unverified Bitcoin transactions from coin dealers in the Bitcoin network. Second, they combine these transactions with other data
to form a block which is only accepted once the miner has found a nonce number hashing to a number in the block with enough leading zeros.

\subsection{Security in ML Systems}

Most work on security in the context of ML~\cite{huang2011adversarial,papernot2018sok,biggio2018wild} has focused on the integrity of model predictions~\cite{szegedy2013intriguing,biggio2013evasion,biggio2012poisoning}
or on  providing guarantees of privacy to the training data~\cite{song2013stochastic}. 
Our efforts on developing a proof-of-learning (or \prf) concept for the training algorithm are instead, as illustrated by both of the use cases discussed in \S~\ref{sec:intro}, most relevant to two previous lines of work: the first is model stealing, the second is Byzantine-tolerant distributed learning. %

\paragraph{Model Ownership \& Extraction}\label{ssec:extraction} 
The intellectual property of model owners can be infringed upon by an adversary using model extraction attacks~\cite{tramer2016stealing}. Most extraction attacks targeting DNNs are learning-based: the adversary collects a substitute dataset (\ie consists of data from a similar distribution or synthetic data), queries the victim model to obtain labels, and then re-trains a surrogate model that is functionally similar to the victim~\cite{tramer2016stealing, 2016arXiv160202697P, pal, copycat, knockoff}. Attacks may also use side-channel information~\cite{sidechannel}. %

There are currently two types of defenses against extraction attacks: (a) restricting information released for each query answered by the model~\cite{tramer2016stealing, 2019arXiv190901838J}, and (b) assessing if a suspected model is indeed a stolen copy of a victim model. The latter can be done in two ways. If the model was  watermarked, one can query for the corresponding triggers~\cite{watermarksurvey, watermarkingdnn, watermarkingdnn2}. If that is not the case, one can use the training data directly to perform dataset inference~\cite{maini2021dataset}. However, decreasing extraction efficiency by restricting information returned by queries comes at the expense of the model's utility~\cite{tramer2016stealing,varun,lee2018defending,alabdulmohsin2014adding}. Similarly, watermarking trades utility with the robustness of the watermark while additionally requiring the modification of the training process~\cite{watermarksurvey, 2020arXiv200212200J, 2019arXiv190106151N, 2019arXiv191001226L, ewe}.
Thus, watermarks may be removed from a deployed model or made ineffective~\cite{finepruning, neuralcleanse, jia2020entangled}. In contrast, our work does not impact training and produces a \prf which is immutable (see \S~\ref{sec:formalizing}), we also do not restrict information released at inference time.

\subsection{Byzantine-tolerant distributed ML}

In the second scenario we described in \S~\ref{sec:intro}, we consider a setting where a model owner wishes to distribute the compute required to train a model across a pool of potentially untrusted workers~\cite{dean2012large}. Each of these workers receives a few batches of training data, performs some gradient descent steps, and then regularly synchronizes parameters with the model owner. In this distributed setting, we note that prior work has studied training algorithms which are robust to the presence of Byzantine~\cite{lamport2019byzantine} workers: such workers may behave arbitrarily and return corrupted model updates to the model owner~\cite{blanchard2017machine}. As we will introduce in \S~\ref{sec:formalizing}, verifiable \prf forms a defense against DoS attacks in this context. In addition, our \prf may be used to provide integrity guarantees by confirming the correctness of computations performed by the workers. %

\section{Background on Machine Learning}
\label{sec:background-on-machine-learning}

Throughout our work, we define $[n] := \{1, \dots, n\}$. Consider a data distribution $\mathcal{D}$ of the form $\mathcal{X} \times \mathcal{Y}$, such that $\mathcal{X}$ is the space of inputs, and $\mathcal{Y}$ is the space of outputs. An ML model is a parameterized function of the form $f_{W}: \mathcal{X} \rightarrow \mathcal{Y}$, where $W$ denotes the model parameters. For the purposes of this work, we assume that these models are deep learning models which requires additional terminology. 

\begin{enumerate}
\item \textit{Model Architecture:} A deep neural network is 
a function comprised of many layers, each performing a linear-transformation on their input with optional non-linear activations~\cite{glorot2011deep}. The structure of these layers, \eg the number of neurons and weights, the number of layers $l$, and activations is termed the model architecture. 

\item \textit{Model Weights:} The parameters of the deep learning model are commonly called its \textit{weights}. Each layer $i \in [l]$ is comprised of learnable weights denoted $w_i$, including the additional bias term. Collectively, we denote the set of per-layer weights $\{w_1, \cdots, w_l\}$ as $W$. %
\item \textit{Random Initialization:} Before training, each weight vector $w_i \in W$ requires an initial value. These values are often randomly assigned by sampling from a distribution. Values are sampled from a zero-centered uniform or Gaussian distribution whose standard deviation is parameterized by either the number of neurons in the input layer, the output layer, or both~\cite{glorot2010understanding,NIPS2017_5d44ee6f,He_2015_ICCV}. 

\end{enumerate}
The final set of parameters are learned by training the ML model using empirical risk minimization~\cite{vapnik2013nature}. A training dataset  is sampled from the data distribution $D_{tr} \sim \mathcal{D}$. The expected risk of a model on this dataset is then quantified using a loss: a real valued function $\mathcal{L}(f_{W}(x), y)$ that is the objective for minimization. The loss characterizes the discrepancy between the model's prediction $f_W(x)$ and the ground truth $y$. A common example is the cross-entropy loss~\cite{goodfellow2016deep}. 

Training occurs in an iterative manner by continuously sampling a (mini)\textit{batch} of training data, without replacement, from $D_{tr}$; each such iteration is called a \textit{step}\footnote{One step corresponds to processing one batch of data.}. 
For each step, stochastic gradient descent~\cite{robbins1951stochastic} updates the model's parameters to minimize the empirical risk by taking the gradient of the loss with respect to the parameters. Thus, at each step $i \in [T]$, we obtain a new set of weights $W_i$ as follows:
\begin{equation}
    W_i = W_{i-1} - \eta\cdot\nabla_{W_{i-1}}, \hat{\mathcal{L}}_{i-1}   
\label{eq:sgd}
\end{equation}
where $\eta$ is the learning rate hyperparameter, and $\hat{\mathcal{L}}_{i-1} = \frac{1}{m}\sum_{(x,y)\sim D_b}\mathcal{L}(f_{W_{i-1}}(x), y) $ denotes the average loss computed over a random batch $D_b \subseteq D_{tr}$ of size $m$. 
An \textit{epoch} is one full pass through $D_{tr}$ which contains $S$ steps. The training process overall has a total of $E$ epochs. Thus, assuming the size $m$ of a batch is fixed during training, training the model requires a total of $T = E \cdot S$ steps.

\section{Formalizing \prf}
\label{sec:formalizing}

We wish to show that \textit{one can verify the integrity of the training procedure used to obtain an ML model}. This in turn can also be used to show proof of \textit{ownership}. We focus on training because it induces the largest computational costs. We note that there is prior work in verifiable computing investigating inference-time computation but that these were not designed for training algorithms and require modifying the algorithm to accommodate cryptographic primitives such as an interactive proof system~\cite{zande, embedded_proofs, vCNN}. Instead, we formulate our approach such that no changes need to be made to the model architecture and training algorithm beyond additional logging. This enables a seamless integration for model owners to create \prf and make claims of having trained a model. 
Our approach for \prf is naturally extended to two scenarios:
\begin{enumerate}
\item A party can claim ownership of a trained model $f_{W_T}$. 
\item An entity outsources computation to some client (as in distributed learning), then the results returned by the client (\ie $f^{c}_{W_T}$) can be trusted.\footnote{The superscript $c$ denotes a computation executed locally at a client.}
\end{enumerate}

The party performing the computation is referred to as the prover $\prover$. To verify the integrity of its computation (either for ownership resolution or in the outsourced computation scenario), $\prover$ generates a certificate, henceforth referred to as the Proof-of-Learning (or \prf) performed to obtain $f_{W_T}$.\footnote{The case with $f^c_{W_T}$ is similar. For generality, we proceed to define our work with reference to $f_{W_T}$.} We denote such a \prf as $\pw[\prover, f_{W_T}]$. When the integrity of the computation (ergo model ownership) is under question, an honest and trusted verifier $\verifier$ analyzes $\pw[\prover, f_{W_T}]$ and determines its validity (\ie a valid \prf implies that $\prover$ performed the computation required to obtain $f_{W_T}$). Formally, a  \textit{valid} \prf is one where each component is well-formed (refer \S~\ref{definition}), and $\verifier$ can reconstruct the \prf in its \textit{entirety}. An adversary $\adv$ is one who wishes to subvert this process.

\subsection{Threat Model}
\label{threat}

\textit{Dishonest spoofing} is any strategy that requires lesser computational expenditure than that made by the prover in generating the proof; we formally define this term in \S~\ref{ssec:defining-a-spoof}. The primary scenario we wish to mitigate against is the ability of an adversary $\adv$ to \textit{efficiently} spoof $\pw[\prover, f_{W_T}]$, \ie we want to verify computation to train the model on the part of the prover. By spoofing, $\adv$ can claim to have performed the computation required (to produce $f_{W_T}$, for example). Since $\adv$ has not expended (significant) computational resources nor trained the model to be able to produce $f_{W_T}$, they are unlikely to have $\pw[\prover, f_{W_T}]$. Thus, $\adv$ tries to create $\pw[\adv, f_{W_T}]$ that passes verification, even if that \prf is not valid. We consider the following scenarios for spoofing:
\begin{enumerate}[label=(\alph*)]
\item \textit{Retraining-based Spoofing:} $\adv$ aims to create the \textit{exact same} \prf for $f_{W_T}$ as $\prover$ \ie $\pw[\adv, f_{W_T}] = \pw[\prover, f_{W_T}]$.

\item \textit{Stochastic Spoofing:} $\adv$ aims to create a \textit{valid} \prf for $f_{W_T}$, but this may not be the same as $\prover$'s \prf \ie $\pw[\adv, f_{W_T}] \neq \pw[\prover, f_{W_T}]$.

\item \textit{Structurally Correct Spoofing:} $\adv$ aims to create an \textit{invalid} \prf for $f_{W_T}$ but such a \prf passes verification for $f_{W_T}$. %

\item \textit{Distillation-based Spoofing:} $\adv$ aims to create a \textit{valid} \prf using a modified version of $f_{W_T}$ (say $f$) \ie $\pw[\adv, f] \neq \pw[\prover, f_{W_T}]$. Note that the adversarial approximation of the model $f (\approx f_{W_T}$) has the same test-time performance. %
\end{enumerate}

In our security analysis (see \S~\ref{sec:security_analysis}), we comment on the \textit{efficiency} of the above spoofing strategies; for the adversary, it is desirable that the aforementioned are dishonest spoofing strategies. We assume the following adversarial capabilities:

\begin{enumerate}
\item $\adv$ has full knowledge of the model architecture and parameters (\ie weights). In addition, $\adv$ has access to the loss function, optimizer, and other hyperparameters.
\item $\adv$ has full access to the training dataset, and can modify it. Note that the objective of $\adv$ is not to infer sensitive information from the dataset, but use it to spoof a \prf. %
\item $\adv$ does not have access to the various sources of randomness used by $\prover$. These sources include randomness associated with batching, parameter initialization, chosen random seeds, and other intrinsic sources of randomness such as hardware accelerators~\cite{DBLP:journals/corr/abs-1912-02919}. 
    
\end{enumerate}

\subsection{Protocol Overview}
\label{ssec:protocol}

We define \prf in ML as a $n\geq 1$-\textit{round} protocol between the prover $\prover$ and verifier $\verifier$. The protocol is initiated by $\prover$ by (a) drawing on some source of randomness, or (b) using some other parameters (with a valid \prf) for initialization of its model parameters ($W_0$); we will more formally define the latter in \S~\ref{ssec:practical-considerations}. $\prover$ then trains their ML model and obtains final parameters $W_T$. Through training, $\prover$ accumulates some \textit{secret information} associated with training; this information is used to construct $\pw[\prover, f_{W_T}]$ which can be used to prove integrity of the computation performed by $\prover$ to obtain $W_T$ from $W_0$. To validate the integrity of the computation, $\verifier$ may query $\prover$ for the \prf and $\prover$ returns a subset (or all of) the secret information obtained during training. Using this knowledge, $\verifier$ should be able to ascertain if the \prf is valid or not.

\vspace{1mm} %

\para{Desired Guarantees}
$\adv$ cannot (a) \textit{easily reconstruct} the secret information associated with $\pw[\prover, f_{W_T}]$ (needed for the retraining-based spoofing strategy), or (b) \textit {efficiently reconstruct} another valid \prf $\pw[\adv, f_{W_T}]$ or $\pw[\adv, f]$ for $f\approx f_{W_T}$. In particular, the computational resources needed should (ideally) be the same or more as the cost of valid proof generation. We formalize the computational requirements below:

\begin{enumerate}[label=\textit{Property} \arabic*.,itemindent=*]
  \item Let $C_\prover$ denote a random variable representing the cost (both computation and storage) associated with $\prover$ training $f_{W_T}$. Let $C_{\verifier}$ denote the cost random variable of the verification procedure. We thus require that \[\mathbb{E}[C_{\verifier}] \leq \mathbb{E}[C_{\prover}]\]
  \item Let $C_\adv$ be the cost random variable associated with any spoofing strategy attempted by any $\adv$ aside from the honest strategy (i.e. training $f_{W_T}$). We require that
  \[\mathbb{E}[C_\prover] \leq \mathbb{E}[C_{\adv}]\]
\end{enumerate}
Note here that the second property should hold no matter which of the four scenarios from \S\ref{threat} we consider: in particular the cost of the adversary should be higher even if they choose scenario (c) and form a structurally correct \prf which is invalid but still passes verification.

\subsection{Defining \prf}
\label{definition}

\begin{definition}[\prf]\label{definition:prf}
For a prover $\prover$, a valid \prf is defined as $\pw[\prover, f_{W_T}] = (\pWeights, \pBIndex, \pSign, \mathbb{A})$ where all the elements of the tuple are ordered sets indexed by the training step $t \in [T]$. In particular, (a) $\pWeights$ is a set of model specific information that is obtained during training, (b) $\pBIndex$ denotes information about the specific data points used to obtain each state in $\pWeights$, (c) $\pSign$ represents signatures of these training data points, and (d) $\mathbb{A}$ that incorporates auxiliary information that may or may not be available to $\adv$, such as hyperparameters $\pMeta$, model architecture, optimizer and loss choices. 
\end{definition}

The information in Definition~\ref{definition:prf} encapsulates all the information required to recreate (and consequently verify) a \prf. $\prover$ publishes some deterministic variant of $\pWeights$ (\eg encrypted $\pWeights$). Our scheme should ensure that recreating the states in $\pWeights$ without knowledge of $\pBIndex, \pSign$ and some designated subset of elements in $\mathbb{A}$ is hard; this should dissuade any adversary in recreating the prover $\prover$'s \prf. In addition to this, we should also ensure recreating $W_T$ without $\pWeights$ is hard so that the adversary cannot spoof (refer \S\ref{sec:security_analysis}) the \prf with a different \prf $\pw[\adv, f_{W_T}] \neq \pw[\prover, f_{W_T}]$. To this end, we require that algorithms included within $\mathbb{A}$ be from a known accepted list of algorithms, or have their own \prf (refer \S~\ref{sec:chain-of-trust}). More concretely,  any \prf (and the strategy to generate the \prf) should satisfy the following properties:

\begin{enumerate}
\label{sec:Gs}
\item[\textbf{G1.}] \textit{Correctness:} A \prf for $f_{W_T}$ should be verifiable with high probability if the prover $\prover$ obtained this \prf by training a model from a random initialization of the model parameters and until their convergence to $f_{W_T}$. %
\item[\textbf{G2.}] \textcolor{black}{\textit{Security:} If
$\adv$ is able to \textit{dishonestly spoof} the \prf, then it will be detected with high probability.}
\item[\textbf{G3.}] \textcolor{black}{\textit{Verification Efficiency:} Verifying the correctness of a proof should ideally be computationally less expensive than generating the proof. Additionally, verification should succeed even if the verifier uses different hardware than the prover.} 
\item[\textbf{G4.}] \textit{Model Agnostic:} A proof generation strategy should be general \ie should be applicable to models of varying nature and complexity.
\item[\textbf{G5.}] \textit{Limited Overhead:} Generating the proof should induce limited overhead to the already computationally expensive training procedure. 
\item[\textbf{G6.}] \textcolor{black}{\textit{Concise Proof:} The proof generated should be small with respect to the number of steps of training (and ideally of constant size).}
\end{enumerate}

\section{A \prf Mechanism based on Gradient Descent}
\label{sec:pow}
Our proposal for generating a \prf is based on gradient descent. At the core, our mechanism relies on the difficulty to invert gradient descent. In this section, we simplify the notation for brevity \ie $\pw[\prover, f_{W_T}]$ is now $\pw[f_{W_T}]$.

\subsection{Mechanism Overview}

In our proposed mechanism, $\prover$ reveals to $\verifier$ some of the intermediate weights achieved during training as its \prf $\pw[f_{W_T}]$. More specifically, $\prover$ releases: (a) the values of the weights (or model updates) at periodic intervals during training, and (b) the corresponding indices of the data points from the training set which were used to compute said model updates. To ensure that $\adv$ cannot copy the \prf as is, we require that $\prover$ encrypt their \prf $\pw[f_{W_T}]$ with $\verifier$'s public key $K_{\verifier}^{pub}$ to obtain $\encryptedPw:= \encrypt{\pw[f_{W_T}], K_{\verifier}^{pub}}$, and then sign it with $\prover$'s own private key before publishing the \prf.  
\textcolor{black}{ The proof (or its signature) can be timestamped, or published in a public ledger. This ensures that verifying its validity is as simple as a lookup operation. This prevents replay attacks, where $\adv$ would claim to have published the \prf first.}

To commence verification, $\verifier$ first verifies the authenticity of the signature using $\prover$'s public key and proceeds to decrypt the encrypted \prf using its private key $K_{\verifier}^{priv}$. It then verifies the provenance of the initial weights $W_0$. These are either (a) sampled from the claimed initialization distribution, or (b) come from a valid external source, \ie have their own \prf. See \S~\ref{sec:random-intialization} and \S~\ref{sec:chain-of-trust}, respectively. 
Next, $\verifier$ queries $\prover$ for the data points required to compute a \text{specific subset} of updates in $\mathbb{W}$. There are two possibilities. Either the dataset is released by $\prover$ along with the \prf and is available to $\verifier$ immediately. Alternatively, in a \textit{lazy verification} scenario
(\S~\ref{sec:lazy-verification})
, $\prover$ can delay the release of the exact data points to $\verifier$ until they are explicitly queried. In such a case, $\prover$ is necessitated to include a signature (represented using function $h(.)$) of the training data as part of the \prf. We require this as an abundance of precaution so that an adversarial prover attempting structurally correct spoofing (see \S~\ref{threat}) cannot release a structurally correct yet invalid \prf and then later attempt to synthesize a dataset which would make this \prf valid. %

The process of obtaining updates in $\mathbb{W}$ is similar to training when aided by the information contained in $\mathbb{I}$, $\mathbb{H}$ and $\mathbb{A}$. In our protocol we only retain hyperparameters $\pMeta$ as our auxiliary information in $\mathbb{A}$. Thus, by querying this information, $\verifier$ can recreate the updates in a specific subset by re-executing the computation. By doing so, $\verifier$ is able to attest the computation performed by $\prover$. We detail this verification in \S~\ref{ssec:verification}.

\subsection{\prf Creation}
\label{ssec:prf-prover}

\begin{algorithm}
\caption{\prf Creation}\label{alg:creation}
\begin{algorithmic}[1]
\Require Dataset $\dataset$, Training metadata $M$
\Require $\verifier$'s public key $K^{pub}_{\mathcal{V}}$
\Require $E, S, k$
\Comment{{\footnotesize Number of epochs, steps per epoch, checkpointing interval}}
\Optional $W_0, \zeta$ \Comment{{\footnotesize Initialization weight and strategy}}
\State $\pWeights\gets\{\}, \pBIndex\gets\{\}, \pSign\gets\{\}, \pMeta\gets\{\}$
\If{$W_0 = \emptyset$}
\State $M_0 \gets \zeta$
\State $W_{0} \gets \texttt{init}(\zeta)$
\EndIf
\For{$e \gets 0,\dots,E-1$} \Comment{{ \footnotesize Training epochs}}
\State $I \gets {\texttt{getBatches}(D, S)}$
\For{$s \gets 0,\dots,S-1$} \Comment{{ \footnotesize steps per epoch}}
\State ${t = e\cdot S + s}$ \label{alg:line:t-calc}
\State $W_{t+1}\gets \texttt{update}(W_{t}, \dataset[I_s], M_{t})$
\State $\pBIndex.\texttt{append}(I_t)$
\State $\pSign.\texttt{append}(\hash{\dataset[I_t]})$
\State $\pMeta.\texttt{append}(M_{t})$
\If{$t \bmod k = 0$}
\State $\pWeights.\texttt{append}(W_{t})$
\Else
\State $\pWeights.\texttt{append}(\textbf{nil})$
\EndIf
\EndFor
\EndFor
\State $\mathbb{A}\gets\{\pMeta\}$
\State $\encryptedPw \gets \encrypt{(\pWeights, \pBIndex, \pSign, \mathbb{A}), K_{\verifier}^{pub}}$
\vspace{2mm}
\State \Return $\encryptedPw, \hash{\encryptedPw, K_\prover^{priv}}$
\end{algorithmic}
\end{algorithm}

In Algorithm~\ref{alg:creation}, we present the concrete mechanism to create \prf $\pw[f_{W_T}]$. $\pWeights$ is a flattened list of all recorded weights across all epochs indexed by the proof step $t$. The mapping from training step $s$ to the proof step $t$ is $t = e\cdot S + s$, where $S$ is the number of training steps per epoch and $e$ is the epoch counter (of a total of $E$ epochs). We only append a weight $W_{t}$ every $k^{th}$ step of the training, and otherwise add $\perp$ at that index. \textcolor{black}{Observe that checkpointing is commonly performed as part of training and adds limited overhead (\hyperref[sec:Gs]{\textbf{G5}})}. $k$ is a parameter which we call checkpointing interval; $\frac{1}{k}$ is then the \textit{checkpointing frequency}. Increasing $k$ helps optimize storage costs (refer \S~\ref{sssec:storage}). $\prover$ may use additional hyperparameters and optimizer specifications (\eg learning rate schedules, etc.), which we denote as metadata $M_{t}$ (to be included in $\pMeta$). To make sure that weights in $\pWeights$ will be verified on the same data samples $f_{W_T}$ was originally trained on, we require that $\pw[f_{W_T}]$ include a signature of the training data, \ie $\hash{\dataset[I_t]}$ in $\pSign$ along with the data indices which are themselves included in $\pBIndex$.

In Algorithm~\ref{alg:creation}, \texttt{init()} is a method that initializes the weights according to an initialization strategy $\zeta$ before training commencement. In scenarios where the initial model state is obtained from elsewhere, we require that a \prf be provided for the initial model state itself as well (see \S~\ref{sec:chain-of-trust}). 
In a similar vein, \texttt{getBatches()}  randomly assigns a set of data indices (over the entire dataset) to each batch. Thus, the output of the method is a list of $T$ sets of indices. Finally, the method \texttt{update()} performs an update to the model parameters using a suitable optimizer implementing one of the variants of gradient descent as in Equation~\eqref{eq:sgd}.

\para{Storage Cost}
The \textit{proof size} is $\frac{ES}{k}|W|$ where $|W|$ indicates the size of a set of model weights \ie a single checkpoint (\hyperref[sec:Gs]{\textbf{G6}}). We note that if the prover would like to delay the verification until requested (see \S\ref{sec:lazy-verification}) then they should maintain a copy of the dataset, which adds $|D|$ to the storage cost, where $|D|$ is the size of the dataset. 
Increasing the checkpointing interval linearly decreases the storage cost, however this can come at the cost of verification accuracy (see \S\ref{ssec:verification}). Storage costs are discussed in detail in \S~\ref{sec:checkpoint_storage_cost_analysis}.

\subsection{\prf Verification}
\label{ssec:verification}
Algorithm~\ref{alg:verification} summarizes the verification algorithm. Every \prf starts from a set of weights either sampled from the claimed initialization distribution, or from previously trained model weights. In the latter case, the prover needs to provide a valid \prf for the pre-trained model weights, \ie $\pw^{0}$ (referenced in encrypted form in Algorithm \ref{alg:verification} as $\encryptedPw^0$). In the case of sampling from the claimed initialization distribution, a statistical test is conducted to verify the claim. We discuss these requirements and their importance in more detail in \S~\ref{sec:chain-of-trust} and \S~\ref{sec:random-intialization}, respectively.
After this initial verification step, we store the distance between each consecutive pair of weights captured in $\pWeights$ in a new list $mag$ using $d_1$ which is a distance measure in a metric space (such as the $p$-norm). Once every epoch, we sort $mag$ to find the largest model updates which we verify using the \textproc{VerifyEpoch} procedure.
To verify, $\verifier$ loads up the index corresponding to the largest model update into its own model $W'_t$. Next, $\verifier$ performs a series of $k$ updates to arrive at $W'_{t+k}$ which is compared to the purported $W_{t+k}$ in the \prf. We tolerate $d_2(W'_{t+k},W_{t+k}) \leq \delta$, where $d_2$ is a distance measure (possibly different from $d_1$). $\delta$ is a slack parameter that should be calibrated before verification starts, as it depends on hardware, model architecture, dataset, checkpointing interval, and the learning hyperparameters. Alternatively, acceptable ranges for $\delta$ may be mandated by law and enforced by the verifier. \textcolor{black}{Since the purpose of $\delta$ is to upper bound the randomness in training, one heuristic is to set $\delta$ as the average of a few gradient updates during training.} \textcolor{black}{We note that for an honest $\prover$ who has obtained all intermediate model weights, the particular choice of $k$ is immaterial. Also since $\delta$ is  chosen to account for hardware and software tolerances, Algorithm \ref{alg:verification} will correctly verify such an honest proof~(\hyperref[sec:Gs]{\textbf{G1}}).
}

\noindent{\bf Why only verify the largest updates?}
We verify the largest model updates because \textit{valid} updates tend to have small magnitude (to avoid overshooting during gradient descent), and we want to save computational cost of $\verifier$. More importantly, any estimation error introduced by an adversary $\adv$ wishing to recreate a proof at a smaller computational cost would be easier to detect for these large model updates. This may be because the adversary tried to spoof a valid \prf by fine-tuning models at large learning rates for few epochs, or because they attempt to spoof a \prf with significant discontinuity to arrive at a new $\widehat{W}$ (see \S~\ref{ssec:pow-concatenation}), 
We assume that the verifier $\verifier$ can verify at most $Q \cdot E$ largest updates (\ie Q per epoch), which we denote as $\verifier$'s \textit{verification budget}. Similar to the slack parameter $\delta$, $Q$ is also a verification hyper-parameter which should be calibrated, and can be mandated by law.

\para{Time Complexity}
The complexity of verification is $\operatorname{\mathcal{O}}(E \cdot Q \cdot k \cdot C_{|W|})$ where $C_{|W|}$ is the time-complexity of one update step of the training loop with parameter size $|W|$ (\hyperref[sec:Gs]{\textbf{G3}}). \textcolor{black}{Figure~\ref{fig:time_analysis} in Appendix~\ref{app:additional} shows a visualization of the bound.}

\begin{algorithm}[t]
	\caption{Verifying a \prf}
	\label{alg:verification}
	\begin{algorithmic}[1]
		\Function{Verify}{$\encryptedPw, \encryptedPw^{0}, K_{\verifier}^{priv}, \model, \dataset, Q, \delta$} \Comment{{\footnotesize encrypted {\prf}s, $\verifier$'s private key, model, dataset, query budget, slack parameter}}
			\State $\pWeights, \pBIndex, \pSign, \pMeta \gets \decrypt{\encryptedPw, K_{\verifier}^{priv}}$
			
			\If{$\encryptedPw^{0} = \emptyset$} \label{alg:line:init-verification-start}
				\If{$\Call{VerifyInitialization}{\pWeights_0} = $ FAIL}
					\State \Return{FAIL}
				\EndIf
			\ElsIf{$\Call{VerifyInitProof}{\encryptedPw^{0}}=$ FAIL}\label{alg:line:verify-init-proof}
				\State \Return {FAIL}
			\EndIf \label{alg:line:init-verification-end}
			\State $e \gets 0$ \Comment{{\footnotesize Epoch counter}}
			\State $mag \gets \{\}$ \Comment{{\footnotesize List of model update magnitudes}}
			\For{$t \gets 0, \dots, T-1$} \Comment{{\footnotesize training step}}
				\If{$t \bmod k = 0 \wedge t \neq 0$}
					\State $mag.\texttt{append}(d_1(\pWeights_t - \pWeights_{t-k}))$ 
				\EndIf
				
				\State $e_t = \floor{\frac{t}{S}}$ \Comment{{\footnotesize Recovering the epoch number}}
				\If{$e_t = e + 1$} 
					\State \Comment{{\footnotesize New epoch started. Verify the last epoch}}
					\State $idx \gets \operatorname{\texttt{sortedIndices}}(mag, \downarrow)$
					\\\Comment{{\footnotesize get indices for decreasing order of magnitude}}
					\If{$\Call{VerifyEpoch}{idx}$ = FAIL}
						\State	\Return FAIL
					\EndIf
					\State $e\gets e_t, mag \gets \{\}$
				\EndIf
			\EndFor

			\State \Return Success
			\vspace{1mm}
			\Function{VerifyEpoch}{$idx$}
				\For{$q \gets 1,\dots,Q$}
					\State $t = idx[q-1]$ \Comment{{\footnotesize index of $q$'th largest update}}
					\State $H_t \gets \pSign_t, I_t \gets \pBIndex_t$
					\vspace{1mm}
					\State $\Call{VerifyDataSignature}{H_t, \dataset[I_t]}$
					\vspace{1mm}
					\vspace{0.5mm}
					\State $W'_t \gets \pWeights_t$
					\For{$i \gets 0,\dots,k-1$}
						\State $I_{t+i} \gets \pBIndex_{t+i}, M_{t+i} \gets \pMeta_{t+i}$ 
						\State $W'_{t+i+1}\hspace{-1mm}\gets\hspace{-0.5mm}\texttt{update}(W'_{t+i}, \dataset[I_{t+i}], M_{t+i})$
					\EndFor
					
					\vspace{1mm}
						\State $W_{t+k}\gets \pWeights_{t+k}$
						\If{$d_2(W'_{t+k}, W_{t+k}) > \delta$} \Comment{{\footnotesize Dist. func. $d_2$}}
						\vspace{1mm}
						\State \Return {FAIL}
					\EndIf
				\EndFor
				\State \Return{Success}
			\EndFunction
		\EndFunction
	\end{algorithmic}
\end{algorithm}

\para{Verification Success}
We define the verification success rate (VSR) of verifier $\verifier$ on a \prf $\pw $ as:
\begin{equation}
	\label{eq:vsr}
    \vsr{\verifier}{\pw[f_{W_T}]} := \Pr[\textproc{Verify}[\pw[f_{W_T}]] = \text{Success}],
\end{equation}
where $\pw[f_{W_T}] = \pw[\prover, f_{W_T}] = (\pWeights, \pBIndex, \pSign, \mathbb{A})$ and \textproc{Verify} is a simplified notation for the same function in Algorithm \ref{alg:verification}.
Verification success can be described as the probability that the verifier accepts a \prf. Note that by nature of the verification Algorithm~\ref{alg:verification}, the probability of acceptance by the verifier depends on the probability of:
\begin{enumerate}
\itemsep0em
\item $W'_{t+k}$ (\ie calculated update from $W_t$ by the verifier) achieved within a $\delta$-ball of the purported weights $W_{t+k}$.
\item $\verifier$ obtaining $W'_{t+k}$ from initial weights $W_t$ in $k$ steps.
\end{enumerate}
Here, $t=idx[q-1]$, denotes the step with the $q^{th}$ largest $k$-step update in the given epoch $e$. As the update for each $t$ in the verification procedure is calculated separately and the value of $W_t$ is directly obtained, these events for different values of $t$ are independent. To ease the notation assume that $\pBIndex$ is reindexed so that $j$ is the index corresponding to the $q^{th}$ largest update. We re-write Equation~\eqref{eq:vsr} with `Success' as 1, and $\phi = (\pBIndex, \pSign, \pMeta)$,
\begin{multline*}
\Pr[\textproc{Verify}[\pWeights, \phi]=1] = \\ 
\prod_{e=1}^{E}\prod_{q=1}^{Q}\Pr[Tr_{e, q, k} \;\wedge\; 
dist_{e, q+k} \leq \delta \mid \phi] \\
= \prod_{e=1}^{E}\prod_{q=1}^{Q}\Pr[dist_{e, q+k} \leq \delta \mid \phi] \cdot \Pr[Tr_{e, q, k} \mid \phi],
\end{multline*}
where (a) $dist_{e, (q)+k} = d_2(W'_{e,(q)+k}, W_{e,(q)+k})$ denotes the distance measurement, and (b) $Tr_{e, (q), i} := W_{e,(q)} \rightrightarrows W'_{e,(q)+i}$ indicates the updates calculated by $\verifier$ of the $i\leq k$ steps in the $q^{th}$ largest $k$-step update in epoch $e$ and has achieved $W'_{e,(q)+i}$. We also have used the fact that the distance between the purported and the calculated updates is independent from $W_{e,(q)} \rightrightarrows W'_{e,(q)+k}$. Additionally, due to the Markovian nature of the gradient descent process (see \S~\ref{sec:correctness_analysis}), the updates $W'_{e, (q)+i}\rightrightarrows W'_{e,(q)+i+1}$ are independent of each other. 
Combining the last two factors, we have:
\begin{multline}
\vsr{\verifier}{\pw[f_{W_T}]} = \Pr[\textproc{Verify}[\pWeights, \phi]=1] \\ =  \prod_{e=1}^{E}\prod_{q=1}^{Q}\Pr[dist_{e, (q)+k} \leq \delta \mid \phi] \cdot \Pr[Tr_{e, (q), k} \mid \phi] \\
= \prod_{e=1}^{E}\prod_{q=1}^{Q}\Pr[dist_{e, (q)+k} \leq \delta \mid \phi] \prod_{i=0}^{k-1} \Pr[Tr_{e, (q)+i, 1} \mid \phi]
\end{multline}
Note that in above $W'_{e, (q)+0} = W_{e, (q)}$, as $\verifier$ is given these weights, so no noise is introduced by reproduced computation. 
We observe that decreasing the checkpointing interval
or 
increasing
the query budget per epoch $Q$ adds to the probability terms in the product, therefore, if there is any uncertainty regarding intermediate updates, their effect is compounded, which in turn makes for a more stringent verification process. This comes at a trade-off with storage cost (see \S\ref{ssec:prf-prover}).

\subsection{Practical Considerations}
\label{ssec:practical-considerations}

Here, we discuss practical considerations to be made when implementing the mechanism we described so far.

\subsubsection{Private Datasets \& Lazy Verification} \label{sec:lazy-verification}

So far we have assumed that the dataset used to train a model is public, so that $\verifier$ can use batch indices $\pBIndex$ in $\pw[f_{W_T}]$ to verify model updates. It is also possible to use our \prf scheme for private datasets. To do so, in addition to $\pw[f_{W_T}]$, $\prover$ needs to publish a signature of their datapoints $\hash{\dataset[I_t]}, I_t \in \pBIndex$ but not the dataset. In this setup, the verification can be delayed until necessary (\ie \textit{lazy verification}), at which time $\prover$ should reveal $\dataset[I_t]$ to $\verifier$ who additionally has to verify their signatures with the published record.

\subsubsection{Amount of Data Needed}
\label{sssec:storage}

With lazy verification, the expected amount of data required to be transferred to the verifier $\verifier$ can be expressed as a function of $S$, $E$, $k$, and $Q$ (for simplicity here we assume $Q$ is the same for all epochs). Let $c_{i}$ denotes the Binomial random variable representing the number of times data points $i$ is sampled by \textproc{VerifyEpoch} in Algorithm~\ref{alg:verification} (thus there are $E$ trials). We assume each data point is equally likely to be chosen such that in a certain trial
$\forall i \in [|D|], \Pr(c_{i}=1) = \frac{Q\cdot k}{S}$. Therefore, the probability of a data point being chosen at least once is 
\begin{equation}
\Pr(c_{i} \geqslant 1) = 1 - \Pr(c_{i} = 0) = 1 - (1 - \frac{Q\cdot k}{S})^{E}
\end{equation}
This means for dataset $D$, the expected amount of data for Algorithm~\ref{alg:verification} is $|D|[1 - (1 - \frac{Q\cdot k}{S})^{E}]$.

\subsubsection{Initial State Provenance and Chain of Trust}
\label{sec:chain-of-trust}
To improve convergence behavior and achieve better performance, most ML models are not initialized from a cold start---an initialization sampled randomly from a particular distribution. Indeed, it is common to start training from a set of weights that have previously achieved good results on a different dataset, and improve upon them (a warm start). A common example of a warm start is transfer learning~\cite{yosinski2014transferable}. If we do not check the provenance of the initial state, we discuss in \S~\ref{sec:security_analysis} how an adversarial prover could fine-tune a stolen model by continuing to train it for a few steps, thus creating a valid \prf, and claim that they have started from a \textit{lucky} initialization---where the lucky initialization is the true owner's final weights. %

In order to establish the \prf in models with a warm start while defending against the said attack scenario, we propose to establish a \textit{chain of trust}: a \prf $\pw[f_{W_T}]$ should come with a previously published $\pw^{0}$, where $\pw^{0}$ denotes the proof needed to verify the initial state $W_0$ used to obtain $\pw[f_{W_T}]$. The verifier keeps a record of previously verified proofs.
Therefore, in Algorithm \ref{alg:verification}, if $\verifier$ has recorded $\pw^{0}$, \textproc{VerifyInitProof} would be a simple record lookup. Otherwise, it would trigger the verification of $\pw^{0}$. The verification success rate follows a chain rule $\vsr{\verifier}{\pw[f_{W_T}] \rightarrow \pw^{0}} = \vsr{\verifier}{\pw[f_{W_T}]} \vsr{\verifier}{\pw^{0}}$, where $\rightarrow$ denotes the dependence. Of course $\pw^{0}$ can depend on a prior \prf $\pw^1$, and so on. 
Concretely, for all $j \geq 0$, let $\pw^{j+1}$ denote the \prf for the first set of model weight needed to obtain $\pw^{j}$, and $\pw^0 = \pw[f_{W_0}]$, \ie $\pw^0$ is the proof for $W_0$. Therefore, the VSR for a chain of $R$ prior \prf{}s can be written as:
\begin{multline}
     \vsr{\verifier}{\pw[f_{W_T}] \rightarrow \pw^{0} \rightarrow \cdots \rightarrow \pw^{R}} \\= \vsr{\verifier}{\pw[f_{W_T}]} \prod_{j=0}^{R}\vsr{\verifier}{\pw^{j}}.
\end{multline}

\textcolor{black}{
If $\prover$ cannot provide such a proof, then it must be the case that they have trained a model starting from a random initial state. In this case, $\prover$ should provide their initialization distribution and strategy and} apply a statistical test to verify that the initial model parameter values contained within the proof sequence were indeed sampled from the claimed distribution.

\subsubsection{Verifying Initialization}
\label{sec:random-intialization}
Most existing initialization strategies for model weights such as Xavier~\cite{glorot2010understanding}, Kaiming~\cite{He_2015_ICCV}, and Orthogonal Initialization~\cite{orthogonal}, involve sampling values from a designated distribution (usually normal or uniform). Such distributions rely on the architecture of the model (\eg dimensionality of a certain layer), so it can be easily obtained given the initialization strategy which must be included in the initial metadata $M_0 \in \pMeta$.

The Kolmogorov–Smirnov (KS) test~\cite{kstest} is a statistical test to check whether samples come from a specific distribution. We use a single-trial KS test to check if the weights of each layer are sampled from the designed distribution. If any layer does not pass the KS test, \ie the p-value is below the chosen significance level,
the verifier can claim that the initialization parameters are not sampled from the prover's claimed initialization distribution, making the \prf invalid. We note that the tests are done under the assumption that the different layers are initialized independently which is often the case~\cite{borisrolnick}. Otherwise, the significance level should be corrected to account for multiple testing using a method such as Bonferroni's method. Along with all other metadata (\eg the optimizer), we assume that $\prover$ and $\adv$ must choose an initialization strategy from a previously chosen (and publicly known) set of strategies (\eg all widely-known strategies), preventing the adversary from creating an arbitrary initialization strategy for their own spoofing purposes. In Algorithm \ref{alg:verification}, \textproc{VerifyInitialization} handles the initialization test.

\section{Correctness Analysis of the Gradient Descent Mechanism for Proof-of-Learning}
\label{sec:correctness_analysis}

Recall that the goal of our proposed verification scheme is for the verifier to gain confidence that each of the steps recorded in the \prf are valid, rather than verifying the end-to-end sequence altogether. We now prove why the verification must be performed step-wise.

\subsection{Stationary Markov Process}
\label{subsec:markov}
Training a neural network using a gradient-based stochastic optimization method is a Markov process, \ie its future progression is independent of its history given its current state. We formalize this property in Appendix~\ref{app:markov}. The Markov assumption is used in ML libraries, including  \texttt{pytorch}~\cite{paszke2019pytorch} and \texttt{tensorflow}~\cite{abadi2016tensorflow}, to enable in-place model updates. 

Gradient-based stochastic optimization method is not only a Markov process but also stationary, assuming that any randomness in the architecture is fixed (\eg using a fixed batching strategy, and with deterministic dropout masks). Without loss of generality, we prove this property for SGD but note that other gradient-based stochastic optimization methods follow (\hyperref[sec:Gs]{\textbf{G4}}). Here, we adopt the notation $\tilde{W_t} :=  (W_t, M_t)$ to denote the model weight and the associated learning hyperparameters at step $t$. Thus, a training step is represented as follows:
\begin{equation}
\tilde{W}_{t+1} = \tilde{W}_{t} - \eta \nabla_{\tilde{W}_{t}} \hat{\mathcal{L}}_{t} + z_{t},
\label{eq:trainin_process}
\end{equation}
where $z_{t}$ is the random variable representing noise arising from the hardware and low-level libraries such as cuDNN~\cite{chetlur2014cudnn} at step $t$ and the set of random variables $\{z_{t} \mid t\in [T]\}$ are independent and identically distributed. Thus, for all steps $t$ and arbitrary $\tilde{w}_{a}, \tilde{w}_{b}$,
\begin{equation*}
    \Pr(\tilde{W}_{t+1}=\tilde{w}_{a}|\tilde{W}_{t}=\tilde{w}_{b}) = \Pr(\tilde{W}_{t}=\tilde{w}_{a}|\tilde{W}_{t-1}=\tilde{w}_{b}).
\end{equation*}
Thus, the  process of training a neural network using gradient-based stochastic optimization is a stationary Markov process.

\subsection{Entropy Growth}
\label{subsec:entropy rate}
Building on our results in \S~\ref{subsec:markov}, we  analyze the entropy growth of training a DNN as a stationary Markov process,
$\Theta_{T}={\tilde{W}_{0}, \cdots, \tilde{W}_{T}}$. Entropy captures the variance, or number of possible paths of the gradient descent sequences~\cite{wiley_informationtheory}. Using the definition of entropy rate (refer Equation~\ref{eq:entropy_rate} Appendix~\ref{app:markov}) and Markovian nature of the training process $\Theta_{T}$, we get the entropy rate as follows: 
\begin{align}
    H'(\Theta_{T}) = \lim_{T\to\infty} H(\tilde{W}_{T}|\tilde{W}_{0},...,\tilde{W}_{T-1}) &= H(\tilde{W}_{1}|\tilde{W}_{0})\label{eq:entropy-markov}\\
         &= H(z_{0})\label{eq:entropy-step}
\end{align}
where we obtain Equation~(\ref{eq:entropy-step}) by plugging in the result stated in Equation~(\ref{eq:trainin_process}). This proves the following result: %
\begin{theorem}[Entropy Growth]
The entropy of the training process $\Theta_{T}$ grows {\em linearly} in the number of training steps ${T}$.
\label{theorem:-entropy-growth}
\end{theorem}

To bound the entropy, our verification scheme performs a step-wise comparison. Otherwise, the entropy would grow unbounded, increasing the difficult of accurately verifying the updates in a training process. Further, Theorem~\ref{theorem:-entropy-growth} also proves that the exact reproducibility of a ML model is difficult because the entropy grows, without bound, as the training sequence grows (rendering retraining-based spoofing impossible). This result holds true even with an identical initialization and batching strategy. Note that our only assumption was the presence of some i.i.d noise in the training process arising due to hardware and low-level libraries. Our result is therefore of interest beyond the setting considered in our work, and in particular explains the negative results observed previously in model extraction research when trying to reproduce a training run exactly~\cite{jagielski2020high}.\\

\para{Interpretation of Entropy Growth}\label{ssec:interpretation-of-entropy} Recall the definition of entropy~\cite{wiley_informationtheory}. The entropy of a training step captures the variance, or number of possible paths from that state (\ie how much information is needed to describe the possibilities). Thus, the expected variance of the sequences grows too. The relation between entropy and number of possible sequences is predominantly exponential as its definition is logarithmic to the probability. Thus the linear growth in entropy in Theorem~\ref{theorem:-entropy-growth} represents an exponential growth in the number of potential sequences of gradient descent.

\subsection{Reproducibility Evaluation}
\label{subsec:reproduce}

To illustrate our analysis, we empirically evaluate our verification scheme and the implications of Theorem~\ref{theorem:-entropy-growth}. We also discussed how to configure hyperparameters of Algorithms~\ref{alg:creation} and~\ref{alg:verification} to analyze trade-offs between storage cost and correctness of \prf verification.

\subsubsection{Experimental Setup}
\label{subsubsec:setup}
A Residual Neural Network (ResNet)~\cite{resnet} is a common deep neural network architecture used for image classification. We evaluated the proposed \prf for ResNet-20 and ResNet-50 on two object classification tasks: CIFAR-10 and CIFAR-100~\cite{cifar} respectively.  Each of the two datasets is composed of 50,000 training images and 10,000 testing images, each of size $32\times32\times3$. The datasets differ in that CIFAR-10 only has 10 classes whereas CIFAR-100 has 100 classes. Thus classifying CIFAR-100 is considered as a harder task. Both models are trained for 200 epochs with batch size being 128 (\ie $E=200$, $S=390$).

\subsubsection{Metrics For Evaluation}
\label{ssec:evalulation-metrics} 

Our goal here is to understand how the entropy growth of training (see Theorem~\ref{theorem:-entropy-growth}) impacts our capability to verify a training update. Formally, we are given (initial) weights $W_t$ which are trained to a state $W_{t+k}$, where $k$ represents some previously chosen and fixed checkpointing interval. The verifier then attempts to reproduce this step by calculating their own $W'_{t+k}$ from $W_t$. The reproduction error here is \ret$=d(W_{t+k}, W'_{t+k})$, using some distance metric $d$, \eg a $p$-norm. With a sufficiently small \ret, $\forall t\in[T]$, a verifier can confirm that indeed $W'_{t+k}\approx W_{t+k}$, $\forall t \in [T]$, which proves that the prover trained this ML model. Specifically, we require that $\max_t $ \ret $ \ll$ \rd, where \rd$=d(W^1_{T}, W^2_{T})$ 
is the reference distance between two models $W^1_{T}$ and $W^{2}_{T}$ of the same architecture, trained to completion (\ie for $T$ steps) using the same architecture, dataset, and initialization strategy, but with a \textit{different batching strategy and not forcing the same initial parameters} (\ie $W^1_{0}\neq W^2_{0}$). If this is the case, then we can set our distance threshold $\delta$ (refer to Algorithm~\ref{alg:verification}) such that $\max_t($\ret$) < \delta <$ \rd.
Note that \rd can be interpreted as the difference between two models trained from scratch by two independent parties, so it is used as our upper bound (\ie if two models differ by about \rd then they should not be considered as related).

Observing Table~\ref{tab:no_checkpoint}, we see that our empirical results corroborate Theorem~\ref{theorem:-entropy-growth}. Reproducing weights trained step by step ($k=1$) leads to a negligible \ret. However, attempting to reproduce an entire sequence leads to a large error due to the linear increase in entropy over the $T$ steps. Note that this error accumulates even when \textit{using the exact same batching strategy, architecture, initial parameters, and training setup}, due to the irreproducible noise $z$ arising from the hardware and low-level libraries. Thus, it is impossible for a verifier to reproduce an entire training sequence and we require that $k$ be sufficiently small to prevent these errors from accumulating and approaching to \rd. Note that our results display a normalized \nret $= \frac{max_{T}(\text{\re})}{\text{\rd}}$ where we require that \nret$ << 1$ for the sufficient condition to hold so that the verifier can select a suitable $\delta$.
\begin{table}[t]
    \centering
    \subfloat[CIFAR-10]{
    \begin{tabular}{c c | c c c}
        & & \multicolumn{2}{c}{Checkpointing Interval, $k$} & Deterministic\\
        &  & $k=E\cdot S$ & $k=1$ & operations\\
        \toprule
        \multirow{4}{*}{\rotatebox[origin=c]{90}{\nret}} 
        &$\ell_{1}$ &$0.974 (\pm0.004)$&$0.001 (\pm0.001)$&$0.582 (\pm0.004)$\\
        &$\ell_{2}$ &$0.955 (\pm0.004)$&$0.001 (\pm0.001)$&$0.569 (\pm0.004)$\\
        &$\ell_{\infty}$ &$0.769 (\pm0.052)$&$0.001 (\pm0.001)$&$0.307 (\pm0.035)$\\
        &cos &$0.914 (\pm0.007)$&$0.0 (\pm0.0)$&$0.46 (\pm0.007)$\\
        \end{tabular}}
    \\
    \centering
    \subfloat[CIFAR-100]{
    \begin{tabular}{c c | c c c}
        & & \multicolumn{2}{c}{Checkpointing Interval, $k$} & Deterministic\\
        & & $k=E\cdot S$ & $k=1$ & operations\\
        \toprule
        \multirow{4}{*}{\rotatebox[origin=c]{90}{\nret}} &$\ell_{1}$ &$0.903 (\pm0.002)$&$0.002 (\pm0.001)$&$0.903 (\pm0.001)$\\
        &$\ell_{2}$ &$0.815 (\pm0.002)$&$0.002 (\pm0.002)$&$0.816 (\pm0.001)$\\
        &$\ell_{\infty}$ &$0.532 (\pm0.07)$&$0.004 (\pm0.004)$&$0.51 (\pm0.055)$\\
        &cos &$0.383 (\pm0.002)$&$0.0 (\pm0.0)$&$0.384 (\pm0.002)$\\

        \end{tabular}}
    \caption{\textit{Normalized reproduction error, \nret, of a valid \prf.} The same initial parameter values $W_0$, batching strategy, model architecture, and training strategy are used. $W'_{t+k}$ is reproduced from $W_{t}$ by retraining $\forall t \in \{0,k,2k,\dots,T\}$ while \nret is computed as the distance between $W'_{t+k}$ and $W_{t+k}$ \textcolor{black}{normalized by \rd (see Table~\ref{tab:reference_distance} in Appendix~\ref{app:additional} for exact values of \rd)}. Deterministic operations used $k=E \cdot S$.}
    \label{tab:no_checkpoint}
\end{table}

\subsubsection{Deterministic Operations}
\label{sssec:determine_ops}
Libraries such as \texttt{pytorch} provide functionality that restrict the amount of randomness~\cite{pytorch_reproducibility} (\eg using deterministic algorithms for convolution operations) to enable reproducibility. We evaluate this with $k=T$ (refer Table~\ref{tab:no_checkpoint}). As seen in Table~\ref{tab:no_checkpoint}, \nre with deterministic operations drops to half of \nre for non-deterministic operations with ResNet-20. However, \nre is still significant and deterministic operations incur a large computational cost in training and a greater than one percentage point accuracy drop. The reduction in \nre is not observed for ResNet-50, which is likely because the main source of randomness for this architecture is not captured by deterministic operations provided by \texttt{pytorch}. Some other libraries use counter-based pseudorandom number generators, which will be discussed in \S~\ref{sec:discussion}.

\subsubsection{Checkpointing Interval and Storage Cost}\label{sec:checkpoint_storage_cost_analysis}
\begin{figure}[t]
    \centering
    \subfloat[CIFAR-10]{\includegraphics[width=0.5\linewidth]{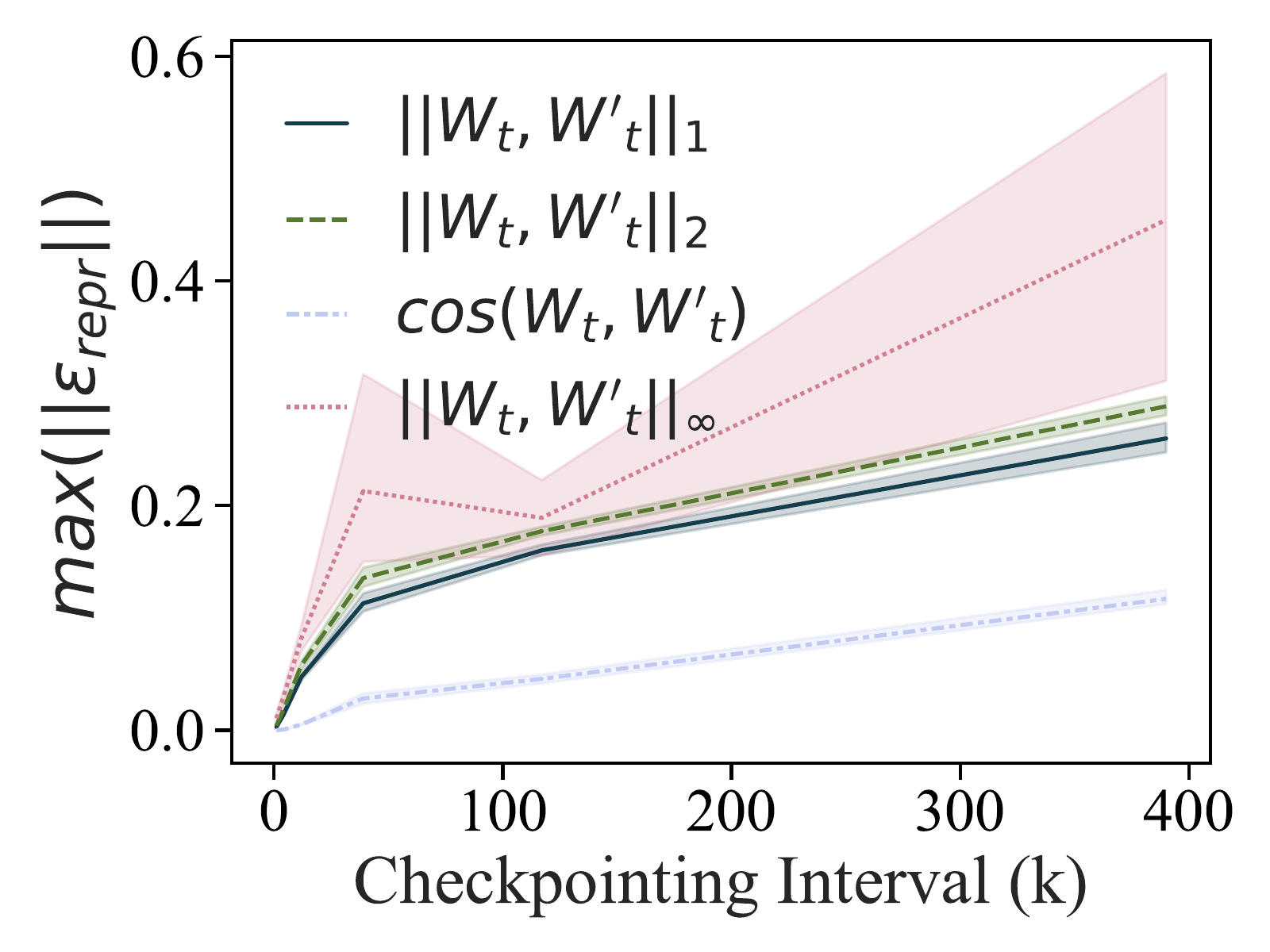}}
    \subfloat[CIFAR-100]{\includegraphics[width=0.5\linewidth]{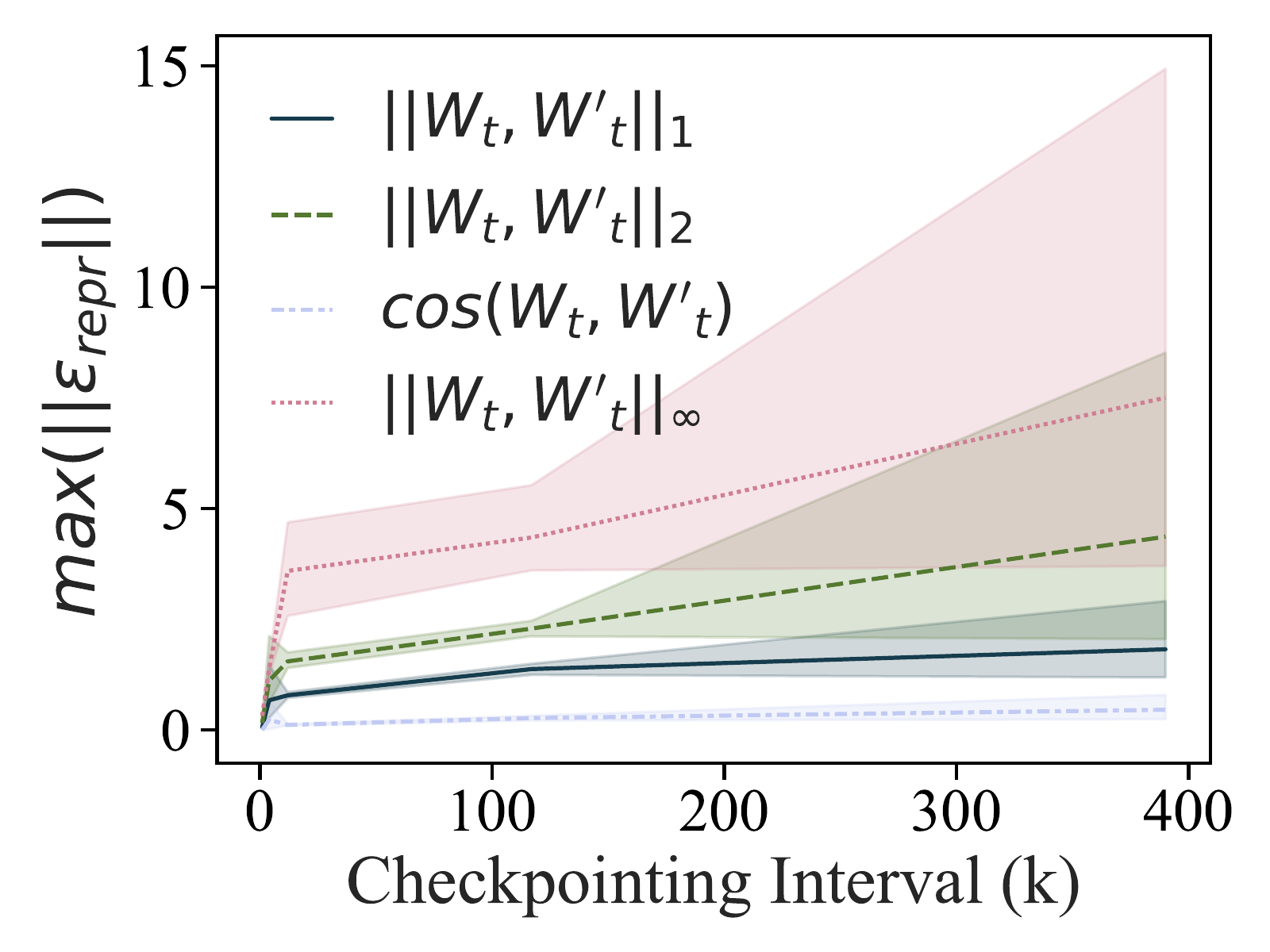}}
    \caption{\textcolor{black}{Normalized reproduction error, \nre,} as a function of the checkpoint interval, k. After choosing $k$, $\delta$ in Alg.~\ref{alg:verification} must be greater than \re($k$). Here, we define $cos = 1 - $cosine similarity.}
    \label{fig:reproducibility_save_freq}
\end{figure}

\begin{figure}[t]
    \centering
    \subfloat[CIFAR-10]{\includegraphics[width=0.5\linewidth]{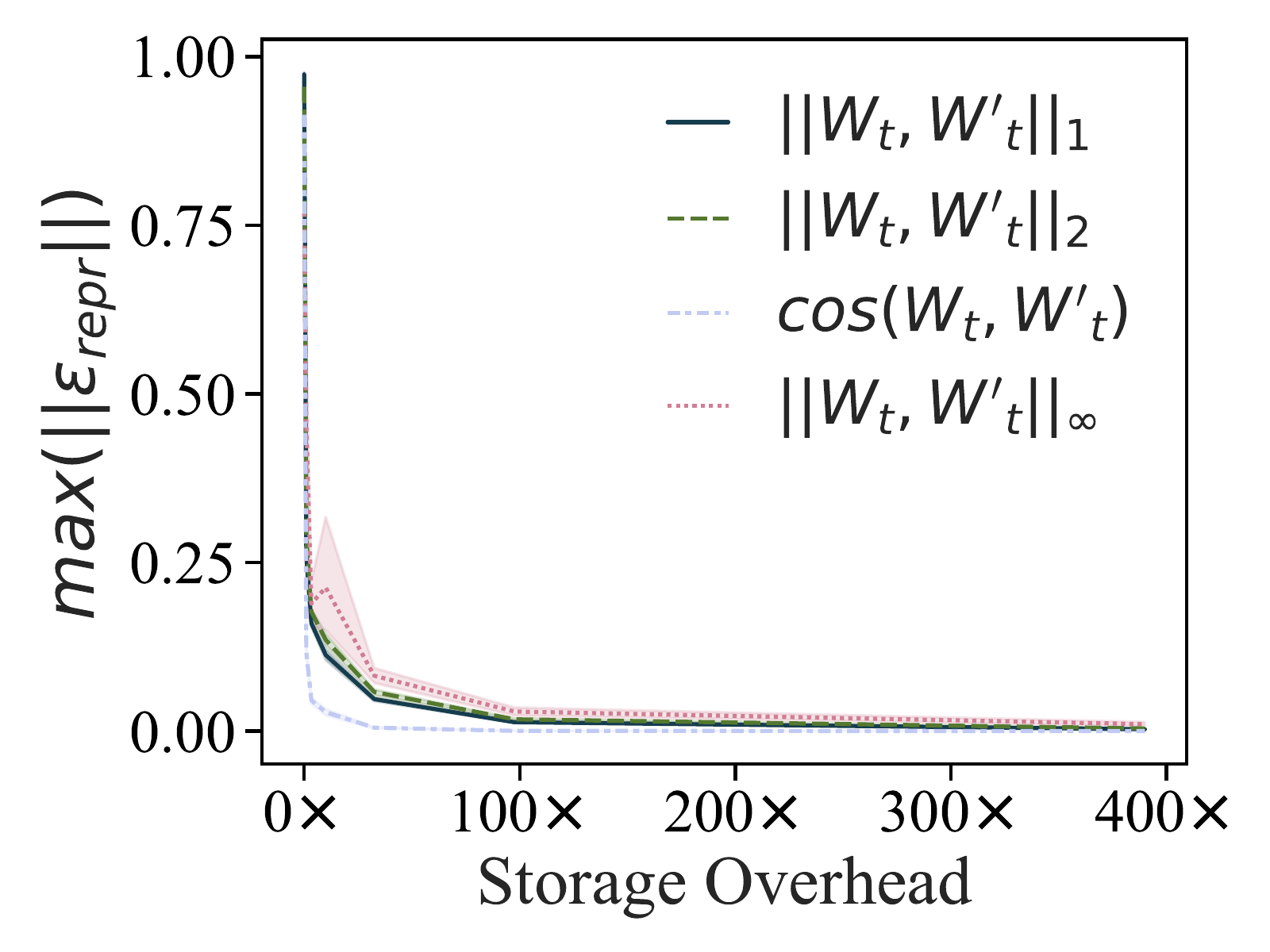}}
    \subfloat[CIFAR-100]{\includegraphics[width=0.5\linewidth]{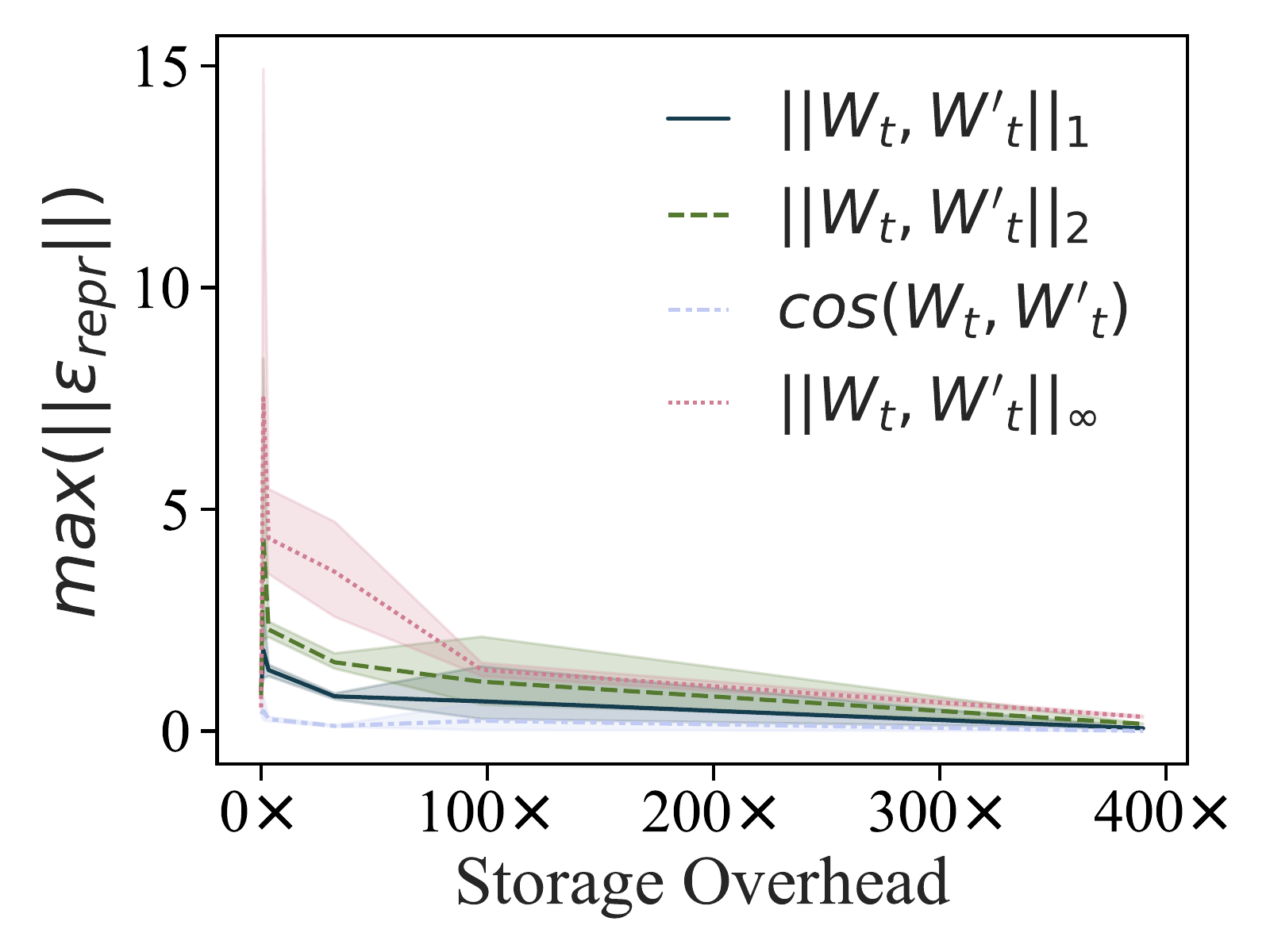}}
    \caption{Relation between \nre and storage overhead, where storage overhead is defined as the required number of checkpoints dividing by the number of epochs (assuming the prover checkpoints at every epoch even if not creating \prf). It can be seen \nre is still significantly lower than 1 when storage overhead = $1\times$ (\ie no storage overhead). \textcolor{black}{Figure~\ref{fig:reproducibility_memory} in Appendix~\ref{app:additional} shows sample values in megabytes.}}
    \label{fig:reproducibility_storage}
\end{figure}

The checkpointing interval $k$ is a hyperparameter of the proposed \prf method and is related to the storage cost, as the prover needs to checkpoint after every $k$ training steps. Common practice when training DNNs is to checkpoint at every epoch (\ie $k = S$) to allow resuming training and pick the model with highest accuracy after training, so we consider $k = S$ as a baseline and define the storage overhead as $\frac{S}{k}$. The relationships between \nre and $k$, and \nre and $\frac{S}{k}$ are shown in Figure~\ref{fig:reproducibility_save_freq} and~\ref{fig:reproducibility_storage} respectively. The most important observation from these figures is that \emph{the prover does not need to spend additional storage to save at every step, \ie $k=1$ suffices}. In particular, if the prover only utilizes the checkpoints saved roughly at every epoch ($k\approx S$), they can still attain \nre substantially below \nre $\approx 1$ for $k=T$. In Figures ~\ref{fig:reproducibility_save_freq}, ~\ref{fig:reproducibility_storage} and Table~\ref{tab:no_checkpoint} for the CIFAR-10 dataset, we observe that using $k=S$ outperforms creating \prf with the deterministic operations described in \S~\ref{sssec:determine_ops} and does not influence the speed of training or model's accuracy. Note that the prover could also save the checkpoints with a precision of \texttt{float16} rather than \texttt{float32} to save a factor of 2 in storage (please see \S~\ref{sec:discussion} for details on related storage considerations).

\subsubsection{Varying Learning Rate}
\begin{figure}[t]
    \centering
    \subfloat[CIFAR-10]{\includegraphics[width=0.5\linewidth]{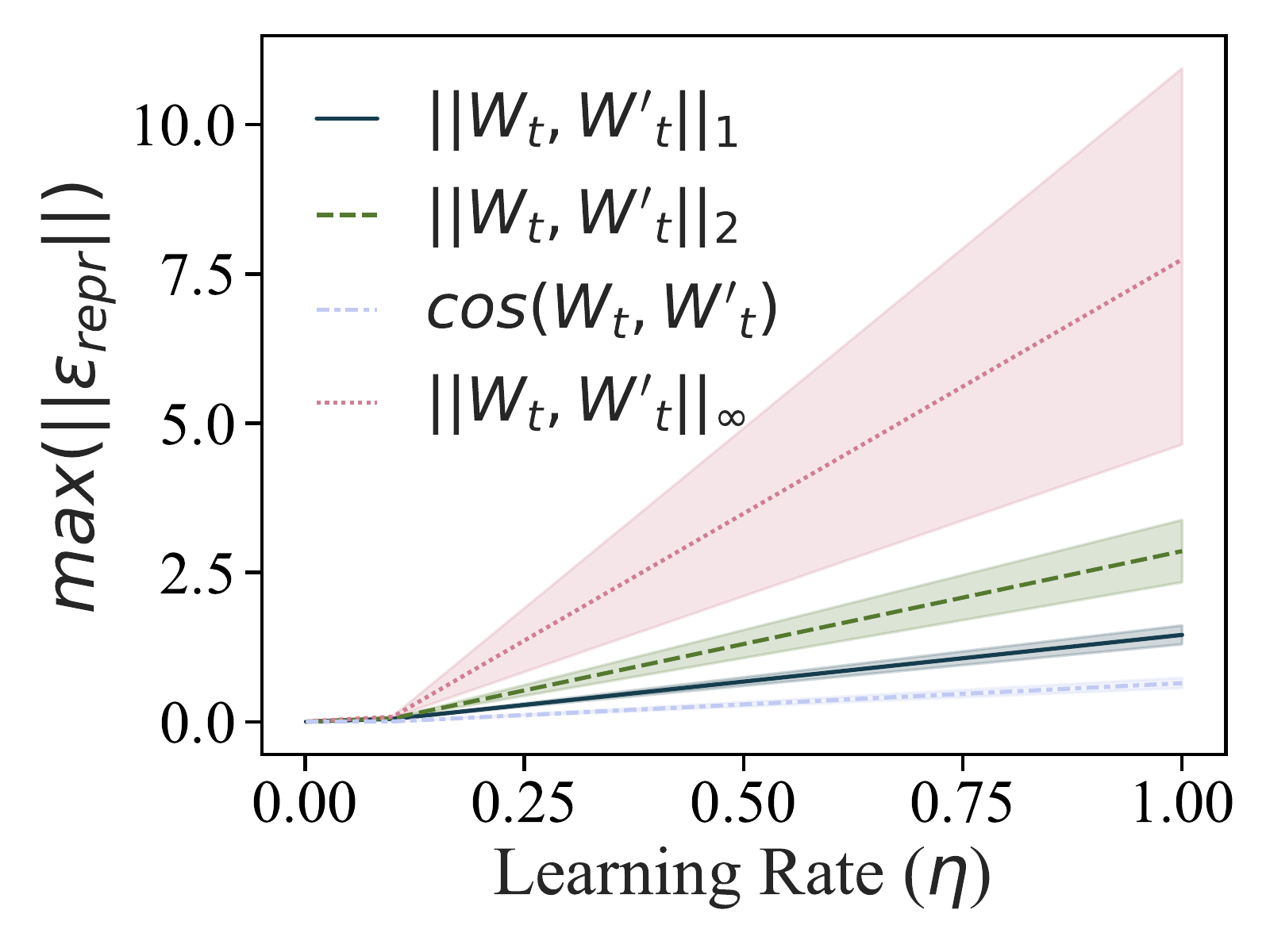}}
    \subfloat[CIFAR-100]{\includegraphics[width=0.5\linewidth]{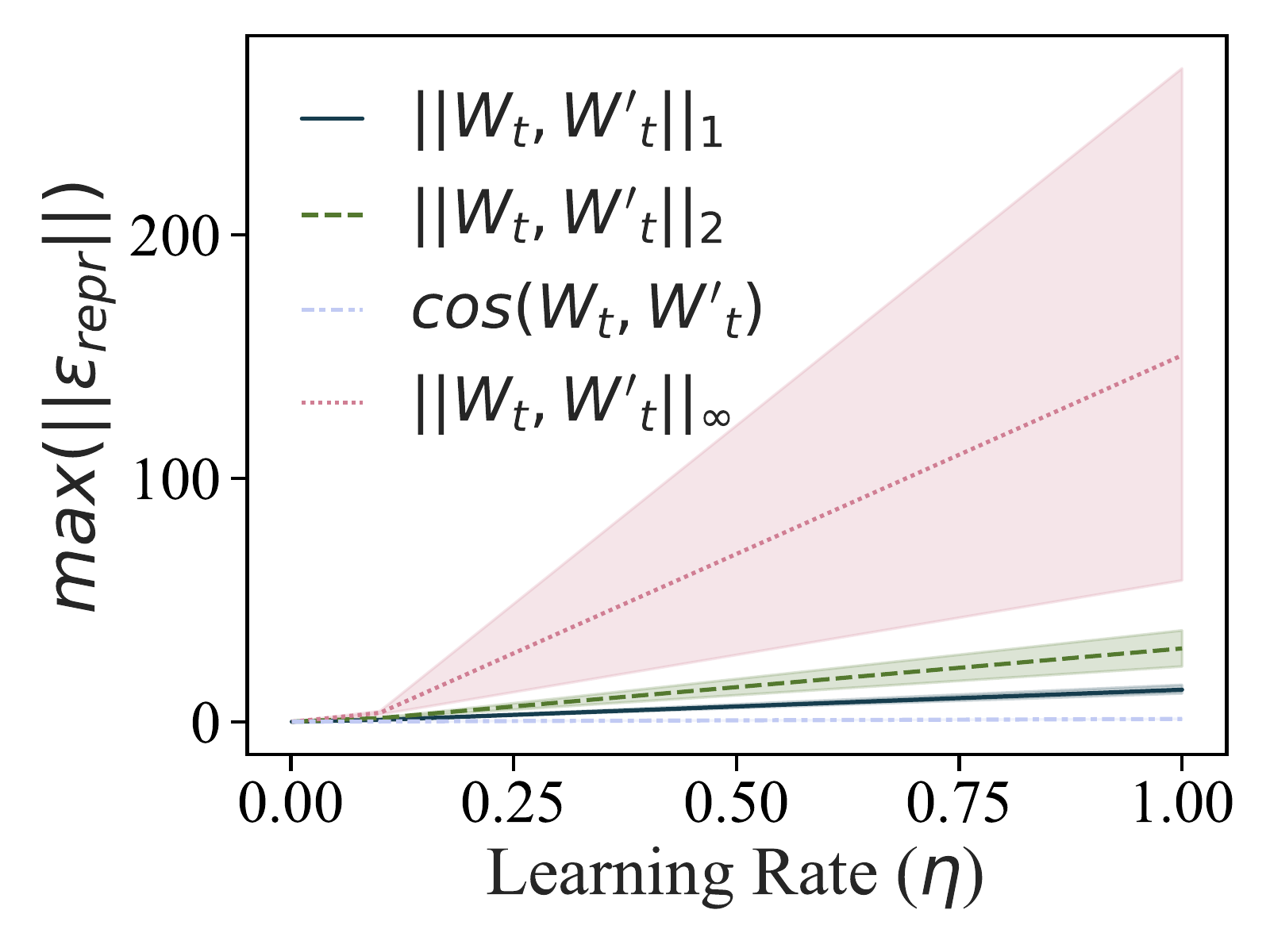}}
    \caption{Influence of learning rate, $\eta$, on \nre ($k=12$): if $\eta$ is in the order of magnitude smaller than $10^0$, $\eta$ does not have significant impact on \re. However, when $\eta$ is set to 1, \nre increases significantly.}
    \label{fig:reproducibility_lr}
\end{figure}

Since the proposed \prf relies on gradient updates, \nre is correlated to learning rate $\eta$, the hyperparameter that controls magnitude of gradient updates. Thus we present the relation between \nre and $\eta$ in Figure~\ref{fig:reproducibility_lr}. It can be seen $\eta$ has a significant impact on \nre only when it is set to 1. This may be because when $\eta$ is too large, the training process is unstable so a tiny difference may lead to distinct parameters after a few steps.

\subsection{Initialization Verification}
As described in \S~\ref{sec:chain-of-trust} and \S~\ref{sec:random-intialization}, if a prover claims their model is trained from cold-start (\ie rather than from pre-trained weights), a KS test is applied to verify whether the initial state in the \prf is sampled from a random distribution per the claimed initialization strategy.

\begin{table}[t]
    \centering
    \centering
    \begin{tabular}{c | c c}
        & CIFAR-10 & CIFAR-100\\
        \toprule
        Step & $7.00 (\pm 3.87)$ & $1(\pm 0)$ \\
        Accuracy &$10.526(\pm 0.953)\%$ & $1.124 (\pm 0.348)\%$\\
        \end{tabular}
    \caption{Index of the training step that p-values of the KS test dropped below the significance level, $\alpha=0.01$, and the corresponding validation accuracy. After this step, at least one layer is statistically different from a newly initialized layer.}
    \label{tab:check_initial}
\end{table}

\begin{figure}[t]
    \centering
    \subfloat[CIFAR-10]{\includegraphics[width=0.5\linewidth]{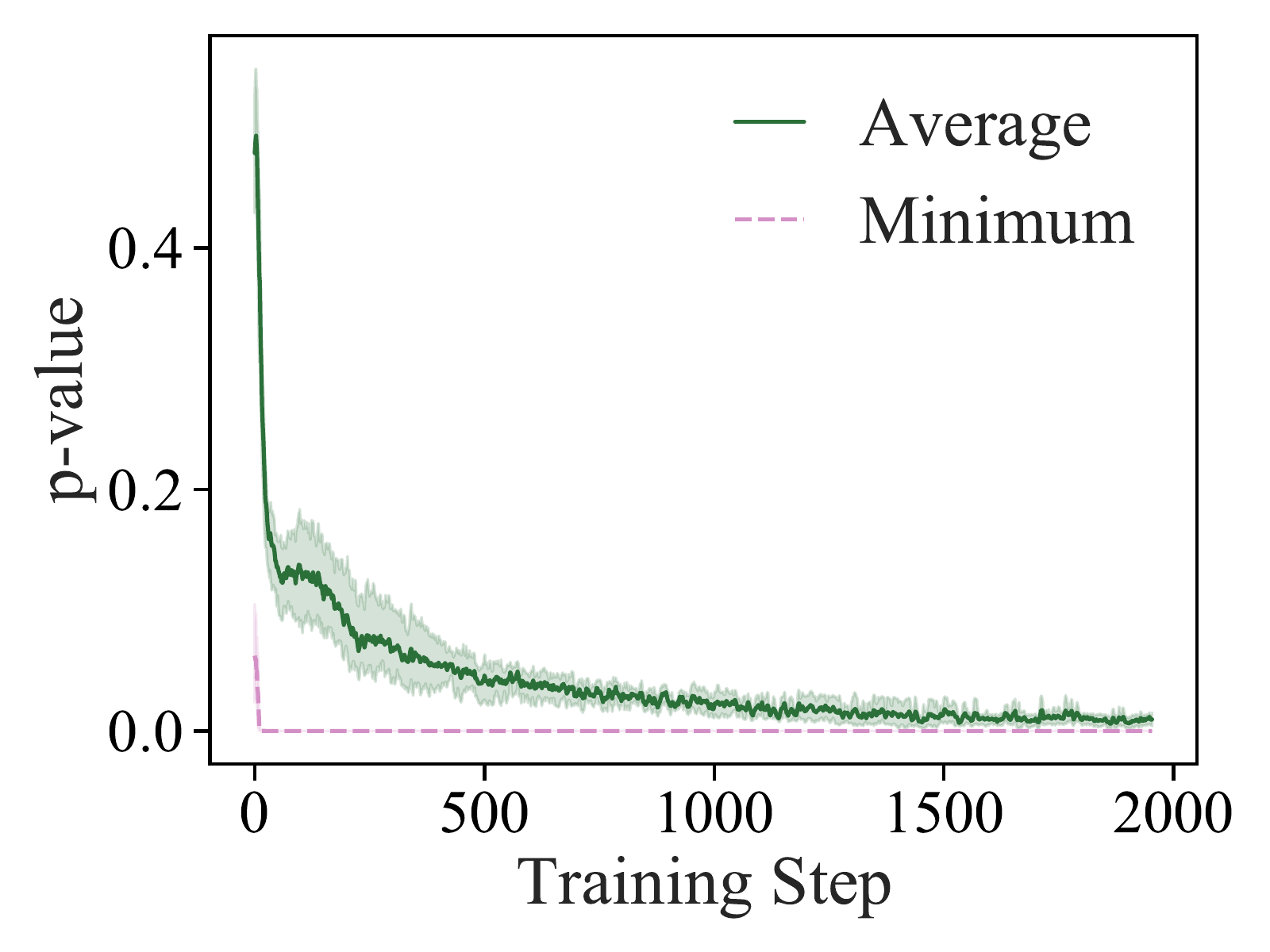}}
    \subfloat[CIFAR-100]{\includegraphics[width=0.5\linewidth]{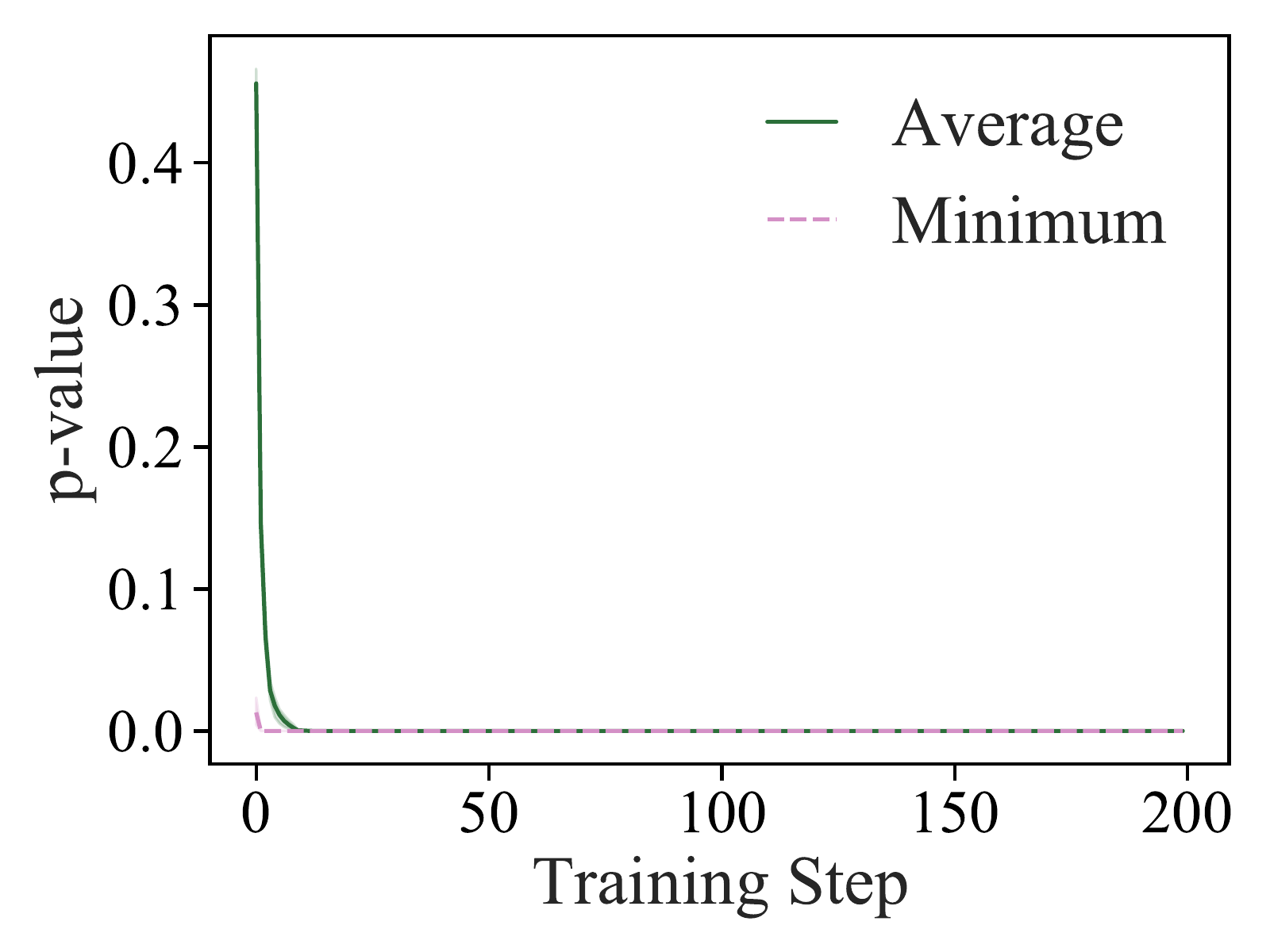}}
    \caption{p-value of Kolmogorov–Smirnov test with the null hypothesis that the model parameters came from the claimed initialization distributions, with respect to number of training steps: one can observe the minimum drops to almost zero within a few steps, meaning at least one layer has weights out of the initialization distributions.}
    \label{fig:initial_verify}
\end{figure}

Using the same setup as in \S~\ref{subsubsec:setup}, we applied a KS test to the early training steps (with $S=390$ for both datasets). As shown in Figure~\ref{fig:initial_verify}, for both models, the minimum p-value across all network layers drops to $0$ rapidly. We interpret this as: the weight distribution for at least one of the layers is statistically different from the initialization distribution. Observing Table~\ref{tab:check_initial}, 7 updates of ResNet-20 and 1 update of ResNet-50 on average would lead to p-value below 0.01, where the validation accuracy is only slightly higher than random guessing (\ie $10\%$ for CIFAR-10 and $1\%$ for CIFAR-100).

\section{Security Analysis of the Gradient Descent Mechanism for Proof-of-Work}
\label{sec:security_analysis}

Choosing a suitably low checkpointing interval allows us to control the entropy growth (in other words, the number of possible sequences of gradient descent). Controlling the entropy growth enables verification of the \prf: the prover $\prover$ can claim ownership in our model stealing scenario, or the model owner can trust the parameters shared by a worker in the distributed learning scenario (see \S\ref{sec:formalizing}).
Here we show that in addition, the entropy growth also creates an asymmetry between the adversary $\adv$ and verifier $\verifier$. This asymmetry disadvantages $\adv$ trying to spoof the \prf to pass verification with lesser computational effort \ie a structurally correct spoof. In light of this observation, we introduce and analyze two classes of spoofing strategies. 

\subsection{Defining a Spoof}
\label{ssec:defining-a-spoof}

Recall from \S~\ref{threat} that $\adv$ has gained access to $f_{W_T}$ (\ie its weights) but does not have a \prf that passes verification. Thus $\adv$ must create a spoof $\pw[\adv, f]$ proving that they trained the model $f$, where $f$ is an approximation of $f_{W_T}$ (denoted $f \approx f_{W_T})$, and has comparable test-time performance (see \S~\ref{ssec:initializations-difficulty}~and~\ref{ssec:pow-concatenation}). An adversary may always (re)perform the training required to obtain $f$. We call this an \textit{honest spoof} because $\mathbb{E}[C_{\adv}] \geq \mathbb{E}[C_\prover]$. Thus, the adversary gains nothing computationally beneficial from having gained access to $f_{W_T}$ and our verification scheme satisfies Property 2. %

\begin{definition}[Dishonest Spoof]\label{def:dishonest-spoof}
Any spoof for a prover's model $f_{W_T}$ that passes verification, and where the adversary expends fewer computational resources than the trainer, \ie $\mathbb{E}[C_{\adv}] < \mathbb{E}[C_\prover]$, is dishonest.
\end{definition}

Intuitively, for an attack to be \textit{dishonest}, the adversary would need to leverage knowing $f_{W_T}$ in order to possibly construct a \prf for $f_{W_T}$ using less computational resources than the $\prover$. Knowing the architecture of $f$ does not inform one on any part of a \prf other than the model one computed gradients on. However $W_T$ is the last state in $\pWeights$; thus, what we will consider are spoofing attacks that leverage knowing $W_T$ to construct a \prf. We will call spoofing attacks that use knowing $W_T$ to make  training less onerous as \textit{directed retraining} and those that attempt to reconstruct a \prf backwards starting from $W_T$ as \textit{inverse gradient methods}. These two methods encapsulate the two directions one could realistically obtain any ordered sequence (\ie a structurally correct \prf that may or may not pass verification): forwards (\ie directed retraining) and backwards (\ie inverse gradient methods).

\subsection{Inverse Gradient Methods}
Recall that Equation~\eqref{eq:sgd} defines a training step with SGD: given weights $W_{t-1}$ we find the next set of weights $W_t$. The inverse gradient step solves the inverse problem to this: given $W_t$ find the $W_{t-1}$ that led to it. We will denote this problem as solving $\beta(W_{t-1})$, where $\beta(W_{t-1})$ is defined as:
\begin{align}
    \beta(W_{t-1}) & \vcentcolon=  W_{t-1} - W_{t}  - \eta \nabla_{W_{t-1}} {\mathcal{L}} = 0  \label{eqn:root-finding}
\end{align}
Note that the batches these gradients are computed on do not necessarily have to be the same as those used in training, which is necessary as we do not assume the adversary has access to the batching strategy used in training (see \S~\ref{threat}).

To construct a complete \prf and pass verification, an adversary will iteratively repeat this inverse step and solve Equation~\eqref{eqn:root-finding} until they obtain a suitable $W_0$ that can be justified to have been either (a) sampled from a random distribution or (b) accompanied with a valid \prf $\pw^{0}$ in the chain-of-trust setting (see \S\ref{sec:chain-of-trust}). We  call this process of obtaining initial weights $W_0$ from the final weights $W_T$ the inverse gradient method. %

This approach is analogous to using the Euler Backward method to iteratively maximize the loss function, and is not new to ML in general \cite{yin2018stochastic}. However, to the best of our knowledge, it is new to DNNs, and we call it by a new name as to emphasize the context we are using it in; we are using this as an inverse procedure. As we will show, the top-Q strategy of verification (refer Algorithm~\ref{alg:verification}) will prevent this spoof.

\subsubsection{Entropy for the Inverse Gradient Process}~\label{ssec:entropy-inverse-gradient}
Recall from Theorem~\ref{theorem:-entropy-growth} that the forward process has a linearly increasing entropy growth with respect to the total number of training steps $T$. We now prove that the inverse gradient process is lower-bounded by this increase.
Recall Equation~\eqref{eq:trainin_process} which accounts for noise in SGD. To formulate the rate of entropy growth per inverse step, we take the conditional probabilities of $W_{t-1}$ with respect to $W_t$, as it was computed previously:
\begin{align}
    H(\tilde{W}_{t-1}|\tilde{W}_{t}) &= H(z_{0}) + H(\eta \nabla_{\tilde{W}_{t-1}} L|\tilde{W}_{t})\label{eq:reverse-entropy-formulation}
\end{align}
The inverse gradient process thus has higher entropy than the forward process if and only if $H(\eta \nabla_{\tilde{W}_{t-1}} L|\tilde{W}_{t}) >0$. This is true if and only if our inverse step (Equation~\eqref{eqn:root-finding}) has more than one solution with non-zero probability. That is, there is more than one training path using $\eta$ that reaches weights $W_t$.

\begin{theorem}[Reverse Entropy Growth]\label{theorem:reverse-entropy}
Similar to Theorem~\ref{theorem:-entropy-growth}, the unconstrained reverse training process, denoted by $\Theta_{-T} = \{W_T, W_{T-1}, \cdots, W_0 \}$, is also a Markov random process. It has equal or greater entropy than the forward training process $\Theta_{T}$, that is $H(\Theta_{-T}) \geq H(\Theta_{T})$, with equality if and only if $\nabla_{\tilde{W}_{t-1}} L|\tilde{W}_{t}$ is deterministic.
\end{theorem}

If the necessary and sufficient condition is true, then we necessarily have that the rate of entropy accumulation in inverting a training step is greater than the rate of entropy accumulation in the forward process: we would expect to see greater 
variance in our inverse paths than in our forward paths.

Given the large confidence intervals in Figures~\ref{fig:LeNet5-setups-error}~and~\ref{fig:LeNet5-comparing-lr} (see~\ref{ssec:general-inverse-gradient-method} for experimental setup), we hypothesize that these necessary and sufficient conditions are true for DNN, \ie there are several training paths passing through the same weights. We leave to future work the rigorous verification of these conditions because they are not necessary to refute the inverse gradient-based spoofing attacks that we propose.

\subsubsection{Retraining-based Spoofing}

Here we show why an inverse gradient approach is not effective to exactly reconstruct a spoof, \ie perform retraining-based spoofing to obtain $\pw[\adv, \model_{W_T}] = \pw[\prover, \model_{W_T}]$. From Theorem~\ref{theorem:reverse-entropy} we know that the entropy of inverting a sequence $H(\Theta_{-T})$ is lower bounded by the entropy of training the sequence $H(\Theta_{T})$, which we know grows linearly with $T$. Recall from \S~\ref{ssec:interpretation-of-entropy} that this entropy represents an exponential increase in the number of paths to reach $W_T$. As DNN training requires thousands of steps, we can safely say that the probability of following any given path is near-zero.
Thus for any sequence sufficiently long, \ie $T \gg 0$, we can dismiss the inverse gradient method for a reconstruction spoof because the probability of recreating a specific sequence is negligible, \ie $\approx 0$. Indeed, our results for reproducability (see Table~\ref{tab:no_checkpoint}) show empirically that the lengths used for training a DNN satisfy this condition.

\subsubsection{Stochastic Spoofing}
\label{ssec:general-inverse-gradient-method}

To overcome the challenges of exactly recreating $\pw[\prover,\model_{W_T}]$, an adversary employing the general inverse spoof instead focuses on obtaining a different \prf $\pw[\adv, f]\neq\pw[\prover, f_{W_T}]$ that regardless passes verification. As we show, this is not beneficial as the adversary  still faces a computational cost at least as large as that for $\prover$ and it is difficult %
to end in a suitable random initialization.

\paragraph{The Computational Costs} 
Any numerical approach to solving  Equation~(\ref{eqn:root-finding}) will require at least one function call to Equation~(\ref{eqn:root-finding}), \eg to check that the provided solution is indeed the correct solution. Since computing Equation~\eqref{eqn:root-finding} requires computing $\nabla_{W_{t-1}}\mathcal{L}$, \ie, one training step, inverting a training step is bounded by the computational load of a training step. We remark that DNNs are highly non-linear and as such there are no known analytical solutions to Equation~\eqref{eqn:root-finding}. Thus attempting to create a \prf such that $\pw[\adv, \model_{W_T}]\neq\pw[\prover, \model_{W_T}]$ but that passes verification would be at least as computationally expensive as what it took $\prover$.

\begin{figure}[t]
\centering
\includegraphics[width=0.78\columnwidth]{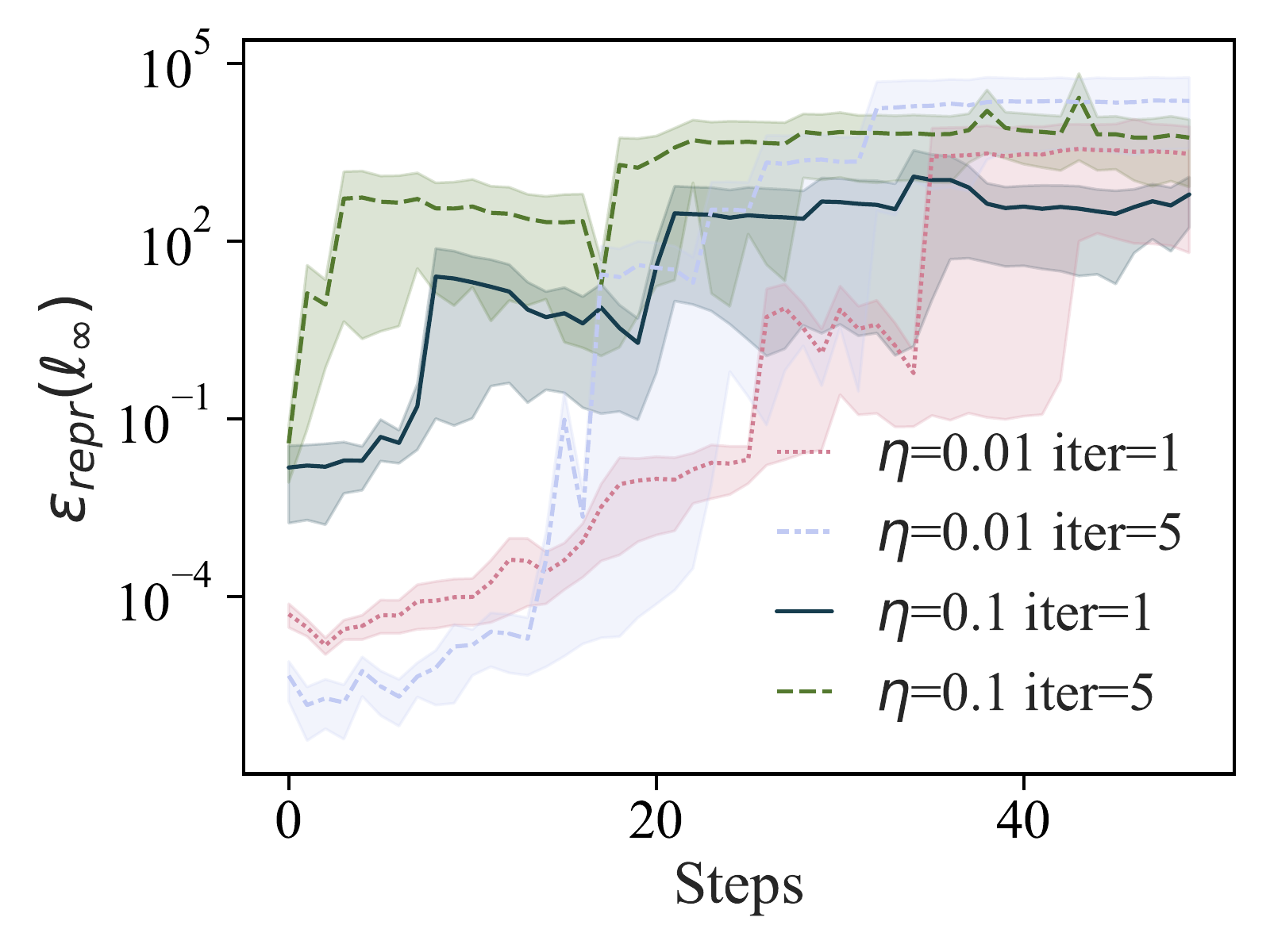}
\caption{Inverting gradients on LeNet5 leads to an $l_{inf}$ error that is several orders of magnitude higher as the learning rate increases from $0.1$ to $0.01$.\vspace{-5mm}}
\label{fig:LeNet5-setups-error}
\end{figure}

\begin{figure}[b]
\centering
\subfloat[$\ell_2$ distance\label{fig:LeNet5_l2}]{\includegraphics[width=0.5\columnwidth]{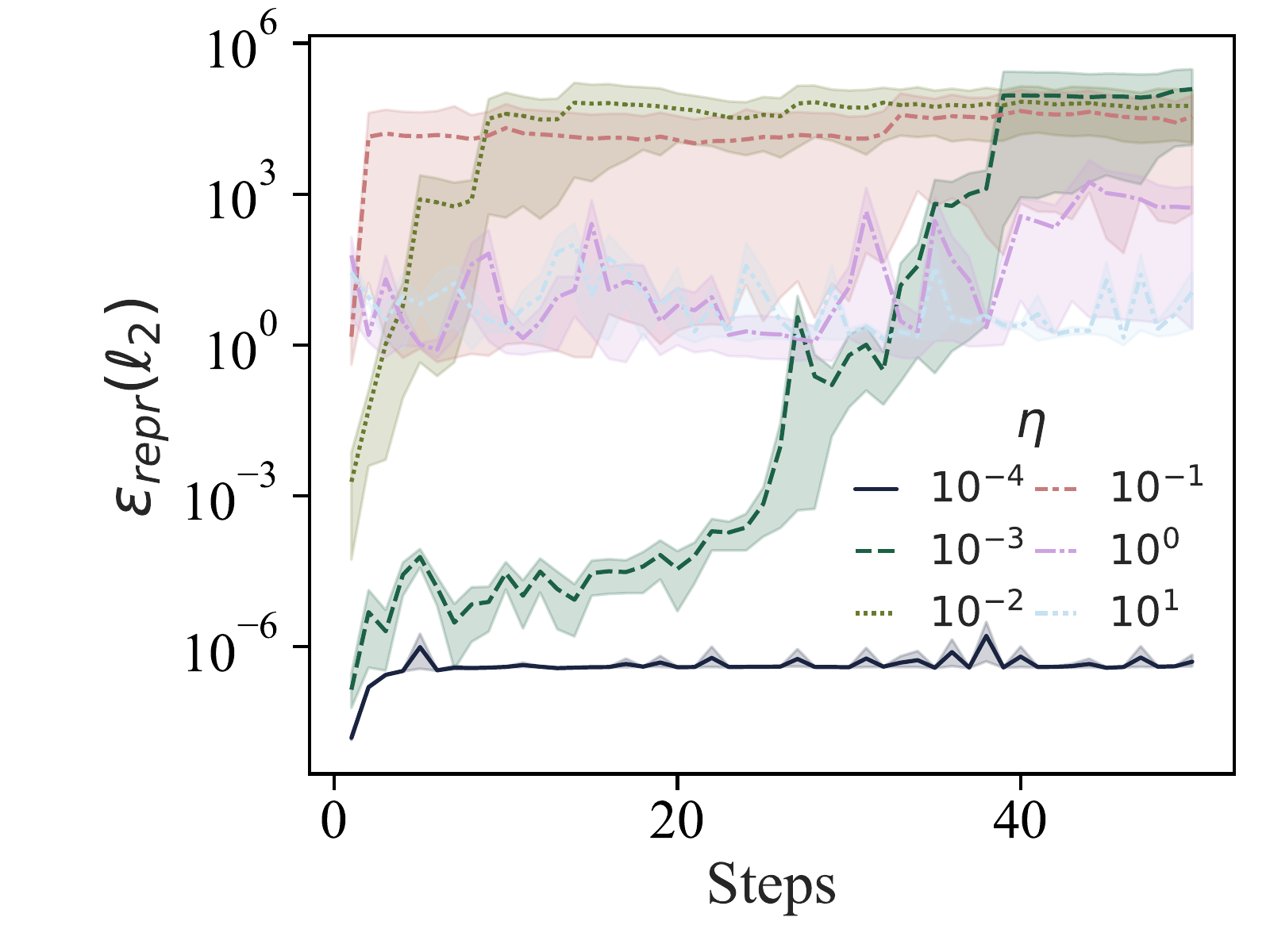}}
\subfloat[$\ell_\infty$ distance\label{fig:LeNet5-comparing-lr}]{\includegraphics[width=0.5\columnwidth]{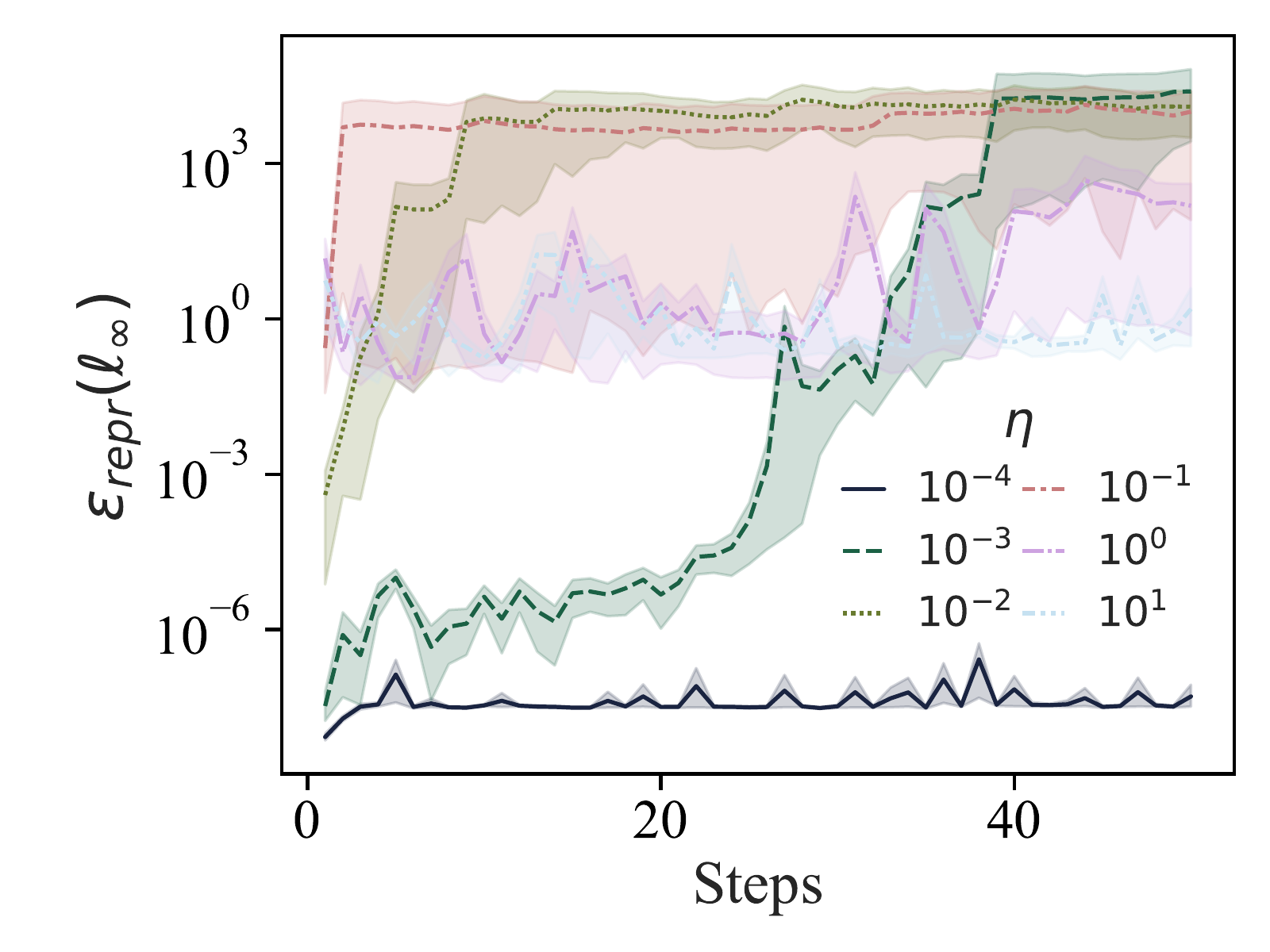}}
\caption{Observe that for larger learning rates, numerical methods are unable to converge to a sufficiently small error. Thus, using large learning rates is infeasible.\vspace{-5mm}}

\end{figure}

\begin{figure}[t]
\centering
\includegraphics[width=0.8\columnwidth]{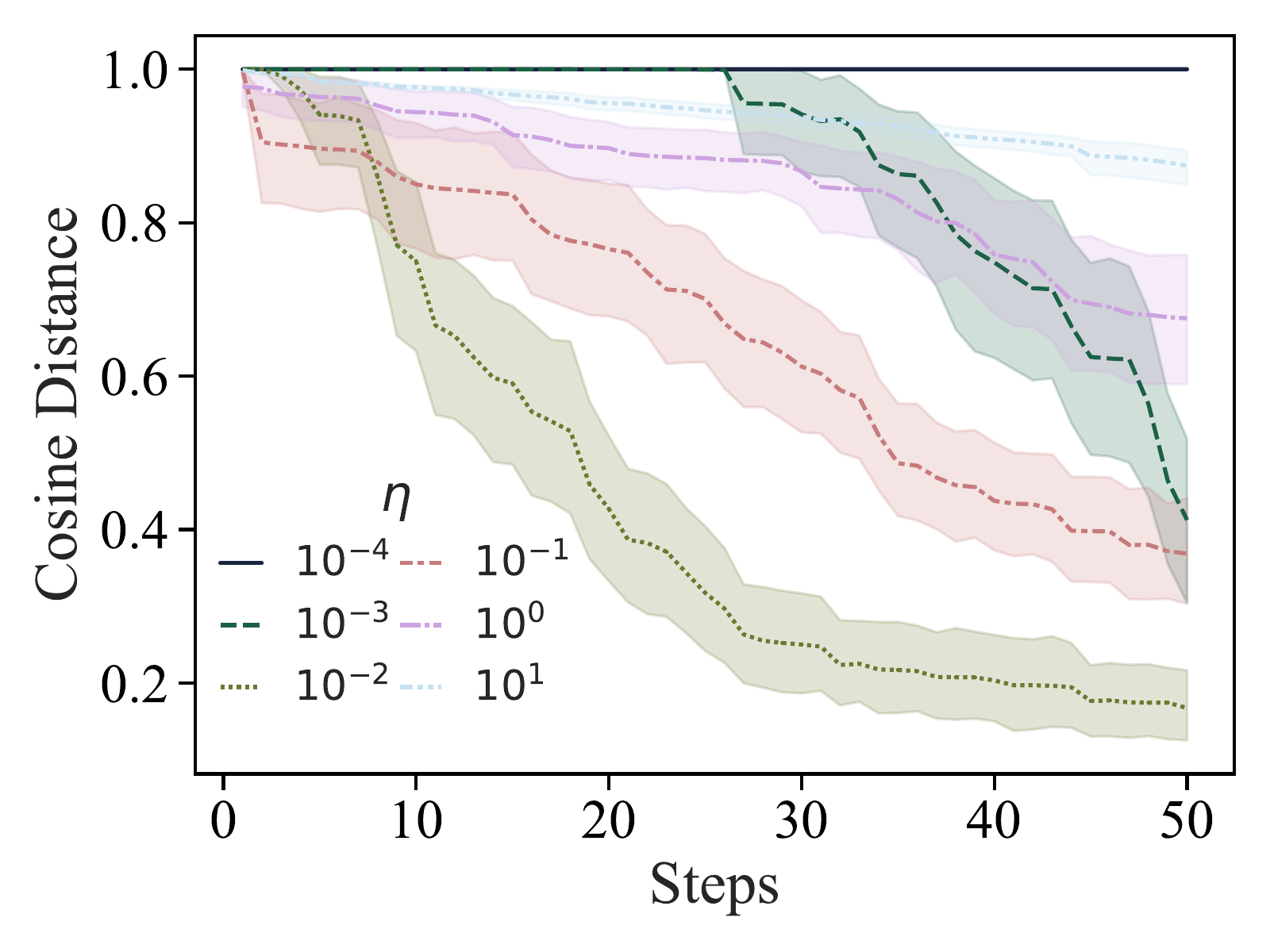}
\caption{Observe that the cosine measure relative to the trained sequence for all learning rates tested steadily decreases.\vspace{-5mm}}
\label{fig:LeNet5_cosine}
\end{figure}

The only remaining strategy to make the computational costs feasible, while maintaining $f=f_{W_T}$, is for an adversary to take larger inverse steps, \ie use larger learning rates so as to reduce the length of the \prf. To disprove this we conducted experiments on a LeNet5 model~\cite{lecun1995comparison}
on the MNIST dataset~\cite{lecun1998mnist}. The first set of experiments compared the effect of the learning rate to reconstruction error \re after each step $t$ (see Figures~\ref{fig:LeNet5-comparing-lr}, \ref{fig:LeNet5_l2}, and~\ref{fig:LeNet5_cosine}), and the second compared the effect of fewer and more iterations of the root solver for moderate learning rates (see Figure~\ref{fig:LeNet5-setups-error}). We ran all these experiments inverting 50 steps (with $k=1$  ) from a state achieved after $5$ epochs of training. All experiments are repeated $35$ times to capture variance as seen in the confidence intervals.
We further evaluated on ResNet models on CIFAR-10 and CIFAR-100, the experimental setup of which is described earlier in \S~\ref{subsubsec:setup}.

As seen from these experiments, the reproducability error (the error between where a training step from $W_{t-1}$ leads and $W_{t}$) quickly increases after a few steps for learning rates above $10^{-4}$, meaning the \prf obtained is not valid. As this was the case for a relatively small model, we also expect this to be the case for larger models; our tests on inverting ResNet models also resulted in average \re larger than those found when training with $k=1$ (see Tables~\ref{tab:no_checkpoint} and~\ref{tab:inverse_error}). Thus, we have empirically determined that an adversary cannot use higher learning rates to decrease the computational load.

From the argument we have made (\hyperref[sec:Gs]{\textbf{G2}}), and given that we are not aware of a mechanism to prove this formally, we present the following as a conjecture:

\begin{conjecture}
Inverting a training sequence using numerical root finding methods will always be at least as computationally expensive as training, given the same model.
\end{conjecture}

\paragraph{Difficulty of Finding a Suitable Initialization}\label{ssec:initializations-difficulty}
As mentioned in \S~\ref{sec:random-intialization}, a valid initialization must pass the KS test~\cite{kstest}. To test the initialization, the verifier compares it against the public pool of known initializations, \eg various forms of zero-centered uniform and normal distributions~\cite{He_2015_ICCV,NIPS2017_5d44ee6f,glorot2010understanding}. Thus, the adversary must in addition successfully spoof the initialization to pass the KS test. Our empirical results indicate that inverse gradient methods are unlikely to find a valid initialization. Specifically, we inverted 50 steps on a model  trained for 50 steps, and applied the KS test to the last state of inverting (corresponding to the first state of training) as described in \S~\ref{sec:random-intialization}. On CIFAR-10 we observe that the the average and minimum p-values are $0.044 (\pm 0.102)$ and $1.077(\pm 1.864)\times 10^{-28}$, respectively. On CIFAR-100, the average and minimum p-values are $0.583(\pm 7.765)\times 10^{-12}$ and $0 (\pm 0)$, respectively. These p-values are far below the required threshold to pass the KS test and thus an adversary is unable to find a valid initialization sampled from a claimed distribution. A clever adversary may attempt to \textit{direct} the inverse gradient method toward a valid initialization. We discuss in \S~\ref{ssec:directed-retraining} below how these directed approaches do not succeed in passing our verification scheme. We remark that the KS test prevents other spoofing strategies, such as leveraging fine-pruning~\cite{liu2018fine} or sparsification~\cite{frankle2018lottery}. These strategies can significantly minimize the computational load of spoofing while maintaining both the model architecture and test-time performance, \ie $f\approx f_{W_T}$. However, they as well fail to pass the KS test and thus are not verified by our scheme.

\begin{table}[t]
    \centering
    \begin{tabular}{c c | c c}
        &  & CIFAR-10 & CIFAR-100\\
        \toprule
        \multirow{4}{*}{\rotatebox[origin=c]{90}{\nre}}
        &$\ell_{1}$&$ 0.023 \pm 0.001 $&$ 0.005 \pm 0.001 $\\
        &$\ell_{2}$&$ 0.048 \pm 0.004 $&$ 0.016 \pm 0.005 $\\
        &$\ell_{\infty}$&$ 0.18 \pm 0.044 $&$ 0.073 \pm 0.014 $\\
        &cos&$ 0.016 \pm 0.002 $&$ 0.0 \pm 0.0 $\\

    \end{tabular}
    \caption{\textit{Normalized reproduction error, \nre of \prf created by \textit{General  Inverse  Gradient  Method.}} The trained models inverted for 50 steps to obtain a \prf with length 50 and $k=1$. The \re is then computed on this \prf. Comparing to the $k=1$ case in Table~\ref{tab:no_checkpoint}, the \re here is larger.}
    \label{tab:inverse_error}
\end{table}

\subsection{Directed Retraining}
\label{ssec:directed-retraining}
Given no extra knowledge, retraining $\model_{W_T}$ would take as much compute as used by $\prover$. However, the adversary always has the additional advantage of knowing the final weights $W_T$. We now explore how the adversary can leverage this knowledge to create a dishonest spoof (see Definition~\ref{def:dishonest-spoof}).

\subsubsection{Approach 1: \prf Concatenation}
\label{ssec:pow-concatenation}

An adversary $\adv$ aware that $\verifier$ does not verify all the updates may try to exploit this and employ structurally correct spoofing (refer \S~\ref{threat}) to obtain a partially valid \prf that may pass the verification. To this end, the adversary can fine-tune~\cite{yosinski2014transferable} or fine-prune~\cite{liu2018fine} the model $f_{W_T}$ to achieve $f$ which is not an exact copy of $f_{W_T}$ but has comparable test-time performance. This step provides the adversary with a valid \prf from $f_{W_T}$ to $f$. However, this would still be detected by Algorithm~\ref{alg:verification} because $\verifier$ also checks the initial state (recall \S~\ref{sec:random-intialization}), which in the adversary's \prf is ${W_T}$ (for which it has no valid \prf).

To adapt, the adversary can train a model with the same architecture as $f_{W_T}$ from a random initialization for some number of steps with minimal cost, providing a second valid \prf, this time starting from a valid random initialization. Then, the adversary concatenates these two \prf{s}. In addition to saving compute, the advantage of this strategy is that there is only one single point of discontinuity in the \prf, which consists of thousands of updates. Thus if $\verifier$ randomly sampled a few updates to check, the $\adv$'s \prf would likely go undetected. However, since $\verifier$ verifies the top-$Q$ updates in Algorithm~\ref{alg:verification}, this discontinuity which is among the largest of the sequence  would be invalidated---as we evaluate next.\\ 

\begin{figure}[t]
    \centering
    \subfloat[CIFAR-10]{\includegraphics[width=0.5\linewidth]{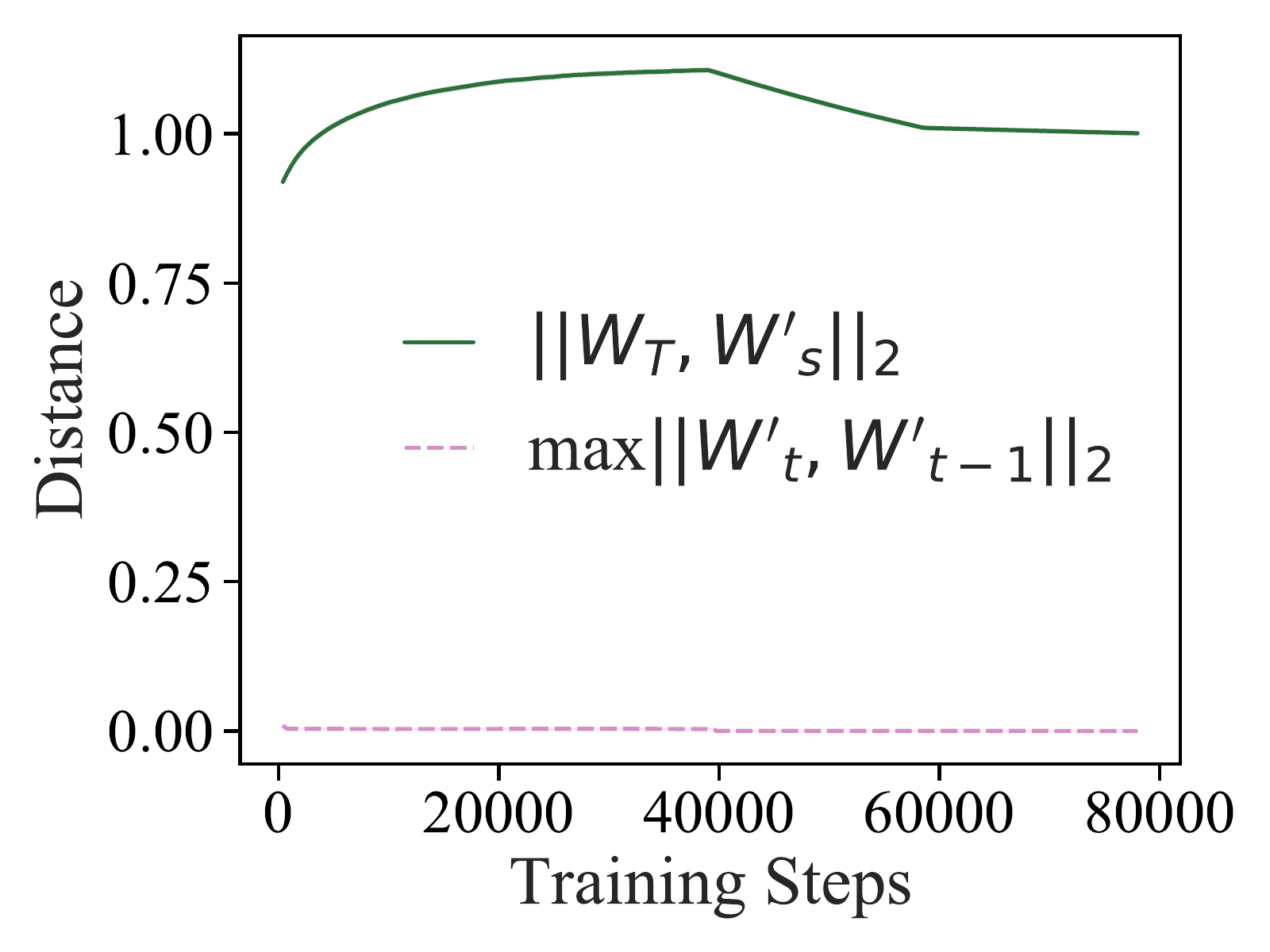}}
    \subfloat[CIFAR-100]{\includegraphics[width=0.5\linewidth]{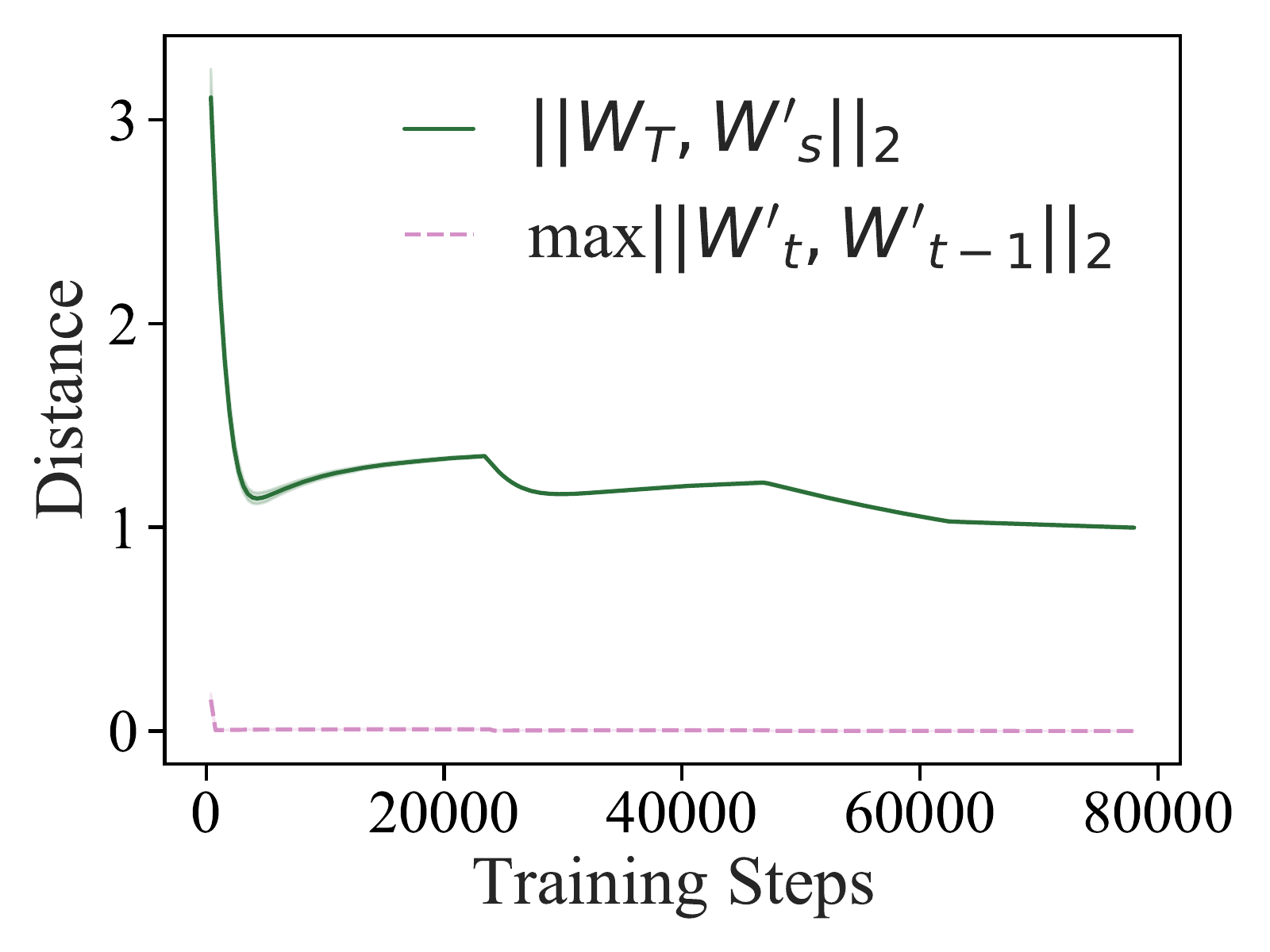}}
    \caption{Magnitude of discontinuity $||W_T - W'_s||_2$ and largest valid update $\max||W'_t - W'_{t-1}||_2$ in a spoofing \prf made by concatenating 2 valid but independent \prf. The discontinuity is significantly larger than the valid updates, and thus easily detected by Algorithm~\ref{alg:verification} which checks the largest updates first.}
    \label{fig:concat_spoof}
\end{figure}

\para{Evaluation} Our evaluation is performed with the setup from \S~\ref{subsubsec:setup}. For each dataset, we first train a model to completion as the prover $\prover$'s model $W_T$. Then we play the role of $\adv$ to spoof a \prf by concatenation: we fine-tune $W_T$ for 1 epoch to get $f$, \textcolor{black}{and train another model (from scratch) with the same architecture for $s$ steps ($s \leq T$) from initialization (\ie $W'_0$ to $W'_s$); $s$ is the number of steps on the x-axis in Figure~\ref{fig:concat_spoof}. We plot $||W_T - W'_s||_2$ and $\max_{t\leq s}||W'_t - W'_{t-k}||_2$ (both normalized (by \rd)) in this figure (with $k=1$).} We observe that: 
\begin{itemize}
\item The discontinuity (\ie $||W_T - W'_s||_2$) is much larger than all valid gradient updates in the \prf, so setting $Q=1$ would be sufficient for the verifier to detect this spoofing. The verification cost is  $E\cdot k = E$ steps of gradient updates (since we set $k=1$ for this experiment). However, if the verifier randomly samples $E$ steps (rather than picking the top-1 step of every epoch), the probability of finding the discontinuity is only $\frac{1}{S}$, with $S=390$ here.
\item The discontinuity has similar magnitude to \rd, revealing the fact that $W_T$ and $W'_s$ are unrelated.
\item \textcolor{black}{$\max_{t\leq s}||W'_t - W'_{t-k}||_2$ does vary significantly with respect to $s$, meaning setting $\delta$ to $||W'_t - W'_{t-k}||_2$ for small $t$ is sufficient to detect this kind of attack.}
\end{itemize}

\textcolor{black}{It is worth noting that if the adversary $\adv$ has knowledge about $Q$, or verifier $\verifier$ sets $Q$ to a small value, $\adv$ may make $Q$ (or more) legitimate updates in every epoch by training with an arbitrarily large learning rate, which will bypass Algorithm~\ref{alg:verification}. 
Solutions to this issue could involve (a) using a large $Q$, (b) randomly verifying some more updates, or (c) checking model performance periodically since the arbitrarily large updates would likely decrease model performance significantly.}

\subsubsection{Approach 2: Directed Weight Minimization} To minimize the discontinuity magnitude, an adversary may attempt to direct the weights of retraining toward $W_T$. To achieve this, they can directly minimize this distance using regularization. This approach fails verification because the custom regularizer requires the final weights prior to them having been achieved, which therefore cannot pass verification. Further, this information cannot be easily distilled into synthetic data because no gradient of the regularization term, with respect to the data, exists (refer to Appendix~\ref{app:infeasibility-directed-retraining} for more details).
By this vain, other tactics, such as optimizing a learning rate $\eta$ to converge $W'$ to $W_T$ also fail verification.

\section{Discussions \& Limitations}
\label{sec:discussion}
\textcolor{black}{
A \prf provides grounds for proving ownership of \textit{any} effortful attempt at learning a model. As shown in \S~\ref{sec:correctness_analysis}, a \prf guarantees that no one but the trainer can lay claim to that exact model. Further, if a chain-of-trust is adopted, this guarantee is extended to the use of the said model as an initial state for the training of a surrogate model. However, a \prf cannot be used to connect the model to its surrogate, neither can it be used avoid extraction. \textcolor{black}{Instead, a \prf provides legal protection:} if the trainer produces a \prf and publishes a time-stamped signature of it, this unchangeable record proves ownership in case of false claim by a surrogate model owner.} 

We now discuss limitations with our proposed scheme for \prf. First, our verification scheme requires that the training data be shared with the verifier. When this data is private, this can be undesirable. To protect the training data's confidentiality, it is possible for the prover to engage in a private inference protocol with the verifier~\cite{secureml} using multi-party computation.
This will incur additional computational overhead but is only limited on the chosen private inference scheme.

Second, we note the considerable storage requirements our proposed proof-of-work imposes. To decrease the approach's footprint by a factor of 2, we downcast the \texttt{float32} values of our parameters to \texttt{float16} when saving them. Verifying \texttt{float16} values introduces minimal error. We acknowledge that other approaches such as hashing could provide significantly better improvement to the storage footprint. For example, follow up work may consider hashing weights sequentially utilizing Merkle tree structure~\cite{merkle1987digital}, i.e. each consecutive set of weights during the training procedure are hashed and then saved as the hash of the concatenation of the current weights and the previously saved hash. %
We do not use Merkle trees due to the error accumulated when the verifier reconstructs the weights: the error in the weights forces the weights of the verifier and 
legitimate worker to hash to different values, losing the ability to verify that the weights match within some bound. This may be addressed with fuzzy extractors or locality sensitive hashing (LSH). However, the use of fuzzy extractors and LSH protocols incurs significant difficulty through the need to find a suitable bound to work over all choices of $E$, $Q$, and $k$. Designing such primitives is future work.

Third, we emphasize that counter-based pseudorandom number generators~\cite{salmon2011parallel,claessen2013splittable} can potentially remove most, if not all, noise in the training process because the pseudorandom numbers are generated based only off the input seed, not any hardware-based source of entropy. Recall that this noise introduces the random variable $z$ in Theorems~\ref{theorem:-entropy-growth}~and~\ref{theorem:reverse-entropy}. While there is currently no ground-truth for all sources of randomness arising in ML training through hardware, low-level libraries, and random number generation, 
such ground-truths 
would make training more reproducible and facilitate our approach.

Finally, we remark that our probability of success for our verification scheme degrades multiplicatively with each usage. This limits its usage for extremely long chains of \prf{}s (e.g., when successively transfer learning between many models) where any given probability of success is significantly below $1$. As there is currently no \prf scheme to gain practical insight on this limitation, we leave this to future work.

\section{Conclusions}
Our analysis  shows gradient descent naturally produces secret information due to its stochasticity, and this information can serve as a proof-of-learning. We find that entropy growth during training creates an asymmetry between the adversary and defender which advantages the defender. Perhaps the strongest advantage of our approach is that it requires no changes to the existing training procedure, and adds little overhead for the prover seeking to prove they have trained a model. We expect that future work will expand on the notion of proof-of-learning introduced here, and propose improved mechanisms applicable beyond the two scenarios which motivated our work (model stealing and distributed training).

\section*{Acknowledgments}
\noindent We thank the reviewers for their insightful feedback. This work was supported by CIFAR (through a Canada CIFAR AI Chair), by NSERC (under the Discovery Program, NFRF Exploration program, and COHESA strategic research network), and by gifts from Intel and Microsoft. We also thank the Vector Institute's sponsors. Varun was supported in part through the following US National Science Foundation grants: CNS-1838733, CNS-1719336, CNS-1647152, CNS-1629833 and CNS-2003129, and the Landweber fellowship.

{
\printbibliography

@inproceedings{
maini2021dataset,
title={Dataset Inference: Ownership Resolution in Machine Learning},
author={Pratyush Maini and Mohammad Yaghini and Nicolas Papernot},
booktitle={International Conference on Learning Representations},
year={2021},
url={https://openreview.net/forum?id=hvdKKV2yt7T}
}

@inproceedings{jia2020entangled,
  title={Entangled watermarks as a defense against model extraction},
  author={Jia, Hengrui and Choquette-Choo, Christopher A and Chandrasekaran, Varun and Papernot, Nicolas},
  booktitle={30th $\{$USENIX$\}$ Security Symposium ($\{$USENIX$\}$ Security 21)},
  year={2021}
}

@inproceedings{jagielski2020high,
  title={High Accuracy and High Fidelity Extraction of Neural Networks},
  author={Jagielski, Matthew and Carlini, Nicholas and Berthelot, David and Kurakin, Alex and Papernot, Nicolas},
  booktitle={29th $\{$USENIX$\}$ Security Symposium ($\{$USENIX$\}$ Security 20)},
  year={2020}
}

@inproceedings{yosinski2014transferable,
  title={How transferable are features in deep neural networks?},
  author={Yosinski, Jason and Clune, Jeff and Bengio, Yoshua and Lipson, Hod},
  booktitle={Advances in neural information processing systems},
  pages={3320--3328},
  year={2014}
}

@inproceedings{huang2011adversarial,
  title={Adversarial machine learning},
  author={Huang, Ling and Joseph, Anthony D and Nelson, Blaine and Rubinstein, Benjamin IP and Tygar, J Doug},
  booktitle={Proceedings of the 4th ACM workshop on Security and artificial intelligence},
  pages={43--58},
  year={2011}
}

@inproceedings{liu2018fine,
  title={Fine-pruning: Defending against backdooring attacks on deep neural networks},
  author={Liu, Kang and Dolan-Gavitt, Brendan and Garg, Siddharth},
  booktitle={International Symposium on Research in Attacks, Intrusions, and Defenses},
  pages={273--294},
  year={2018},
  organization={Springer}
}

@article{lecun1998mnist,
  title={The MNIST database of handwritten digits},
  author={LeCun, Yann},
  journal={http://yann. lecun. com/exdb/mnist/},
  year={1998}
}

@inproceedings{merkle1987digital,
  title={A digital signature based on a conventional encryption function},
  author={Merkle, Ralph C},
  booktitle={Conference on the theory and application of cryptographic techniques},
  pages={369--378},
  year={1987},
  organization={Springer}
}

@article{rumelhart1986learning,
  title={Learning representations by back-propagating errors},
  author={Rumelhart, David E and Hinton, Geoffrey E and Williams, Ronald J},
  journal={nature},
  volume={323},
  number={6088},
  pages={533--536},
  year={1986},
  publisher={Nature Publishing Group}
}

@article{dean2012large,
  title={Large scale distributed deep networks},
  author={Dean, Jeffrey and Corrado, Greg and Monga, Rajat and Chen, Kai and Devin, Matthieu and Mao, Mark and Ranzato, Marc'aurelio and Senior, Andrew and Tucker, Paul and Yang, Ke and others},
  journal={Advances in neural information processing systems},
  volume={25},
  pages={1223--1231},
  year={2012}
}

@incollection{lamport2019byzantine,
  title={The Byzantine generals problem},
  author={Lamport, Leslie and Shostak, Robert and Pease, Marshall},
  booktitle={Concurrency: the Works of Leslie Lamport},
  pages={203--226},
  year={2019}
}

@inproceedings{blanchard2017machine,
  title={Machine learning with adversaries: Byzantine tolerant gradient descent},
  author={Blanchard, Peva and Guerraoui, Rachid and Stainer, Julien and others},
  booktitle={Advances in Neural Information Processing Systems},
  pages={119--129},
  year={2017}
}

@article{biggio2012poisoning,
  title={Poisoning attacks against support vector machines},
  author={Biggio, Battista and Nelson, Blaine and Laskov, Pavel},
  journal={arXiv preprint arXiv:1206.6389},
  year={2012}
}

@inproceedings{song2013stochastic,
  title={Stochastic gradient descent with differentially private updates},
  author={Song, Shuang and Chaudhuri, Kamalika and Sarwate, Anand D},
  booktitle={2013 IEEE Global Conference on Signal and Information Processing},
  pages={245--248},
  year={2013},
  organization={IEEE}
}

@article{biggio2018wild,
  title={Wild patterns: Ten years after the rise of adversarial machine learning},
  author={Biggio, Battista and Roli, Fabio},
  journal={Pattern Recognition},
  volume={84},
  pages={317--331},
  year={2018},
  publisher={Elsevier}
}

@inproceedings{papernot2018sok,
  title={SoK: Towards the Science of Security and Privacy in Machine Learning},
  author={Papernot, Nicolas and McDaniel, Patrick and Sinha, Arunesh and Wellman, Michael P},
  booktitle={2018 IEEE European Symposium on Security and Privacy (EuroS\&P)},
  year={2018},
  organization={IEEE}
}

@article{szegedy2013intriguing,
  title={Intriguing properties of neural networks},
  author={Szegedy, Christian and Zaremba, Wojciech and Sutskever, Ilya and Bruna, Joan and Erhan, Dumitru and Goodfellow, Ian and Fergus, Rob},
  journal={arXiv preprint arXiv:1312.6199},
  year={2013}
}

@inproceedings{biggio2013evasion,
  title={Evasion attacks against machine learning at test time},
  author={Biggio, Battista and Corona, Igino and Maiorca, Davide and Nelson, Blaine and {\v{S}}rndi{\'c}, Nedim and Laskov, Pavel and Giacinto, Giorgio and Roli, Fabio},
  booktitle={Joint European conference on machine learning and knowledge discovery in databases},
  pages={387--402},
  year={2013},
  organization={Springer}
}

@inproceedings{putnam2014reconfigurable,
  title={A reconfigurable fabric for accelerating large-scale datacenter services},
  author={Putnam, Andrew and Caulfield, Adrian M and Chung, Eric S and Chiou, Derek and Constantinides, Kypros and Demme, John and Esmaeilzadeh, Hadi and Fowers, Jeremy and Gopal, Gopi Prashanth and Gray, Jan and others},
  booktitle={2014 ACM/IEEE 41st International Symposium on Computer Architecture (ISCA)},
  pages={13--24},
  year={2014},
  organization={IEEE}
}

@inproceedings{ishai2007efficient,
  title={Efficient arguments without short PCPs},
  author={Ishai, Yuval and Kushilevitz, Eyal and Ostrovsky, Rafail},
  booktitle={Twenty-Second Annual IEEE Conference on Computational Complexity (CCC'07)},
  pages={278--291},
  year={2007},
  organization={IEEE}
}

@inproceedings{tan2017efficient,
  title={The efficient server audit problem, deduplicated re-execution, and the web},
  author={Tan, Cheng and Yu, Lingfan and Leners, Joshua B and Walfish, Michael},
  booktitle={Proceedings of the 26th Symposium on Operating Systems Principles},
  pages={546--564},
  year={2017}
}

@inproceedings{setty2012taking,
  title={Taking proof-based verified computation a few steps closer to practicality},
  author={Setty, Srinath and Vu, Victor and Panpalia, Nikhil and Braun, Benjamin and Blumberg, Andrew J and Walfish, Michael},
  booktitle={Presented as part of the 21st $\{$USENIX$\}$ Security Symposium ($\{$USENIX$\}$ Security 12)},
  pages={253--268},
  year={2012}
}

@inproceedings{braun2013verifying,
  title={Verifying computations with state},
  author={Braun, Benjamin and Feldman, Ariel J and Ren, Zuocheng and Setty, Srinath and Blumberg, Andrew J and Walfish, Michael},
  booktitle={Proceedings of the Twenty-Fourth ACM Symposium on Operating Systems Principles},
  pages={341--357},
  year={2013}
}

@inproceedings{setty2012making,
  title={Making argument systems for outsourced computation practical (sometimes).},
  author={Setty, Srinath TV and McPherson, Richard and Blumberg, Andrew J and Walfish, Michael},
  booktitle={NDSS},
  volume={1},
  number={9},
  pages={17},
  year={2012}
}

@inproceedings{hawblitzel2015ironfleet,
  title={IronFleet: proving practical distributed systems correct},
  author={Hawblitzel, Chris and Howell, Jon and Kapritsos, Manos and Lorch, Jacob R and Parno, Bryan and Roberts, Michael L and Setty, Srinath and Zill, Brian},
  booktitle={Proceedings of the 25th Symposium on Operating Systems Principles},
  pages={1--17},
  year={2015}
}

@inproceedings{jouppi2017datacenter,
  title={In-datacenter performance analysis of a tensor processing unit},
  author={Jouppi, Norman P and Young, Cliff and Patil, Nishant and Patterson, David and Agrawal, Gaurav and Bajwa, Raminder and Bates, Sarah and Bhatia, Suresh and Boden, Nan and Borchers, Al and others},
  booktitle={Proceedings of the 44th Annual International Symposium on Computer Architecture},
  pages={1--12},
  year={2017}
}

@inproceedings{markidis2018nvidia,
  title={Nvidia tensor core programmability, performance \& precision},
  author={Markidis, Stefano and Der Chien, Steven Wei and Laure, Erwin and Peng, Ivy Bo and Vetter, Jeffrey S},
  booktitle={2018 IEEE International Parallel and Distributed Processing Symposium Workshops (IPDPSW)},
  pages={522--531},
  year={2018},
  organization={IEEE}
}

@inproceedings{li2014scaling,
  title={Scaling distributed machine learning with the parameter server},
  author={Li, Mu and Andersen, David G and Park, Jun Woo and Smola, Alexander J and Ahmed, Amr and Josifovski, Vanja and Long, James and Shekita, Eugene J and Su, Bor-Yiing},
  booktitle={11th $\{$USENIX$\}$ Symposium on Operating Systems Design and Implementation ($\{$OSDI$\}$ 14)},
  pages={583--598},
  year={2014}
}

@article{walfish2015verifying,
  title={Verifying computations without reexecuting them},
  author={Walfish, Michael and Blumberg, Andrew J},
  journal={Communications of the ACM},
  volume={58},
  number={2},
  pages={74--84},
  year={2015},
  publisher={ACM New York, NY, USA}
}

@inproceedings{glorot2011deep,
  title={Deep sparse rectifier neural networks},
  author={Glorot, Xavier and Bordes, Antoine and Bengio, Yoshua},
  booktitle={Proceedings of the fourteenth international conference on artificial intelligence and statistics},
  pages={315--323},
  year={2011}
}

@article{vCNN,
  author    = {Seunghwa Lee and
               Hankyung Ko and
               Jihye Kim and
               Hyunok Oh},
  title     = {vCNN: Verifiable Convolutional Neural Network},
  journal   = {{IACR} Cryptol. ePrint Arch.},
  volume    = {2020},
  pages     = {584},
  year      = {2020},
  url       = {https://eprint.iacr.org/2020/584},
  bibsource = {dblp computer science bibliography, https://dblp.org}
}

@article{DBLP:journals/corr/abs-1912-02919,
  author    = {Stephanie L. Hyland and
               Shruti Tople},
  title     = {On the Intrinsic Privacy of Stochastic Gradient Descent},
  journal   = {CoRR},
  volume    = {abs/1912.02919},
  year      = {2019},
  url       = {http://arxiv.org/abs/1912.02919},
  archivePrefix = {arXiv},
  eprint    = {1912.02919},
  bibsource = {dblp computer science bibliography, https://dblp.org}
}

@InProceedings{He_2015_ICCV,
author = {He, Kaiming and Zhang, Xiangyu and Ren, Shaoqing and Sun, Jian},
title = {Delving Deep into Rectifiers: Surpassing Human-Level Performance on ImageNet Classification},
booktitle = {Proceedings of the IEEE International Conference on Computer Vision (ICCV)},
month = {December},
year = {2015}
}

@inproceedings{paszke2019pytorch,
  title={Pytorch: An imperative style, high-performance deep learning library},
  author={Paszke, Adam and Gross, Sam and Massa, Francisco and Lerer, Adam and Bradbury, James and Chanan, Gregory and Killeen, Trevor and Lin, Zeming and Gimelshein, Natalia and Antiga, Luca and others},
  booktitle={Advances in neural information processing systems},
  pages={8026--8037},
  year={2019}
}

@article{frankle2018lottery,
  title={The lottery ticket hypothesis: Finding sparse, trainable neural networks},
  author={Frankle, Jonathan and Carbin, Michael},
  journal={arXiv preprint arXiv:1803.03635},
  year={2018}
}

@article{chetlur2014cudnn,
  title={cudnn: Efficient primitives for deep learning},
  author={Chetlur, Sharan and Woolley, Cliff and Vandermersch, Philippe and Cohen, Jonathan and Tran, John and Catanzaro, Bryan and Shelhamer, Evan},
  journal={arXiv preprint arXiv:1410.0759},
  year={2014}
}

@inproceedings{abadi2016tensorflow,
  title={Tensorflow: A system for large-scale machine learning},
  author={Abadi, Mart{\'\i}n and Barham, Paul and Chen, Jianmin and Chen, Zhifeng and Davis, Andy and Dean, Jeffrey and Devin, Matthieu and Ghemawat, Sanjay and Irving, Geoffrey and Isard, Michael and others},
  booktitle={12th $\{$USENIX$\}$ symposium on operating systems design and implementation ($\{$OSDI$\}$ 16)},
  pages={265--283},
  year={2016}
}

@article{robbins1951stochastic,
  title={A stochastic approximation method},
  author={Robbins, Herbert and Monro, Sutton},
  journal={The annals of mathematical statistics},
  pages={400--407},
  year={1951},
  publisher={JSTOR}
}

@inproceedings{tramer2016stealing,
  title={Stealing machine learning models via prediction apis},
  author={Tram{\`e}r, Florian and Zhang, Fan and Juels, Ari and Reiter, Michael K and Ristenpart, Thomas},
  booktitle={25th $\{$USENIX$\}$ Security Symposium ($\{$USENIX$\}$ Security 16)},
  pages={601--618},
  year={2016}
}

@book{vapnik2013nature,
  title={The nature of statistical learning theory},
  author={Vapnik, Vladimir},
  year={2013},
  publisher={Springer science \& business media}
}

@book{goodfellow2016deep,
  title={Deep learning},
  author={Goodfellow, Ian and Bengio, Yoshua and Courville, Aaron and Bengio, Yoshua},
  volume={1},
  number={2},
  year={2016},
  publisher={MIT press Cambridge}
}

@inproceedings{NIPS2017_5d44ee6f,
 author = {Klambauer, G\"{u}nter and Unterthiner, Thomas and Mayr, Andreas and Hochreiter, Sepp},
 booktitle = {Advances in Neural Information Processing Systems},
 editor = {I. Guyon and U. V. Luxburg and S. Bengio and H. Wallach and R. Fergus and S. Vishwanathan and R. Garnett},
 pages = {971--980},
 publisher = {Curran Associates, Inc.},
 title = {Self-Normalizing Neural Networks},
 url = {https://proceedings.neurips.cc/paper/2017/file/5d44ee6f2c3f71b73125876103c8f6c4-Paper.pdf},
 volume = {30},
 year = {2017}
}

@inproceedings{glorot2010understanding,
  title={Understanding the difficulty of training deep feedforward neural networks},
  author={Glorot, Xavier and Bengio, Yoshua},
  booktitle={Proceedings of the thirteenth international conference on artificial intelligence and statistics},
  pages={249--256},
  year={2010}
}

@article{embedded_proofs,
  author    = {Herv{\'{e}} Chabanne and
               Julien Keuffer and
               Refik Molva},
  title     = {Embedded Proofs for Verifiable Neural Networks},
  journal   = {{IACR} Cryptol. ePrint Arch.},
  volume    = {2017},
  pages     = {1038},
  year      = {2017},
  url       = {http://eprint.iacr.org/2017/1038},
  bibsource = {dblp computer science bibliography, https://dblp.org}
}

@misc{zande,
           title = {Leveraging zero-knowledge succinct arguments of knowledge for efficient verification of outsourced training of artificial neural networks},
          author = {M.J. van de {Zande}},
            year = {2019},
           month = {May},
             url = {http://essay.utwente.nl/79180/}
}

@article{ghodsi_safetynets_nodate,
	title = {{SafetyNets}: Verifiable Execution of Deep Neural Networks on an Untrusted Cloud},
	pages = {10},
	author = {Ghodsi, Zahra and Gu, Tianyu and Garg, Siddharth},
	langid = {english}
}

@online{li__chuan_openais_2020,
	title = {{OpenAI}'s {GPT}-3 Language Model: A Technical Overview},
	url = {https://lambdalabs.com/blog/demystifying-gpt-3/},
	shorttitle = {{OpenAI}'s {GPT}-3 Language Model},
	titleaddon = {Lambda Blog},
	author = {Li,  Chuan},
	urldate = {2020-10-01},
	date = {2020-06-03},
	langid = {english},
	note = {Library Catalog: lambdalabs.com}
}

@incollection{preneel_proofs_1999,
	location = {Boston, {MA}},
	title = {Proofs of Work and Bread Pudding Protocols(Extended Abstract)},
	isbn = {978-1-4757-6487-1 978-0-387-35568-9},
	url = {http://link.springer.com/10.1007/978-0-387-35568-9_18},
	pages = {258--272},
	booktitle = {Secure Information Networks},
	publisher = {Springer {US}},
	author = {Jakobsson, Markus and Juels, Ari},
	editor = {Preneel, Bart},
	urldate = {2020-11-09},
	date = {1999},
	langid = {english},
	doi = {10.1007/978-0-387-35568-9_18}
}

@article{2016arXiv160202697P,
       author = {{Papernot}, Nicolas and {McDaniel}, Patrick and {Goodfellow}, Ian and
         {Jha}, Somesh and {Berkay Celik}, Z. and {Swami}, Ananthram},
        title = "{Practical Black-Box Attacks against Machine Learning}",
      journal = {arXiv e-prints},
         year = "2016",
        month = "Feb",
          eid = {arXiv:1602.02697},
        pages = {arXiv:1602.02697},
archivePrefix = {arXiv},
       eprint = {1602.02697},
 primaryClass = {cs.CR},
       adsurl = {https://ui.adsabs.harvard.edu/abs/2016arXiv160202697P}
}

@article{pal,
  author    = {Soham Pal and
               Yash Gupta and
               Aditya Shukla and
               Aditya Kanade and
               Shirish K. Shevade and
               Vinod Ganapathy},
  title     = {A framework for the extraction of Deep Neural Networks by leveraging
               public data},
  journal   = {CoRR},
  volume    = {abs/1905.09165},
  year      = {2019},
  url       = {http://arxiv.org/abs/1905.09165},
  archivePrefix = {arXiv},
  eprint    = {1905.09165},
  bibsource = {dblp computer science bibliography, https://dblp.org}
}

@inproceedings{copycat,
  title={Copycat CNN: Stealing Knowledge by Persuading Confession with Random Non-Labeled Data},
  author={Correia-Silva, Jacson Rodrigues and Berriel, Rodrigo F and Badue, Claudine and de Souza, Alberto F and Oliveira-Santos, Thiago},
  booktitle={2018 International Joint Conference on Neural Networks (IJCNN)},
  pages={1--8},
  year={2018},
  organization={IEEE}
}

@inproceedings{knockoff,
  title={Knockoff nets: Stealing functionality of black-box models},
  author={Orekondy, Tribhuvanesh and Schiele, Bernt and Fritz, Mario},
  booktitle={Proceedings of the IEEE Conference on Computer Vision and Pattern Recognition},
  pages={4954--4963},
  year={2019}
}

@inproceedings {sidechannel,
author = {Lejla Batina and Shivam Bhasin and Dirmanto Jap and Stjepan Picek},
title = {{CSI} {NN}: Reverse Engineering of Neural Network Architectures Through Electromagnetic Side Channel},
booktitle = {28th {USENIX} Security Symposium ({USENIX} Security 19)},
year = {2019},
isbn = {978-1-939133-06-9},
address = {Santa Clara, CA},
pages = {515--532},
url = {https://www.usenix.org/conference/usenixsecurity19/presentation/batina},
publisher = {{USENIX} Association},
month = aug,
}

@ARTICLE{2019arXiv190901838J,
       author = {{Jagielski}, Matthew and {Carlini}, Nicholas and {Berthelot}, David and
         {Kurakin}, Alex and {Papernot}, Nicolas},
        title = "{High-Fidelity Extraction of Neural Network Models}",
      journal = {arXiv e-prints},
         year = "2019",
        month = "Sep",
          eid = {arXiv:1909.01838},
        pages = {arXiv:1909.01838},
archivePrefix = {arXiv},
       eprint = {1909.01838},
 primaryClass = {cs.LG},
       adsurl = {https://ui.adsabs.harvard.edu/abs/2019arXiv190901838J}
}

@ARTICLE{watermarksurvey,
       author = {{Boenisch}, Franziska},
        title = "{A Survey on Model Watermarking Neural Networks}",
      journal = {arXiv e-prints},
         year = 2020,
        month = sep,
          eid = {arXiv:2009.12153},
        pages = {arXiv:2009.12153},
archivePrefix = {arXiv},
       eprint = {2009.12153},
 primaryClass = {cs.CR},
       adsurl = {https://ui.adsabs.harvard.edu/abs/2020arXiv200912153B}
}

@article{varun,
  author    = {Varun Chandrasekaran and
               Kamalika Chaudhuri and
               Irene Giacomelli and
               Somesh Jha and
               Songbai Yan},
  title     = {Model Extraction and Active Learning},
  journal   = {CoRR},
  volume    = {abs/1811.02054},
  year      = {2018},
  url       = {http://arxiv.org/abs/1811.02054},
  archivePrefix = {arXiv},
  eprint    = {1811.02054},
  bibsource = {dblp computer science bibliography, https://dblp.org}
}

@article{lee2018defending,
  title={Defending Against Machine Learning Model Stealing Attacks Using Deceptive Perturbations},
  author={Lee, Taesung and Edwards, Benjamin and Molloy, Ian and Su, Dong},
  journal={arXiv preprint arXiv:1806.00054},
  year={2018}
}

@inproceedings{alabdulmohsin2014adding,
  title={Adding robustness to support vector machines against adversarial reverse engineering},
  author={Alabdulmohsin, Ibrahim M and Gao, Xin and Zhang, Xiangliang},
  booktitle={Proceedings of the 23rd ACM International Conference on Conference on Information and Knowledge Management},
  pages={231--240},
  year={2014}
}

@ARTICLE{2020arXiv200212200J,
       author = {{Jia}, Hengrui and {Choquette-Choo}, Christopher A. and {Papernot}, Nicolas},
        title = "{Entangled Watermarks as a Defense against Model Extraction}",
      journal = {arXiv e-prints},
         year = 2020,
        month = feb,
          eid = {arXiv:2002.12200},
        pages = {arXiv:2002.12200},
archivePrefix = {arXiv},
       eprint = {2002.12200},
 primaryClass = {cs.CR},
       adsurl = {https://ui.adsabs.harvard.edu/abs/2020arXiv200212200J}
}

@ARTICLE{2019arXiv190106151N,
       author = {{Namba}, Ryota and {Sakuma}, Jun},
        title = "{Robust Watermarking of Neural Network with Exponential Weighting}",
      journal = {arXiv e-prints},
         year = 2019,
        month = jan,
          eid = {arXiv:1901.06151},
        pages = {arXiv:1901.06151},
archivePrefix = {arXiv},
       eprint = {1901.06151},
 primaryClass = {cs.CR},
       adsurl = {https://ui.adsabs.harvard.edu/abs/2019arXiv190106151N}
}

@ARTICLE{2019arXiv191001226L,
       author = {{Li}, Huiying and {Wenger}, Emily and {Zhao}, Ben Y. and {Zheng}, Haitao},
        title = "{Piracy Resistant Watermarks for Deep Neural Networks}",
      journal = {arXiv e-prints},
         year = 2019,
        month = oct,
          eid = {arXiv:1910.01226},
        pages = {arXiv:1910.01226},
archivePrefix = {arXiv},
       eprint = {1910.01226},
 primaryClass = {cs.CR},
       adsurl = {https://ui.adsabs.harvard.edu/abs/2019arXiv191001226L}
}

@inproceedings{cifar,
  title={Learning Multiple Layers of Features from Tiny Images},
  author={Alex Krizhevsky},
  year={2009}
}

@ARTICLE{resnet,
author = {{He}, Kaiming and {Zhang}, Xiangyu and {Ren}, Shaoqing and {Sun}, Jian},
title = "{Deep Residual Learning for Image Recognition}",
journal = {arXiv e-prints},
year = 2015,
month = dec,
eid = {arXiv:1512.03385},
pages = {arXiv:1512.03385},
archivePrefix = {arXiv},
eprint = {1512.03385},
primaryClass = {cs.CV},
adsurl = {https://ui.adsabs.harvard.edu/abs/2015arXiv151203385H}
}

@inproceedings{dworknaor1992,
author = {Dwork, Cynthia and Naor, Moni},
title = {Pricing via Processing or Combatting Junk Mail},
year = {1992},
isbn = {3540573402},
publisher = {Springer-Verlag},
address = {Berlin, Heidelberg},
booktitle = {Proceedings of the 12th Annual International Cryptology Conference on Advances in Cryptology},
pages = {139–147},
numpages = {9},
series = {CRYPTO '92}
}

@ARTICLE{orthogonal,
       author = {{Saxe}, Andrew M. and {McClelland}, James L. and {Ganguli}, Surya},
        title = "{Exact solutions to the nonlinear dynamics of learning in deep linear neural networks}",
      journal = {arXiv e-prints},
         year = 2013,
        month = dec,
          eid = {arXiv:1312.6120},
        pages = {arXiv:1312.6120},
archivePrefix = {arXiv},
       eprint = {1312.6120},
 primaryClass = {cs.NE},
       adsurl = {https://ui.adsabs.harvard.edu/abs/2013arXiv1312.6120S}
}

@article{kstest,
author = { Frank J.   Massey   Jr. },
title = {The Kolmogorov-Smirnov Test for Goodness of Fit},
journal = {Journal of the American Statistical Association},
volume = {46},
number = {253},
pages = {68-78},
year  = {1951},
publisher = {Taylor & Francis},
doi = {10.1080/01621459.1951.10500769},

URL = { 
        https://www.tandfonline.com/doi/abs/10.1080/01621459.1951.10500769
    
},
eprint = { 
        https://www.tandfonline.com/doi/pdf/10.1080/01621459.1951.10500769
    
}
}

@misc{hokkaido,
    author = {Fabien Coelho},
    title = {Exponential Memory-Bound Functions for Proof of Work Protocols},
    howpublished = {Cryptology ePrint Archive, Report 2005/356},
    year = {2005},
    note = {\url{https://eprint.iacr.org/2005/356}},
}

@misc{merkle_tree_based_pow,
    author = {Fabien Coelho},
    title = {An (Almost) Constant-Effort Solution-Verification Proof-of-Work Protocol based on Merkle Trees},
    howpublished = {Cryptology ePrint Archive, Report 2007/433},
    year = {2007},
    note = {\url{https://eprint.iacr.org/2007/433}},
}

@article{moderate_hard_mbound,
author = {Abadi, Martín and Burrows, Michael and Manasse, Mark and Wobber, Ted},
year = {2005},
month = {05},
pages = {299-327},
title = {Moderately Hard, Memory-bound Functions},
volume = {5},
journal = {ACM Trans. Internet Techn.},
doi = {10.1145/1064340.1064341}
}

@inproceedings{cuckoo_cycle,
author = {Tromp, John},
year = {2015},
month = {01},
pages = {49-62},
title = {Cuckoo Cycle: A Memory Bound Graph-Theoretic Proof-of-Work},
volume = {8976},
isbn = {978-3-662-48050-2},
doi = {10.1007/978-3-662-48051-9_4}
}

@INPROCEEDINGS{guided_tour,
  author={M. {Abliz} and T. {Znati}},
  booktitle={2009 Annual Computer Security Applications Conference}, 
  title={A Guided Tour Puzzle for Denial of Service Prevention}, 
  year={2009},
  volume={},
  number={},
  pages={279-288},
  doi={10.1109/ACSAC.2009.33}}

@inproceedings{diffiehellman,
author = {Waters, Brent and Juels, Ari and Halderman, J. and Felten, Edward},
year = {2004},
month = {01},
pages = {246-256},
title = {New client puzzle outsourcing techniques for DoS resistance},
journal = {Proceedings of the ACM Conference on Computer and Communications Security},
doi = {10.1145/1030083.1030117}
}

@misc{Dua:2019 ,
author = "Dua, Dheeru and Graff, Casey",
year = "2017",
title = "{UCI} Machine Learning Repository",
url = "http://archive.ics.uci.edu/ml",
institution = "University of California, Irvine, School of Information and Computer Sciences" }

@ARTICLE{2020SciPy-NMeth,
  author  = {Virtanen, Pauli and Gommers, Ralf and Oliphant, Travis E. and
            Haberland, Matt and Reddy, Tyler and Cournapeau, David and
            Burovski, Evgeni and Peterson, Pearu and Weckesser, Warren and
            Bright, Jonathan and {van der Walt}, St{\'e}fan J. and
            Brett, Matthew and Wilson, Joshua and Millman, K. Jarrod and
            Mayorov, Nikolay and Nelson, Andrew R. J. and Jones, Eric and
            Kern, Robert and Larson, Eric and Carey, C J and
            Polat, {\.I}lhan and Feng, Yu and Moore, Eric W. and
            {VanderPlas}, Jake and Laxalde, Denis and Perktold, Josef and
            Cimrman, Robert and Henriksen, Ian and Quintero, E. A. and
            Harris, Charles R. and Archibald, Anne M. and
            Ribeiro, Ant{\^o}nio H. and Pedregosa, Fabian and
            {van Mulbregt}, Paul and {SciPy 1.0 Contributors}},
  title   = {{{SciPy} 1.0: Fundamental Algorithms for Scientific
            Computing in Python}},
  journal = {Nature Methods},
  year    = {2020},
  volume  = {17},
  pages   = {261--272},
  adsurl  = {https://rdcu.be/b08Wh},
  doi     = {10.1038/s41592-019-0686-2},
}

@article{hashcash,
author = {Back, Adam},
year = {2002},
month = {09},
pages = {},
title = {Hashcash - A Denial of Service Counter-Measure}
}

@article{bitcoin,
author = {Nakamoto, Satoshi},
year = {2009},
month = {03},
pages = {},
title = {Bitcoin: A Peer-to-Peer Electronic Cash System},
journal = {Cryptography Mailing list at https://metzdowd.com}
}

@article{yin2018stochastic,
  title={Stochastic backward Euler: an implicit gradient descent algorithm for k-means clustering},
  author={Yin, Penghang and Pham, Minh and Oberman, Adam and Osher, Stanley},
  journal={Journal of Scientific Computing},
  volume={77},
  number={2},
  pages={1133--1146},
  year={2018},
  publisher={Springer}
}

@article{claessen2013splittable,
  title={Splittable pseudorandom number generators using cryptographic hashing},
  author={Claessen, Koen and Pa{\l}ka, Micha{\l} H},
  journal={ACM SIGPLAN Notices},
  volume={48},
  number={12},
  pages={47--58},
  year={2013},
  publisher={ACM New York, NY, USA}
}

@inproceedings{salmon2011parallel,
  title={Parallel random numbers: as easy as 1, 2, 3},
  author={Salmon, John K and Moraes, Mark A and Dror, Ron O and Shaw, David E},
  booktitle={Proceedings of 2011 International Conference for High Performance Computing, Networking, Storage and Analysis},
  pages={1--12},
  year={2011}
}

@INPROCEEDINGS{secureml,
  author={P. {Mohassel} and Y. {Zhang}},
  booktitle={2017 IEEE Symposium on Security and Privacy (SP)}, 
  title={SecureML: A System for Scalable Privacy-Preserving Machine Learning}, 
  year={2017},
  volume={},
  number={},
  pages={19-38},
  doi={10.1109/SP.2017.12}}

@article{shannon1948mathematical,
  title={A mathematical theory of communication},
  author={Shannon, Claude E},
  journal={The Bell system technical journal},
  volume={27},
  number={3},
  pages={379--423},
  year={1948},
  publisher={Nokia Bell Labs}
}

@online{pytorch_reproducibility,
	title = {Reproducibility — {PyTorch} 1.7.0 documentation},
	url = {https://pytorch.org/docs/stable/notes/randomness.html},
	urldate = {2020-12-04}
}

@inproceedings{lecun1995comparison,
  title={Comparison of learning algorithms for handwritten digit recognition},
  author={LeCun, Yann and Jackel, LD and Bottou, Leon and Brunot, A and Cortes, Corinna and Denker, J and Drucker, Harris and Guyon, I and Muller, UA and Sackinger, Eduard and others},
  booktitle={International conference on artificial neural networks},
  volume={60},
  pages={53--60},
  year={1995},
  organization={Perth, Australia}
}

@inproceedings{borisrolnick,
 author = {Hanin, Boris and Rolnick, David},
 booktitle = {Advances in Neural Information Processing Systems},
 editor = {S. Bengio and H. Wallach and H. Larochelle and K. Grauman and N. Cesa-Bianchi and R. Garnett},
 pages = {571--581},
 publisher = {Curran Associates, Inc.},
 title = {How to Start Training: The Effect of Initialization and Architecture},
 url = {https://proceedings.neurips.cc/paper/2018/file/d81f9c1be2e08964bf9f24b15f0e4900-Paper.pdf},
 volume = {31},
 year = {2018}
}

@book{wiley_informationtheory,
author = {Cover, Thomas M. and Thomas, Joy A.},
title = {Elements of Information Theory (Wiley Series in Telecommunications and Signal Processing)},
year = {2006},
isbn = {0471241954},
publisher = {Wiley-Interscience},
address = {USA}
}

@inproceedings {watermarkingdnn,
author = {Yossi Adi and Carsten Baum and Moustapha Cisse and Benny Pinkas and Joseph Keshet},
title = {Turning Your Weakness Into a Strength: Watermarking Deep Neural Networks by Backdooring},
booktitle = {27th {USENIX} Security Symposium ({USENIX} Security 18)},
year = {2018},
isbn = {978-1-939133-04-5},
url = {https://www.usenix.org/conference/usenixsecurity18/presentation/adi},
publisher = {{USENIX} Association},
month = aug,
}

@inproceedings{watermarkingdnn2,
 author = {Zhang, Jialong and Gu, Zhongshu and Jang, Jiyong and Wu, Hui and Stoecklin, M. P. and Huang, H. and Molloy, I.},
 booktitle = {Proceedings of the 2018 on Asia Conference on Computer and Communications Security},
 doi = {10.1145/3196494.3196550},
 title = {Protecting Intellectual Property of Deep Neural Networks with Watermarking},
 url = {https://app.dimensions.ai/details/publication/pub.1104339451},
 year = {2018}
}

@inproceedings{finepruning,
  title={Fine-Pruning: Defending Against Backdooring Attacks on Deep Neural Networks},
  author={K. Liu and Brendan Dolan-Gavitt and Siddharth Garg},
  booktitle={21st International Symposium on Research in Attacks, Intrusions, and Defenses},
  year={2018}
}

@article{neuralcleanse,
  title={Neural Cleanse: Identifying and Mitigating Backdoor Attacks in Neural Networks},
  author={Bolun Wang and Yuanshun Yao and Shawn Shan and Huiying Li and B. Viswanath and H. Zheng and B. Zhao},
  journal={2019 IEEE Symposium on Security and Privacy (SP)},
  year={2019},
  pages={707-723}
}

@ARTICLE{ewe,
       author = {{Jia}, Hengrui and {Choquette-Choo}, Christopher A. and {Chandrasekaran}, Varun and {Papernot}, Nicolas},
        title = "{Entangled Watermarks as a Defense against Model Extraction}",
      journal = {arXiv e-prints},
         year = 2020,
        month = feb,
          eid = {arXiv:2002.12200},
        pages = {arXiv:2002.12200},
archivePrefix = {arXiv},
       eprint = {2002.12200},
 primaryClass = {cs.CR},
       adsurl = {https://ui.adsabs.harvard.edu/abs/2020arXiv200212200J}
}
}

\appendices
\section{Markov Processes and Entropy}
\label{app:markov}
We include additional definitions as they pertain to our proofs in Section~\ref{sec:correctness_analysis}.

\begin{definition}[Markov Process]
A stochastic process is said to have the Markov property if its future is independent of its past, when conditioned on its current state, \ie $Pr(\tilde{W}_{i+1} | \tilde{W}_{0} \hdots \tilde{W}_{i})=Pr(\tilde{W}_{i+1} | \tilde{W}_{i})$. A stochastic process with the Markov property is said to be a Markov process.
\end{definition}

Entropy has many interpretations, but one is the amount of information needed to describe a random variable. We now provide the formal definition based on~\cite{shannon1948mathematical}.

\begin{definition}[Entropy]
~\cite{wiley_informationtheory} For a discrete random variable $X$, its entropy is defined as
\begin{align} \label{eq:entropy}
    H(X) = -\sum_{x\in X} P(x)log_{b}(P(x)).
\end{align}
\end{definition}

\begin{definition}[Cross-Entropy]
For discrete random variables $X$ and $Y$ the cross-entropy of $X$ given $Y$ is defined as
\begin{align}
    H(X|Y) &= -\sum_{x\in X, y\in Y} p(x, y) log_b \frac{P(x,y)}{P(y)}\nonumber\\
    &= -\sum_{x\in X, y\in Y} P(x|Y)log_{b}(P(x|Y))\nonumber\\
    &= H(X,Y) - H(Y)\label{eq:cross-entropy}
\end{align}
\end{definition}

\begin{definition}[Entropy Rate of Stationary Stochastic Process]
Entropy rate of a stationary stochastic process $\{W_i\}$~\cite{wiley_informationtheory} is defined by
\begin{align*}
    H'(W) = \lim_{n \to \infty} H (W_n|W_{n\text{-}1}, W_{n\text{-}2},...,W_1)
\end{align*}
and the limit always exists.
\end{definition}

\begin{definition}[Entropy Rate of Stationary Markov Process]
For a stationary Markov process $\{W_i\}$~\cite{wiley_informationtheory}, the entropy rate is defined as
\begin{align} 
    H'(W) &= \lim_{n \to \infty} H (W_n|W_{n\text{-}1}, W_{n\text{-}2},...,W_1) \nonumber\\ 
    &= \lim_{n \to \infty} H (W_n|W_{n\text{-}1}) \nonumber\\ 
    &= \lim_{n \to \infty} H (W_2|W_1) \nonumber\\ 
    &= H (W_2|W_1)\label{eq:entropy_rate}
\end{align}
\end{definition}

\section{Infeasibility of Directed Retraining}\label{app:infeasibility-directed-retraining}
For an adversary to ensure that weights $W'_t$ converge to the prover's obtained final weights $W_T$, an adversary can directly minimize the difference in their values. This strategy can be easily carried out by regularization, a common strategy in machine learning to limit the effective capacity of a model~\cite{}. To this end, a practitioner will include an additional term in their loss function that minimizes an $l_p$ norm of the weights. An adversary may minimally modify any common regularizer to instead minimize $d(W'_t, W_T)$, as shown in Eq.~(\ref{eqn:total-loss}). This regularizer cannot pass verification because it requires an additional state, consisting of the final model weights $W_T$, that does not pass the KT test and which does not have its own valid \prf. Thus, an adversary may attempt to distill the contained information into a component of the \prf that does not require test: the inputs $x$. However, this strategy cannot be implemented with any gradient-based optimization techniques as no gradient exists (see Eq.~(\ref{eqn:regularization-gradient})). An adversary may attempt to use gradient-free optimization techniques; our experiments show that this requires far more function calls than the training process itself, due to the highly nonlinear relation between $x$ and $\nabla_{f(x)}(\mathcal{L})$.

\begin{equation}
    \label{eqn:total-loss} loss = \mathcal{L}(f_{w_t}(x), y) + d(W'_t, W_T)
\end{equation}
\begin{equation}
    \label{eqn:regularization-gradient} \nabla_x(d(W'_t, W_T)) = 0
\end{equation}

\section{Table of Notations}
\label{sec:notations}

\begin{table}[H]
    \centering
    \begin{tabular}{ll}
        \toprule
        \textbf{Symbol} &  \textbf{Explanation}\\
        \midrule
         \pow &  Proof-of-learning \\
         $\prover$ &  Prover \\
         $\verifier$ & Verifier\\
         $\verifier(.)$ & Verifier $\verifier$'s \textproc{Verify} function\\
         $\model\;/\;\model_\prover$  & Model / of prover $\prover$ \\
         $\dataset\in \mathbb{R}^{n \times d}$ & $d$-dimensional dataset of $n$ samples \\
         $\pw\;/\;\pw[\model]$ & Proof-of-learning / of model $\model$\\
         $W\;/\; W_{t}$ & Model weights / Model weights at step $t$ \\
         $M$ & Meta-data \\
         $\tilde{W} = \{W, M\}$ & Model weights and learning meta-data\\
         $\mathcal{L}$ / $\hat{\mathcal{L}}$ & Loss function / Average loss\\
         \ret & reproduction error of a training step.\\
         \nre & normalized reproduction error.\\
         \rd & average distance between 2 irrelevant models \\
         & with the same architecture and dataset\\
         $c_{i}$ & the random variable that represents the \\
         & number of times data points $i$ is chosen by\\
         &$VerifyEpoch$ in Algorithm~\ref{alg:verification}\\
        \bottomrule
    \end{tabular}
    \caption{Notations}
    \label{tab:notations}
\end{table}

\section{Notes from Spoofing sections}
\begin{figure}[t]
\centering
\includegraphics[width=0.7\columnwidth]{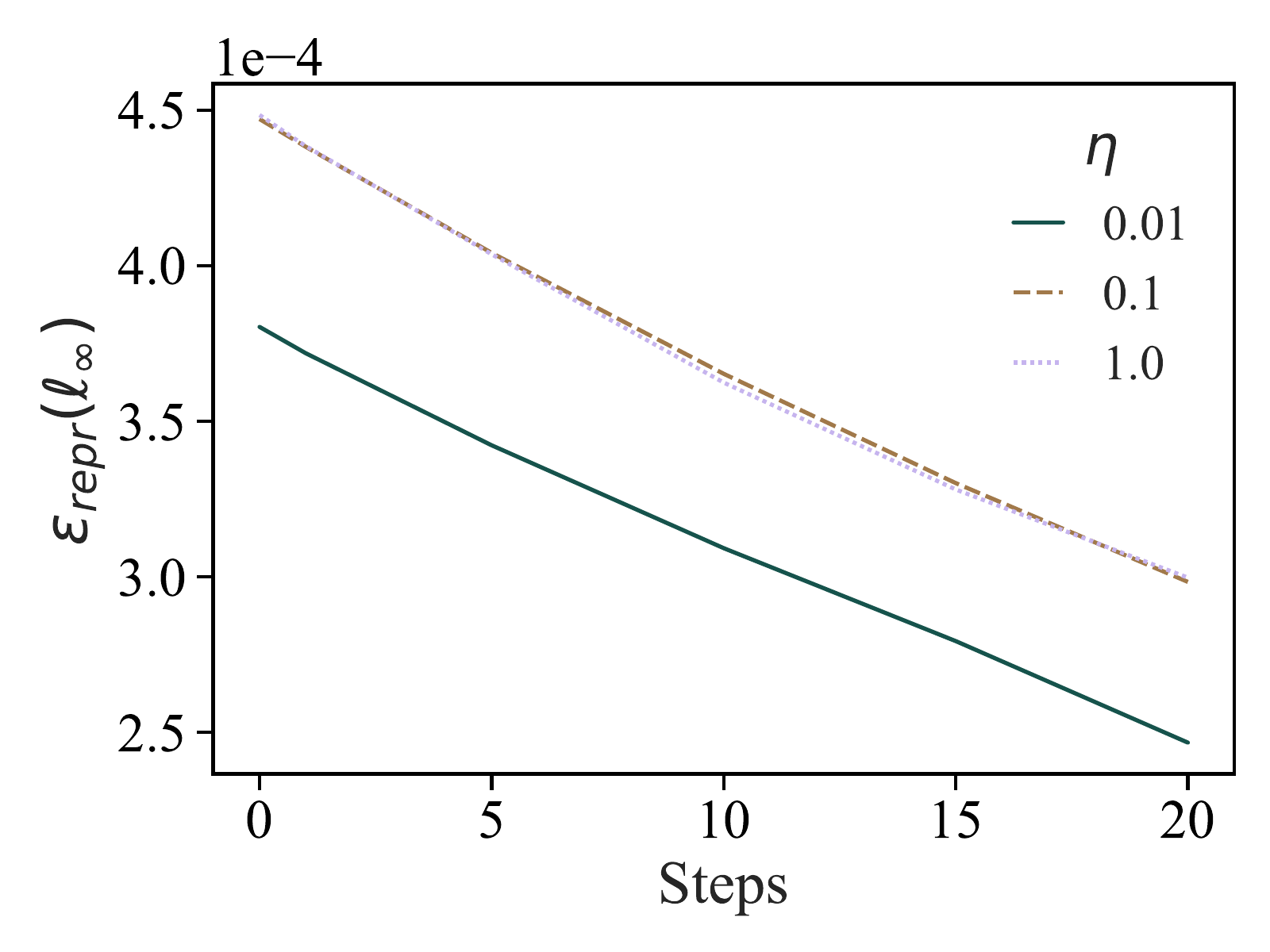}
\caption{Gradient descent has a linear convergence rate when measuring the $l_{\infty}$ norm.}
\label{fig:grad-descent}
\end{figure}

\begin{figure}[t]
\centering
\includegraphics[width=0.75\columnwidth]{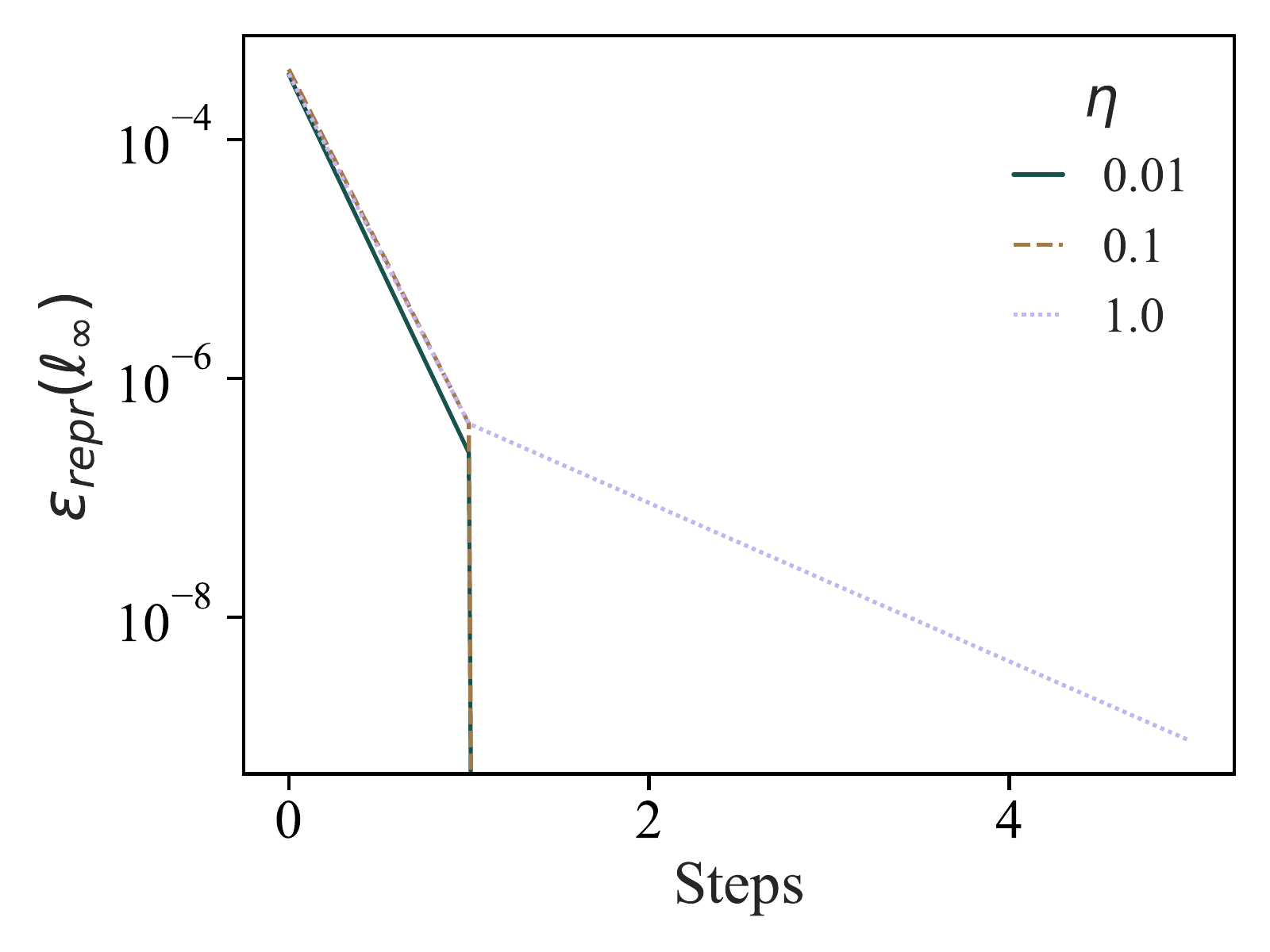}
\caption{Newton Krylov completely converges to $0$ when measuring the $l_{\infty}$ norm.\vspace{-5mm}}
\label{fig:Newton-Krlov}
\end{figure}

\begin{figure}[t]
\centering
\includegraphics[width=0.75\columnwidth]{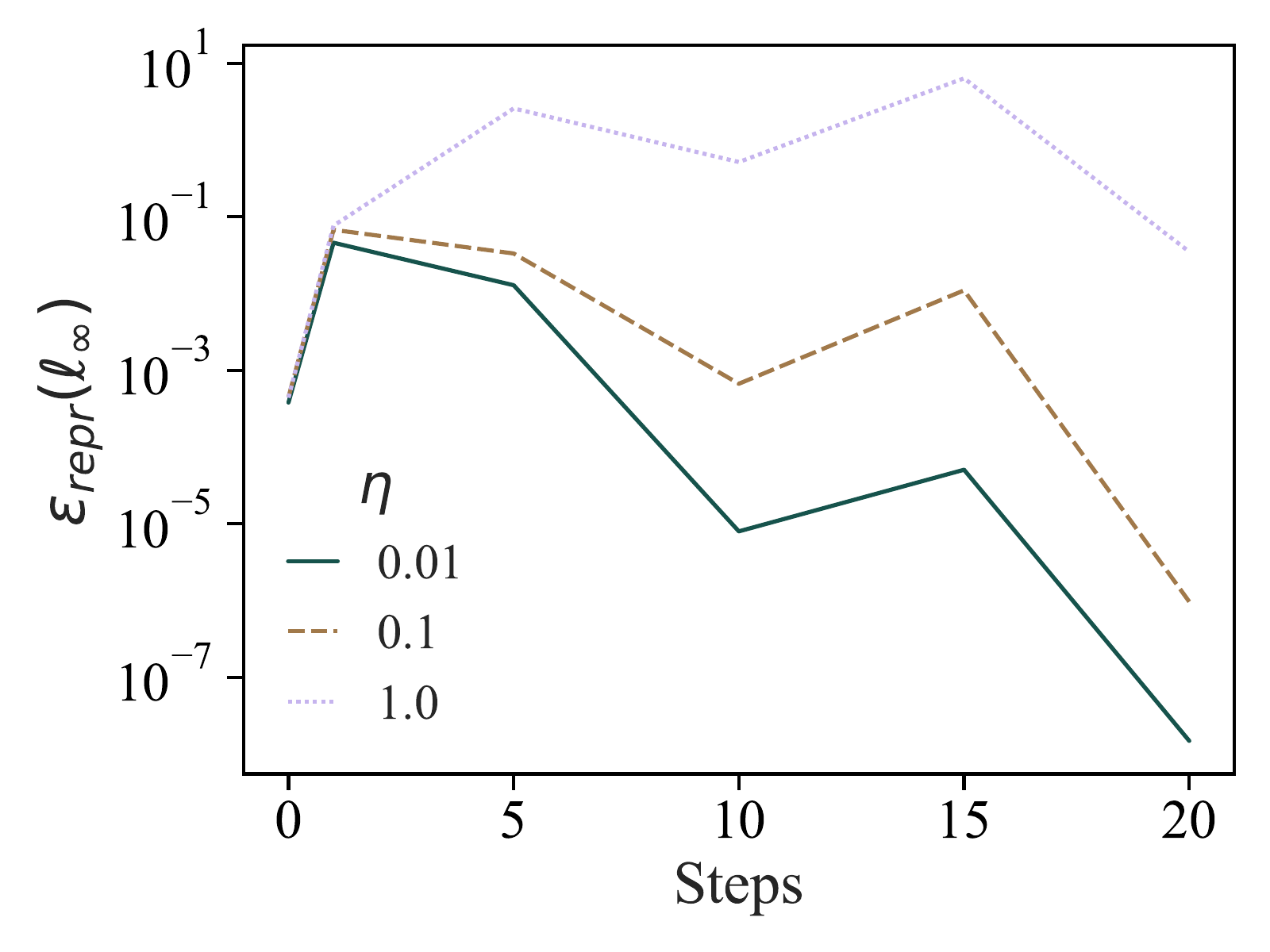}
\caption{Broyden's method converges to below $1e-7$ when measuring the $l_{\infty}$ norm.\vspace{-5mm}}
\label{fig:broyden1}
\end{figure}

\paragraph{Choosing a Root Solver}
We choose three representative root solvers: Gradient Descent on the $l2$-norm, Newton Krylov~\cite{2020SciPy-NMeth}, and Broyden1~\cite{2020SciPy-NMeth} to solve Eq.~\eqref{eqn:root-finding}, \ie find its roots. We use a logistic regression model with $22$ neurons on the Iris dataset~\cite{Dua:2019} and calculated $e_{reproduce}$ (see \S~\ref{ssec:evalulation-metrics}) at each iteration of the root solver. From Figure~\ref{fig:grad-descent},~\ref{fig:Newton-Krlov},~and~\ref{fig:broyden1} we observed that Newton-Krlov performed the best, \ie converged the fastest, and so was the solver we used.

\paragraph{Measuring Computational Complexity of Inverting}
Computational complexity necessarily varies with the method used and implementation of that method, alongside other miscellaneous overhead. However, a lower bound for the computational complexity is simply the number of function calls it takes per step. As computing $\beta(w_{T-1})$ is dominated by computing a training step (+ any overhead), we have that the number of function calls effectively measures how much more computationally intensive inverting is (ex. 5 function calls per step would be at least 5 times as intensive as training). This can be converted to flops by simply taking the flops per backward pass and multiplying that by the number of function calls if comparisons between architectures are needed (in which case the ratio is simply the ratio of flops per backward pass).

\begin{figure}[t]
\centering
\includegraphics[width=0.75\columnwidth]{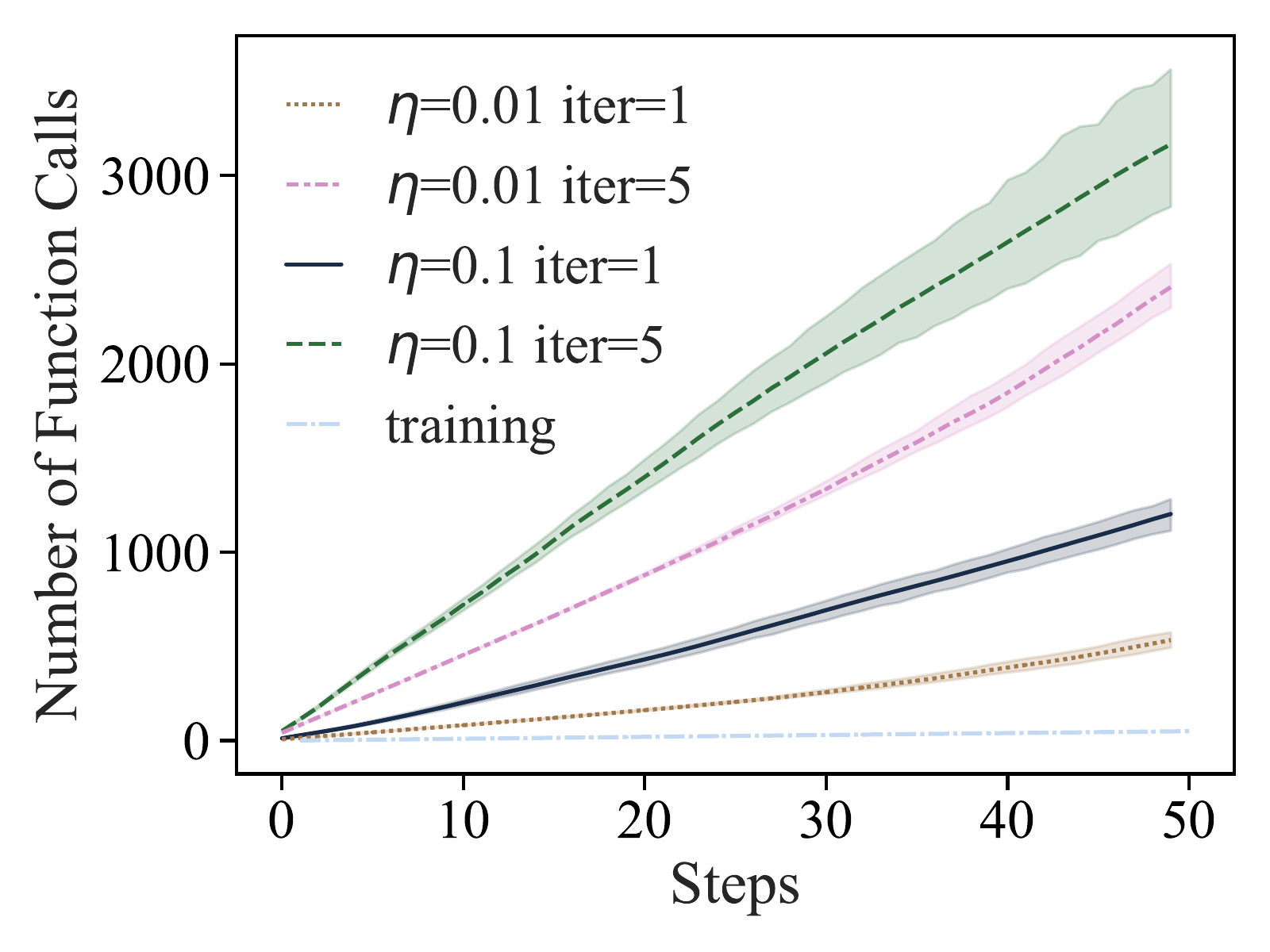}
\caption{Observe that the function calls grow linearly with the steps, and that compared to the baseline of training, they are an order of magnitude steeper.\vspace{-5mm}}
\label{fig:LeNet5-setup-calls}
\end{figure}

\begin{figure}[t]
\centering
\includegraphics[width=0.8\columnwidth]{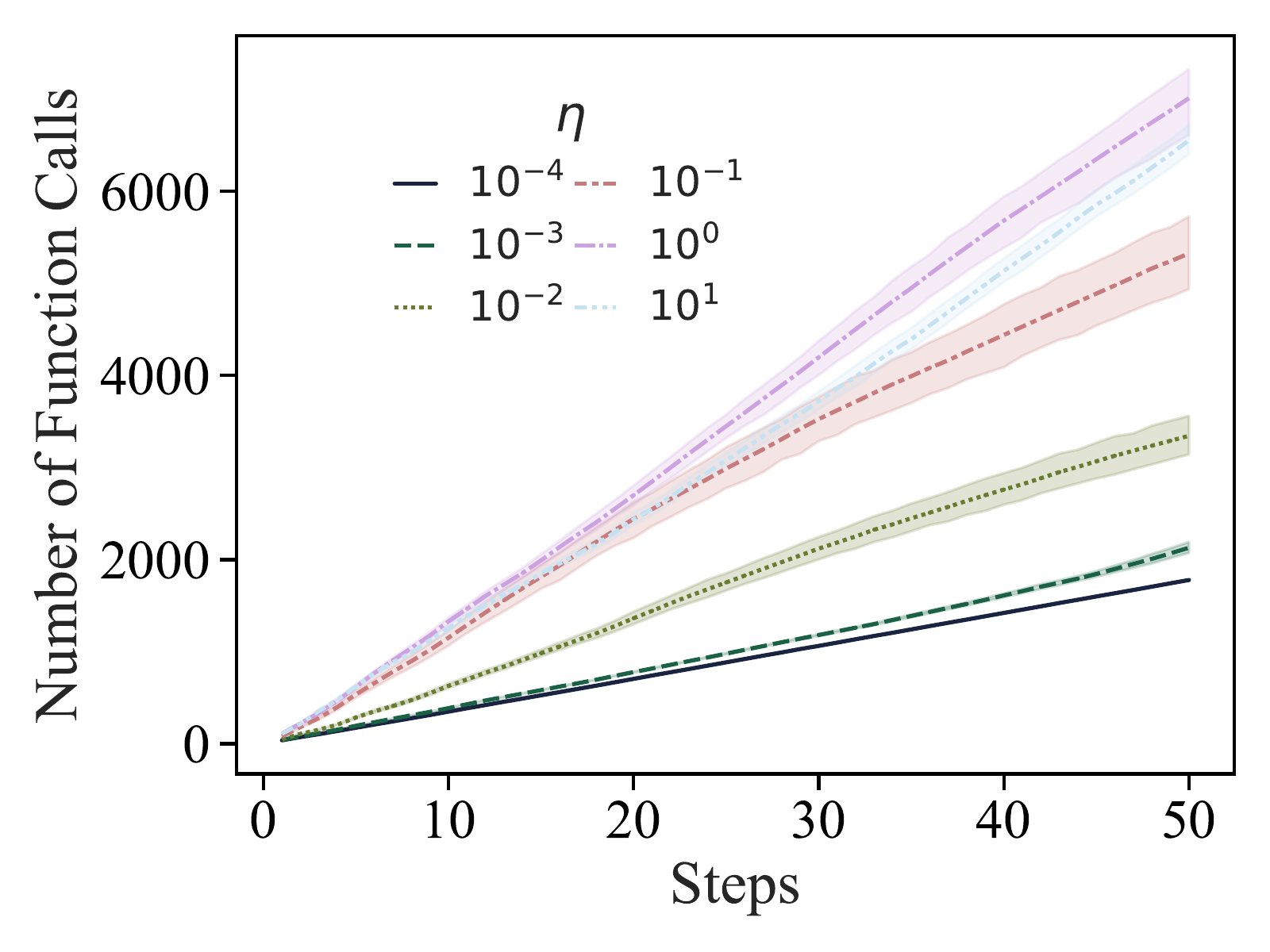}
\caption{Observe that the function calls of all learning rates tested are a magnitude or more larger than the baseline of training, which would be the line $y=x$.\vspace{-5mm}}
 \label{fig:LeNet5_function_calls_lr}
\end{figure}

As seen in Figures~\ref{fig:LeNet5_function_calls_lr} and~\ref{fig:LeNet5-setup-calls}, and noting the baseline for training is simply a $y=x$ line, i.e slope is 1, our current setup is magnitudes more expensive than training. We leave improving this for future work.

\section{Additional Figures and Tables}
\label{app:additional}

\begin{figure}[t]
    \centering
    \subfloat[CIFAR-10]{\includegraphics[width=0.5\linewidth]{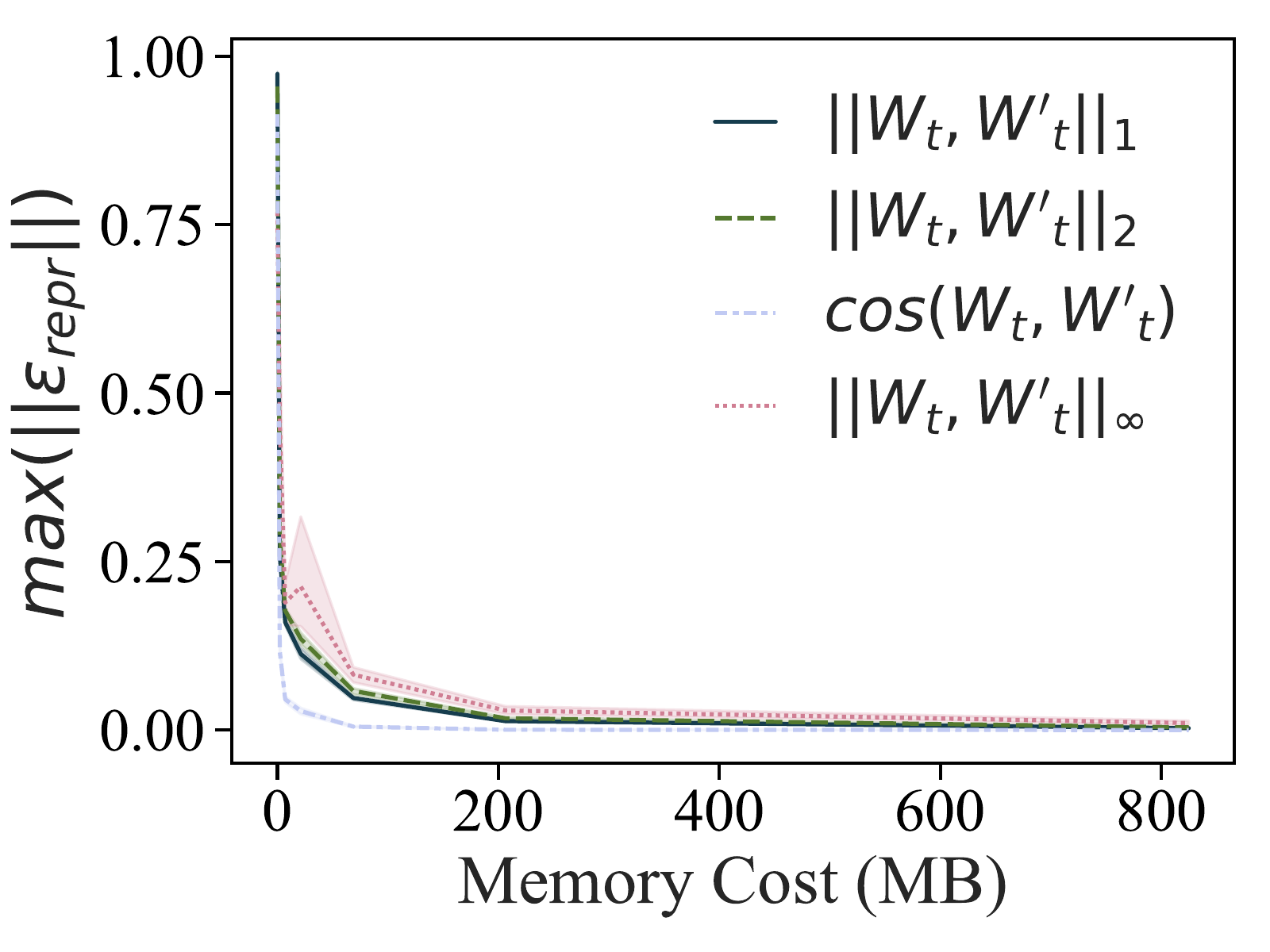}}
    \subfloat[CIFAR-100]{\includegraphics[width=0.5\linewidth]{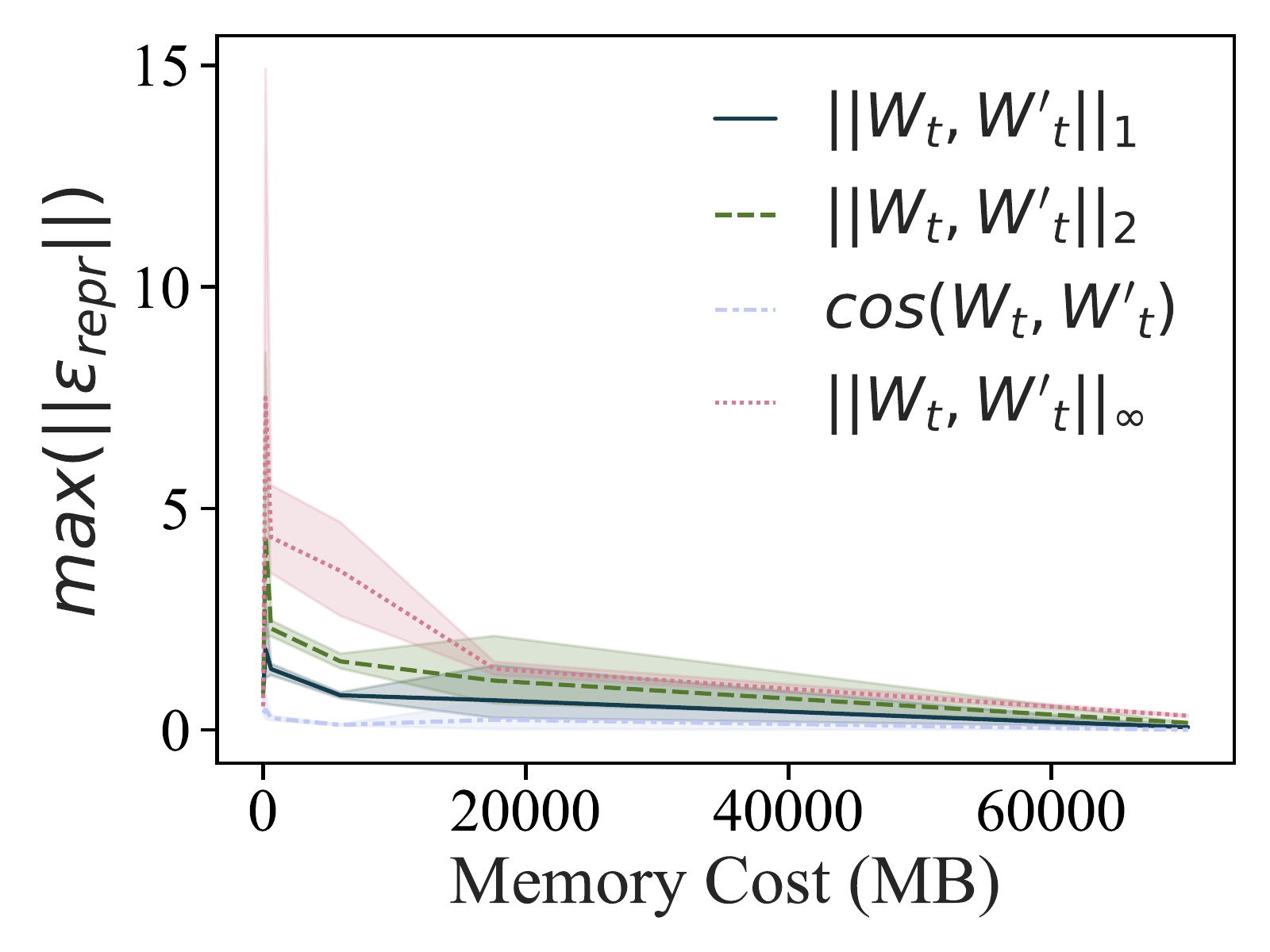}}
    \caption{This is the same as Figure~\ref{fig:reproducibility_storage} except the x-axis is in megabytes (MB). The memory cost is significantly higher for CIFAR-100 because we used a much deeper model than the one used for CIFAR-10.\vspace{-5mm}}
    \label{fig:reproducibility_memory}
\end{figure}

\begin{figure}[t]
    \centering
    \includegraphics[width=0.75\linewidth]{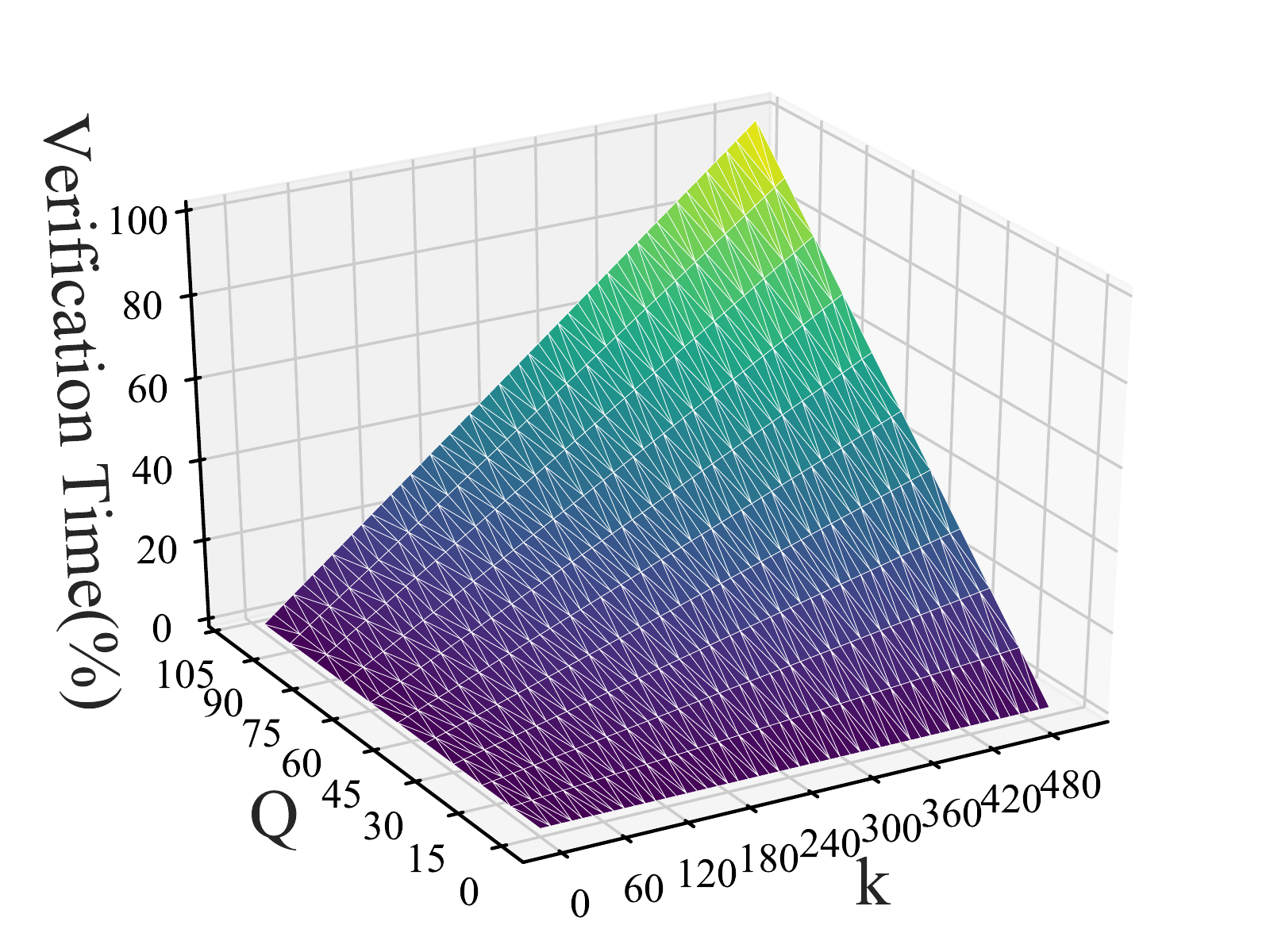}
    \caption{Analytical relation among verification time cost, checkpointing interval(k), and Q. Note here the verification time is measured in proportion to the training time (\ie $100\%$ means verifing the \prf takes the same time as training the model). By doing so, verification only depends on k, Q, and size of the training dataset. Therefore, this figure is applicable to both CIFAR-10 and CIFAR-100 (or any dataset with 50,000 training samples).}
    \label{fig:time_analysis}
\end{figure}

\begin{table}[b]
    \centering
    \begin{tabular}{c c | c c}
        && CIFAR-10 & CIFAR-100\\
        \toprule
        \multirow{4}{*}{\rotatebox[origin=c]{90}{\rd}} 
		&$\ell_{1}$ &$27204.55 (\pm57.384)$&$189093.15 (\pm558.821) $\\
		&$\ell_{2}$ &$71.431 (\pm0.243)$&$58.17 (\pm0.142) $\\
		&$\ell_{\infty}$ &$2.678 (\pm0.267)$&$0.898 (\pm0.135) $\\
		&cos &$0.83 (\pm0.005)$&$0.847 (\pm0.003) $\\
        \end{tabular}
    \caption{Reference distance, \rd, of CIFAR-10 and CIFAR-100. \rd is defined as the average distance between parameters of two models with the same architecture and dataset, but trained independently.\vspace{-7mm}}
    \label{tab:reference_distance}
\end{table}

\end{document}